\documentclass[10pt,journal,compsoc]{IEEEtran}

\usepackage{graphicx}
\usepackage{amsmath}
\usepackage{amssymb}
\usepackage{booktabs}

\usepackage[table]{xcolor}
\usepackage{xcolor}
\definecolor{cvprblue}{RGB}{82,208,83}
\usepackage[breaklinks,colorlinks,citecolor=cvprblue]{hyperref}

\usepackage{bm}
\usepackage{bbding}
\usepackage{amsmath}
\usepackage{algorithmic}
\usepackage{algorithm}
\usepackage{multirow}
\usepackage{graphicx}
\usepackage{wrapfig}
\usepackage{amsthm}
\usepackage{caption}
\usepackage{subcaption}
\usepackage{pifont}
\newcommand{\cmark}{\ding{51}}%
\newcommand{\xmark}{\ding{55}}%

\usepackage{bbding}

\usepackage{hhline}
\usepackage{enumitem}
\usepackage{lipsum}
\setlist[itemize]{leftmargin=0.11in}
\usepackage[capitalize]{cleveref}
\usepackage{pdfpages}

\newtheorem{proposition}{Proposition}

\definecolor{baselinecolor}{gray}{.9}
\newcommand{\baseline}[1]{\cellcolor{baselinecolor}{#1}}
\newlength\savewidth\newcommand\shline{\noalign{\global\savewidth\arrayrulewidth
  \global\arrayrulewidth 1.25pt}\hline\noalign{\global\savewidth\arrayrulewidth
  \global\arrayrulewidth 0.575pt}}

\ifCLASSOPTIONcompsoc
  \usepackage[nocompress]{cite}
\else
  \usepackage{cite}
\fi

\ifCLASSINFOpdf
\else
\fi
\vspace{7.5in}
\IEEEpubid{
\fbox{
\parbox{0.975\textwidth}{
\copyright 2024 IEEE. Personal use of this material is permitted. Permission from IEEE must be obtained for all other uses, in any current or future media, including\hfill
reprinting/republishing this material for advertising or promotional purposes, creating new collective works, for resale or redistribution to servers or lists, or reuse of\hfill
any copyrighted component of this work in other works. \vspace{-0.5mm}
}%
}}

\begin{document}

\title{EfficientTrain++: Generalized Curriculum Learning for Efficient Visual Backbone Training}


\author{
  Yulin~Wang,
  Yang~Yue,
  Rui~Lu,
  Yizeng~Han,
  Shiji~Song,~\IEEEmembership{Senior~Member,~IEEE},
  and~Gao~Huang$^{\dagger}$,~\IEEEmembership{Member,~IEEE}
  \IEEEcompsocitemizethanks{
  \IEEEcompsocthanksitem Y. Wang, Y. Yue, R. Lu, Y. Han, S. Song, and G. Huang are with the Department of Automation, BNRist, Tsinghua University, Beijing 100084, China. 
  G. Huang is also with Beijing Academy of Artificial Intelligence. 
  Email: wang-yl19@mails.tsinghua.edu.cn, gaohuang@tsinghua.edu.cn. 
  \IEEEcompsocthanksitem ${\dagger}$: corresponding author.
}}

%
%

\markboth{EfficientTrain++: Generalized Curriculum Learning for Efficient Visual Backbone Training}
{EfficientTrain++: Generalized Curriculum Learning for Efficient Visual Backbone Training}

\IEEEtitleabstractindextext{%
\begin{abstract}

  The superior performance of modern computer vision backbones (\emph{e.g.}, vision Transformers learned on ImageNet-1K/22K) usually comes with a costly training procedure. This study contributes to this issue by generalizing the idea of curriculum learning beyond its original formulation, \emph{i.e.}, training models using easier-to-harder data. Specifically, we reformulate the training curriculum as a soft-selection function, which uncovers progressively more difficult patterns within each example during training, instead of performing easier-to-harder sample selection. Our work is inspired by an intriguing observation on the learning dynamics of visual backbones: during the earlier stages of training, the model predominantly learns to recognize some `easier-to-learn' discriminative patterns in the data. These patterns, when observed through frequency and spatial domains, incorporate lower-frequency components, and the natural image contents without distortion or data augmentation. Motivated by these findings, we propose a curriculum where the model always leverages all the training data at every learning stage, yet the exposure to the `easier-to-learn' patterns of each example is initiated first, with harder patterns gradually introduced as training progresses. To implement this idea in a computationally efficient way, we introduce a cropping operation in the Fourier spectrum of the inputs, enabling the model to learn from only the lower-frequency components. Then we show that exposing the contents of natural images can be readily achieved by modulating the intensity of data augmentation. Finally, we integrate these two aspects and design curriculum learning schedules by proposing tailored searching algorithms. Moreover, we present useful techniques for deploying our approach efficiently in challenging practical scenarios, such as large-scale parallel training, and limited input/output or data pre-processing speed. The resulting method, EfficientTrain++, is simple, general, yet surprisingly effective. As an off-the-shelf approach, it reduces the training time of various popular models (\emph{e.g.}, ResNet, ConvNeXt, DeiT, PVT, Swin, CSWin, and CAFormer) by $\bm{1.5\!-\!3.0\times}$ on ImageNet-1K/22K without sacrificing accuracy. It also demonstrates efficacy in self-supervised learning (\emph{e.g.}, MAE). Code is available at: \url{https://github.com/LeapLabTHU/EfficientTrain}.

\end{abstract}

\begin{IEEEkeywords}
Deep networks, visual backbones, efficient training, curriculum learning.
\end{IEEEkeywords}}

\maketitle

\IEEEdisplaynontitleabstractindextext

%
\IEEEpeerreviewmaketitle

\section{Introduction}


The remarkable success of modern visual backbones is largely fueled by the interest in exploring big models on comprehensive, large-scale benchmark datasets \cite{He_2016_CVPR, 2016arXiv160806993H, dosovitskiy2021an, liu2021swin}. In particular, the recent introduction of vision Transformers (ViTs) scales up the number of model parameters to more than 1.8 billion, concurrently expanding the training data to 3 billion samples \cite{dosovitskiy2021an, DBLP:journals/corr/abs-2106-04560}. While this has facilitated the attainment of state-of-the-art accuracy, this huge-model and high-data regime results in a time-consuming and expensive training process. For example, it takes 2,500 TPUv3-core-days to train ViT-H/14 on JFT-300M \cite{dosovitskiy2021an}, which may be unaffordable for practitioners in both academia and industry. Moreover, the considerable power consumption for training deep learning models will lead to significant carbon emissions \cite{strubell2019energy, li2022automated}. Due to both economic and environmental concerns, there has been a growing demand for reducing the training cost of modern deep networks.

\begin{figure}[t]
    \begin{minipage}[t]{\linewidth}
        \centering
        \includegraphics[width=0.975\textwidth]{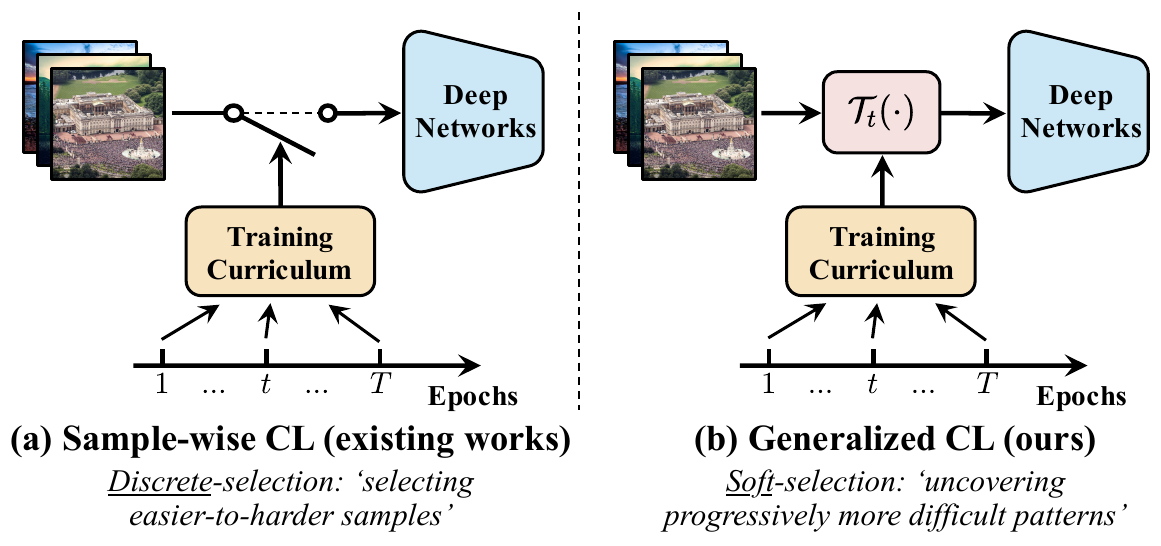}
        \vskip -0.075in
        \captionsetup{font={footnotesize}}
        \caption{\label{fig:ig1}\textbf{(a) Sample-wise curriculum learning (CL):} making a \textit{\underline{discrete}} decision on whether each example should be leveraged to train the model. 
        \textbf{(b) Generalized CL:} we consider a \textit{\underline{continuous}} function $\mathcal{T}_t(\cdot)$, which only exposes the \textit{`easier-to-learn'} patterns within each example at the beginning of training (\textit{e.g.}, \textit{lower-frequency} components; see: Section \ref{sec:EfficientTrain_sec4}), while gradually introducing relatively \textit{more difficult} patterns as learning progresses. 
        }
    \end{minipage}
    \vskip -0.125in
  \end{figure}


In this paper, we contribute to this issue by revisiting the concept of curriculum learning \cite{bengio2009curriculum}, which reveals that a model can be trained efficiently by starting with the easier aspects of a given task or certain easier subtasks, and increasing the difficulty level gradually. The majority of extant research efforts implement this approach by introducing easier-to-harder examples progressively during training \cite{wang2021survey, soviany2022curriculum, graves2017automated, hacohen2019power, jiang2015self, jiang2018mentornet, gong2016multi, fan2018learning}. However, obtaining a light-weighted and generalizable difficulty measurer is typically non-trivial \cite{wang2021survey, soviany2022curriculum}. In general, these methods have not exhibited the capacity to be a universal efficient training technique for modern visual backbones.



In contrast to prior works, this paper seeks a simple yet broadly applicable efficient learning approach with the potential for widespread implementation. To attain this goal, our investigation delves into a generalization of curriculum learning. In specific, we extend the notion of training curricula beyond only differentiating between `easier' and `harder' examples, and adopt a more flexible hypothesis, which indicates that the discriminative features of each training sample comprise both `easier-to-learn' and `harder-to-learn' patterns. Instead of making a \textit{discrete} decision on whether each example should appear in the training set, we argue that it would be more proper to establish a \textit{continuous} function that adaptively extracts the simpler and more learnable discriminative patterns within every example. In other words, a curriculum may always leverage all examples at any stage of learning, but it should eliminate the relatively more difficult or complex patterns within inputs at earlier learning stages. Our idea is illustrated in Figure \ref{fig:ig1}.


Driven by our hypothesis, a straightforward yet surprisingly effective algorithm is derived. We first demonstrate that the `easier-to-learn' patterns incorporate the lower-frequency components of images. We further show that a lossless extraction of these components can be achieved by introducing a cropping operation in the frequency domain. This operation not only retains exactly all the lower-frequency information, but also yields a smaller input size for the model to be trained. By triggering this operation at earlier training stages, the overall computational/time cost for training can be considerably reduced while the final performance of the model will not be sacrificed. Moreover, we show that the original information before heavy data augmentation is more learnable, and hence starting the training with weaker augmentation techniques is beneficial. Finally, these theoretical and experimental insights are integrated into a unified `\emph{EfficientTrain}' learning curriculum by leveraging a greedy-search algorithm.


Built upon our aforementioned exploration, we further introduce an enhanced EfficientTrain++ approach that improves EfficientTrain from two aspects. First, we propose an advanced searching algorithm, which not only dramatically saves the cost for solving for the curriculum learning schedule, but also contributes to a well-generalizable and more effective resulting curriculum. Second, we present an efficient low-frequency down-sampling operation, aiming to alleviate the problem of high CPU-GPU input/output (I/O) costs caused by EfficientTrain. Beyond these methodological advancements, we develop two implementation techniques for EfficientTrain++. They are useful for achieving more significant practical training speedup, \emph{e.g.}, through 1) facilitating efficient large-scale parallel training using an increasing number of GPUs (\emph{e.g.}, 32 or 64 GPUs), and 2) reducing the data pre-processing loads for CPUs/memory.


One of the most appealing advantages of our method may be its simplicity and generalizability. Our method can be conveniently applied to most deep networks \emph{without additional hyper-parameter tuning}, but significantly improves their training efficiency. Empirically, for the supervised learning on ImageNet-1K/22K \cite{deng2009imagenet}, EfficientTrain++ reduces the training cost of a wide variety of visual backbones (\emph{e.g.}, ResNet \cite{He_2016_CVPR}, ConvNeXt \cite{liu2022convnet}, DeiT \cite{touvron2021training}, PVT \cite{wang2021pyramid}, Swin \cite{liu2021swin}, CSWin \cite{dong2021cswin}, and CAFormer \cite{yu2022metaformer}) by $\bm{1.5\!-\!3.0\times}$, while achieving competitive or better performance compared to the baselines. Importantly, our method is also effective for self-supervised learning (\emph{e.g.}, MAE \cite{he2022masked}).

Parts of the results in this paper were published originally in its conference version \cite{wang2023efficienttrain}. However, this paper extends our earlier works in several important aspects:
\begin{itemize}
    \item An improved EfficientTrain++ approach is presented. We introduce a computational-constrained sequential searching algorithm for solving for the training curriculum (Section \ref{sec:ET_plus_1}), which is not only significantly more efficient, but also outperforms the greedy-search algorithm in EfficientTrain in terms of the resulting curriculum. Moreover, we propose an efficient low-frequency down-sampling operation (Section \ref{sec:ET_plus_2}), which effectively addresses the issue of high CPU-GPU I/O costs caused by EfficientTrain. We conduct new experiments to investigate the effectiveness of these two components (Tables \ref{tab:ET_vs_ETplus}, \ref{tab:ablation}, \ref{tab:ET_plus_abl}, \ref{tab:design_of_alg2}).
    \item We develop two implementation techniques for EfficientTrain++ (Section \ref{sec:ET_plus_3}). The first one improves the scalability of the practical training efficiency of EfficientTrain++ with respect to the number of GPUs used for training. The second one can significantly reduce the data pre-processing loads for CPUs/memory, while preserving competitive performance. Extensive experimental evidence has been provided to demonstrate their utility (Tables \ref{tab:results_early_large_bs}, \ref{tab:replay_buffer}). 
    \item Our experiments have been improved by extensively evaluating EfficientTrain++, \emph{e.g.}, via supervised learning on ImageNet-1K (Tables \ref{tab:img_1k_main_result}, \ref{tab:high_acc}, and Figure \ref{fig:acc_vs_cost}), ImageNet-22K pre-training (Table \ref{tab:img_22K_main_result}), self-supervised learning (Table \ref{tab:mae_result}), transfer learning (Tables \ref{tab:transferability}, \ref{tab:coco}, \ref{tab:ade20k}), and being compared with existing approaches (Table \ref{tab:img1k_vs_baseline}). Besides, new analytical investigations have been added (Tables \ref{tab:alg2_generalizability}, \ref{tab:compatibility_with_sample_selection}). These comprehensive results demonstrate that EfficientTrain++ outperforms both vanilla EfficientTrain and other competitive baselines, attaining state-of-the-art training efficiency. 
\end{itemize}

\section{Related Work}
\label{sec:related}

\textbf{Curriculum learning}
is a training paradigm inspired by the organized learning order of examples in human curricula \cite{elman1993learning, krueger2009flexible, bengio2009curriculum}. This idea has been widely explored in the context of training deep networks from easier data to harder data \cite{kumar2010self, graves2017automated, fan2018learning, hacohen2019power, platanios2019competence, wang2021survey, soviany2022curriculum}. Typically, a pre-defined \cite{bengio2009curriculum, chen2015webly, wei2016stc, tudor2016hard} or automatically-learned \cite{kumar2010self, tullis2011effectiveness, jiang2014easy, graves2017automated, weinshall2018curriculum, fan2018learning, jiang2018mentornet, ren2018learning, zhang2019leveraging, hacohen2019power, matiisen2019teacher} difficulty measurer is deployed to differentiate between easier and harder samples, while a scheduler \cite{bengio2009curriculum, graves2017automated, platanios2019competence, wang2021survey} is defined to determine when and how to introduce harder training data.

For example, DIHCL \cite{NEURIPS2020_62000dee} and InfoBatch \cite{qin2024infobatch} determine the hardness of each sample by leveraging its online training loss or the change in model outputs. MCL \cite{zhou2018minimax} adaptively selects a sequence of training subsets by repeatedly solving joint continuous-discrete minimax optimization problems, whose objective combines both a continuous training loss that reflects training set hardness and a discrete submodular promoter of diversity for the chosen subset. CurriculumNet \cite{guo2018curriculumnet} designs a curriculum that measures the difficulty of different data using the distribution density in the feature space obtained by training an initial model with all data. Our work is based on a similar `starting small' spirit \cite{elman1993learning} to these methods, but we reformulate the training curriculum as a soft-selection function that uncovers progressively more difficult patterns within each example, rather than performing easier-to-harder data selection.

Our work is also related to curriculum by smoothing \cite{sinha2020curriculum}, curriculum dropout \cite{morerio2017curriculum} and label-similarity curriculum \cite{dogan2020label}, which do not perform example selection as well. However, our method is orthogonal to them since we propose to reduce the training cost by modifying the model inputs. In contrast, they focus on regularizing the deep features during training (\emph{e.g.}, via anti-aliasing smoothing \cite{sinha2020curriculum} and feature dropout \cite{morerio2017curriculum}), or progressively configuring the learning objective (\emph{e.g.}, through computing the similarity of the word embeddings of labels \cite{dogan2020label}).



\textbf{Efficient progressive learning algorithms.}
It has been observed that smaller networks are typically more efficient to train at earlier learning stages. More specifically, previous research has revealed that deep networks can be trained efficiently with the increasing model size during training, \emph{e.g.}, a progressively growing number or layers \cite{fahlman1989cascade, lengelle1996training, bengio2006greedy, hinton2006fast, simonyan2014very, wang2017deep, karras2018progressive, li2022automated}, a growing network width \cite{chen2015net2net}, or a dynamically changing topology of network connections \cite{wei2016network, wei2020modularized}. For example, AutoProg \cite{li2022automated} and budgeted ViT \cite{xia2023budgeted} demonstrate that the training efficiency of vision Transformers can be effectively improved by increasing network depth or width progressively during training. LipGrow \cite{dong2020towards} conducts theoretical analyses of network growth from an ordinary differential equation perspective, and proposes a novel performance measure, yielding an adaptive training algorithm for residual networks, which increases depth automatically and accelerates training. Moreover, the budgeted training algorithm \cite{Li2020Budgeted} shows that properly configuring learning rate schedules is beneficial for efficient training. In general, our work is orthogonal to these efforts since we do not modify network architectures or learning rates.


Furthermore, some works \cite{tan2021efficientnetv2, Touvron2022DeiTIR, li2022automated} propose to down-sample early training inputs to save training cost.
Our work differs from them in several important aspects: 
1) EfficientTrain++ is drawn from a distinctly different motivation of generalized curriculum learning, based on which we present a novel frequency-domain-inspired analysis; 
2) we introduce a cropping operation in the frequency domain, which is not only theoretically different from image down-sampling (see: Proposition \ref{prop:downsampling}), but also outperforms it empirically (see: Table \ref{tab:ablation} (b)); 
3) we propose a novel efficient searching algorithm (see: Algorithm \ref{alg:greedy_search_v2}) and design an EfficientTrain++ curriculum, achieving a significantly higher training efficiency than existing works on a variety of state-of-the-art models (see: Table \ref{tab:img1k_vs_baseline}). 
Besides, FixRes \cite{touvron2019FixRes} shows that a smaller training resolution improves accuracy by fixing the discrepancy between the scale of training and test inputs. However, we do not borrow gains from FixRes as we adopt a standard resolution at final learning stages. Our method is actually orthogonal to FixRes (see: Table \ref{tab:img1k_vs_baseline}).

AutoProg \cite{li2022automated} proposes a novel and effective algorithm that fine-tunes the model at each training stage to search for a proper progressive learning strategy. Our work is relevant to AutoProg \cite{li2022automated} in this paradigm. However, we introduce a novel formulation of computational-constrained searching (\emph{i.e.}, maximizing validation accuracy under a fixed ratio of training cost saving; see: Section \ref{sec:ET_plus_1}), leading to a simple searching objective as well as a significantly better training curriculum. An analysis on this issue is given in Table \ref{tab:design_of_alg2}.

\textbf{Frequency-based analysis of deep networks.}
Our observation that deep networks tend to capture the low-frequency components first is inline with \cite{wang2020high}, but the discussions in \cite{wang2020high} focus on the robustness of ConvNets and are mainly based on some small models and tiny datasets. Towards this direction, several existing works also explore decomposing the inputs of models in the frequency domain \cite{yin2019fourier, lopes2019improving, paul2021vision} in order to understand or improve the robustness of the networks. In contrast, our aim is to improve the training efficiency of modern deep visual backbones.




\section{Generalized Curriculum Learning}

\label{sec:GCL}

As uncovered in previous research, machine learning algorithms generally benefit from a `starting small' strategy, \emph{i.e.}, to first learn certain easier aspects of a task, and increase the level of difficulty progressively \cite{elman1993learning, krueger2009flexible, bengio2009curriculum}. The dominant implementation of this idea, curriculum learning, proposes to introduce gradually more difficult examples during training \cite{hacohen2019power, wang2021survey, soviany2022curriculum}. In specific, a curriculum is defined on top of the training process to determine whether or not each sample should be leveraged at a given epoch (Figure \ref{fig:ig1} (a)).

\textbf{On the limitations of sample-wise curriculum learning.}
Although curriculum learning has been widely explored from the lens of the sample-wise regime, its extensive application is usually limited by two major issues. \textit{First}, differentiating between `easier' and `harder' training data is non-trivial. It typically requires deploying additional deep networks as a `teacher' or exploiting specialized automatic learning approaches \cite{kumar2010self, graves2017automated, fan2018learning, jiang2018mentornet, hacohen2019power}. The resulting implementation complexity and the increased overall computational cost are both noteworthy weaknesses in terms of improving the training efficiency. \textit{Second}, it is challenging to attain a principled approach that specifies which examples should be attended to at the earlier stages of learning. As a matter of fact, the `easy to hard' strategy is not always helpful \cite{wang2021survey}. The hard-to-learn samples can be more informative and may be beneficial to be emphasized in many cases \cite{freund1996experiments, alain2015variance, loshchilov2015online, shrivastava2016training, gopal2016adaptive, NEURIPS2020_62000dee}, sometimes even leading to a `hard to easy' anti-curriculum \cite{pi2016self, braun2017curriculum, zhou2018minimax, zhang2018empirical, wang2019dynamically, liu2022acpl}.

Our work is inspired by the above two issues. In the following, we start by proposing a generalized hypothesis for curriculum learning, aiming to address the second issue. Then we demonstrate that an implementation of our idea naturally addresses the first issue.


\textbf{Generalized curriculum learning.}
We argue that simply measuring the easiness of training samples tends to be ambiguous and may be insufficient to reflect the effects of a sample on the learning process. As aforementioned, even the difficult examples may provide beneficial information for guiding the training, and they do not necessarily need to be introduced after easier examples. To this end, we hypothesize that every training sample, either `easier' or `harder', contains both \emph{easier-to-learn} or \emph{more accessible} patterns, as well as certain \emph{difficult} discriminative information which may be challenging for the deep networks to capture. Ideally, a curriculum should be a continuous function on top of the training process, which starts with a focus on the `easiest' patterns of the inputs, while the `harder-to-learn' patterns are gradually introduced as learning progresses. 

A formal illustration is shown in Figure \ref{fig:ig1} (b). Any input data $\boldsymbol{X}$ will be processed by a transformation function $\mathcal{T}_t(\cdot)$ conditioned on the training epoch $t\ (t\!\leq\!T)$ before being fed into the model, where $\mathcal{T}_t(\cdot)$ is designed to dynamically filter out the excessively difficult and less learnable patterns within the training data. We always let $\mathcal{T}_T(\boldsymbol{X})\!=\!\boldsymbol{X}$. Notably, our approach can be seen as a generalized form of the sample-wise curriculum learning. It reduces to example-selection by setting $\mathcal{T}_t(\boldsymbol{X}) \!\in\! \{\emptyset, \boldsymbol{X}\}$. 

\textbf{Overview.}
In the rest of this paper, we will demonstrate that an algorithm drawn from our hypothesis dramatically improves the implementation efficiency and generalization ability of curriculum learning. We will show that a zero-cost criterion pre-defined by humans is able to effectively measure the difficulty level of different patterns within images. Based on such simple criteria, even a surprisingly straightforward implementation of introducing `easier-to-harder' patterns yields significant and consistent improvements on the training efficiency of modern visual backbones.

\section{The EfficientTrain Approach}
\label{sec:EfficientTrain_sec4}

To obtain a training curriculum following our aforementioned hypothesis, we need to solve two challenges: 1) identifying the `easier-to-learn' patterns and designing transformation functions to extract them; 2) establishing a curriculum learning schedule to perform these transformations dynamically during training. This section will demonstrate that a proper transformation for 1) can be easily found in both the frequency and the spatial domain, while 2) can be addressed with a greedy search algorithm. Implementation details of the experiments in this section: see Appendix \textcolor{red}{A}.

\subsection{Easier-to-learn Patterns: Frequency Domain}
\label{sec:freq_inspired}

Image-based data can naturally be decomposed in the frequency domain \cite{campbell1968application, sweldens1998lifting, mallat1999wavelet, chen2019drop}. In this subsection, we reveal that the patterns in the lower-frequency components of images, which describe the smoothly changing contents, are relatively easier for the networks to learn to recognize.

\begin{figure}[t!]
  \begin{center}
  \centerline{
    \hskip 0.15in
    \includegraphics[width=0.985\linewidth]{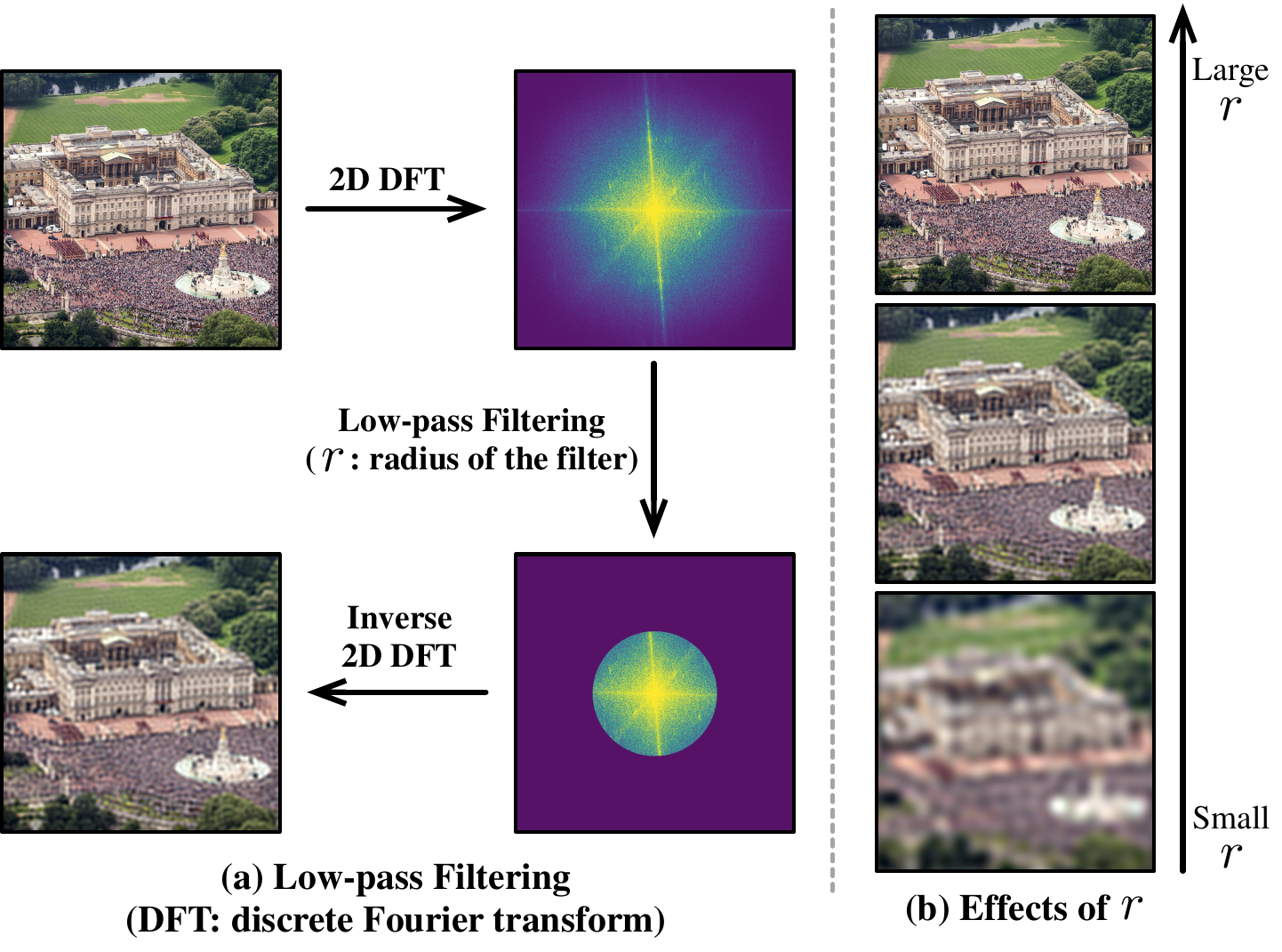}
    }
  \vskip -0.09in
  \captionsetup{font={footnotesize}}
  \caption{\textbf{Low-pass filtering.} Following \cite{wang2020high}, we adopt a circular filter. \label{fig:low_pass_filtering}
  }
  \end{center}
  \vspace{-4ex}
\end{figure}
\begin{figure}[t]
  \begin{center}
  \centerline{\includegraphics[width=1\linewidth]{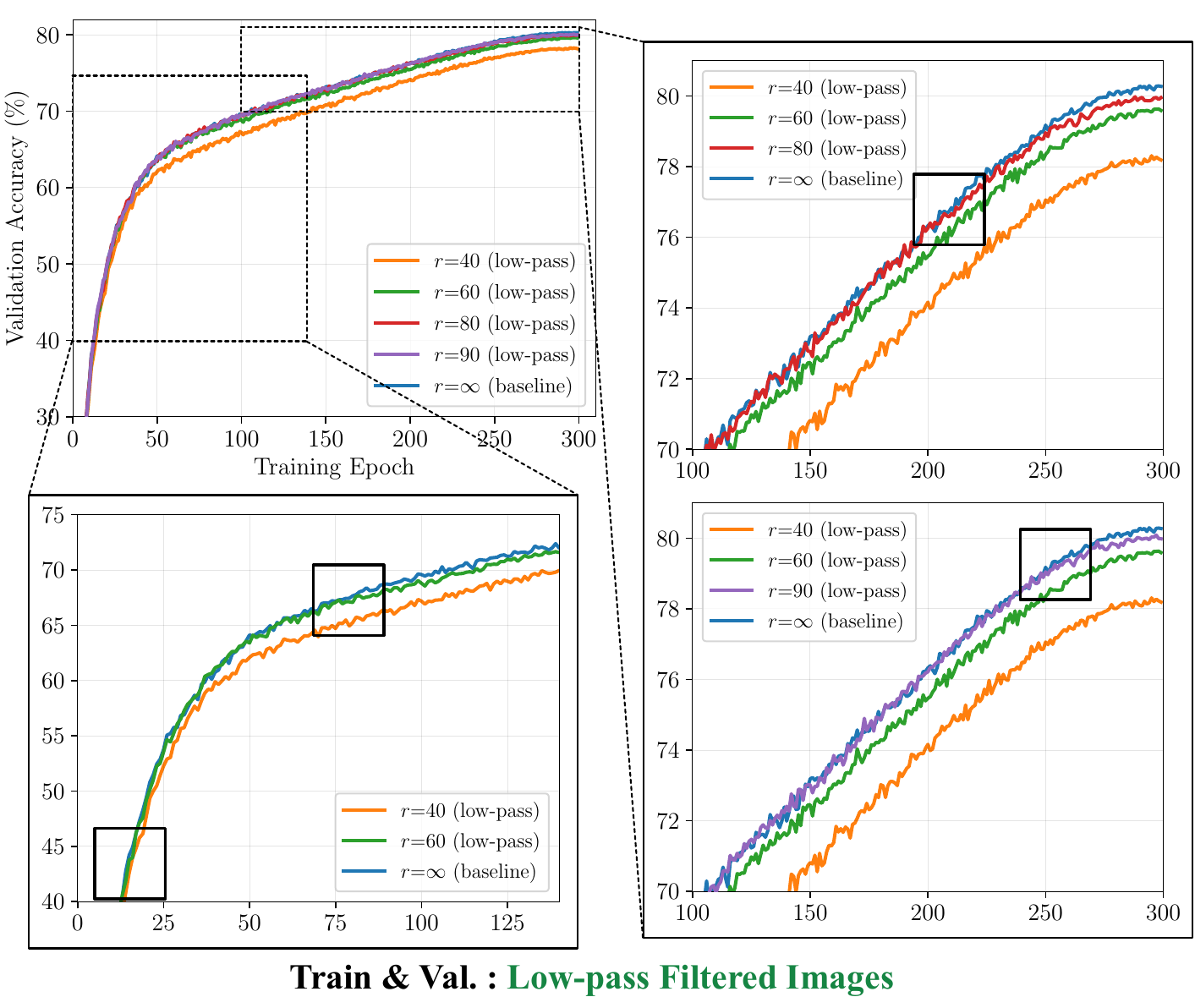}}
  \vskip -0.09in
  \captionsetup{font={footnotesize}}
  \caption{\textbf{Ablation study results with low-pass filtering} ($r$: bandwidth of the filter, see Figure \ref{fig:low_pass_filtering} for details). We ablate the higher-frequency components of the inputs for a DeiT-Small \cite{touvron2021training}, and present the curves of validation accuracy v.s. training epochs on ImageNet-1K. We highlight the separation points of the curves with black boxes.\label{fig:low_pass_training} 
  }
  \end{center}
  \vspace{-4ex}
\end{figure}

\textbf{Ablation studies with the low-pass filtered input data.}
We first consider an ablation study, where the low-pass filtering is performed on the data we use. As shown in Figure \ref{fig:low_pass_filtering} (a), we map the images to the Fourier spectrum with the lowest frequency at the centre, set all the components outside a centred circle (radius: $r$) to zero, and map the spectrum back to the pixel space. Figure \ref{fig:low_pass_filtering} (b) illustrates the effects of $r$. The curves of accuracy v.s. training epochs on top of the low-pass filtered data are presented in Figure \ref{fig:low_pass_training}. Here both training and validation data is processed by the filter to ensure the compatibility with the i.i.d. assumption.

\textbf{Lower-frequency components are captured first.}
The models in Figure \ref{fig:low_pass_training} are imposed to leverage only the lower-frequency components of the inputs. However, an appealing phenomenon arises: their training process is approximately identical to the original baseline at the beginning of training. Although the baseline finally outperforms, this tendency starts midway in the training process, instead of from the very beginning. In other words, the learning behaviors at earlier epochs remain unchanged even though the higher-frequency components of images are eliminated. 
Moreover, consider increasing the filter bandwidth $r$, which preserves progressively more information about the images from the lowest frequency. The separation point between the baseline and the training process on low-pass filtered data moves towards the end of training. To explain these observations, we postulate that, \emph{in a natural learning process where the input images contain both lower- and higher-frequency information, a model tends to first learn to capture the lower-frequency components, while the higher-frequency information is gradually exploited on the basis of them}.



\begin{figure}[!t]
  \begin{center}
  \begin{minipage}{.525\linewidth}
    \centering
    \includegraphics[width=\textwidth]{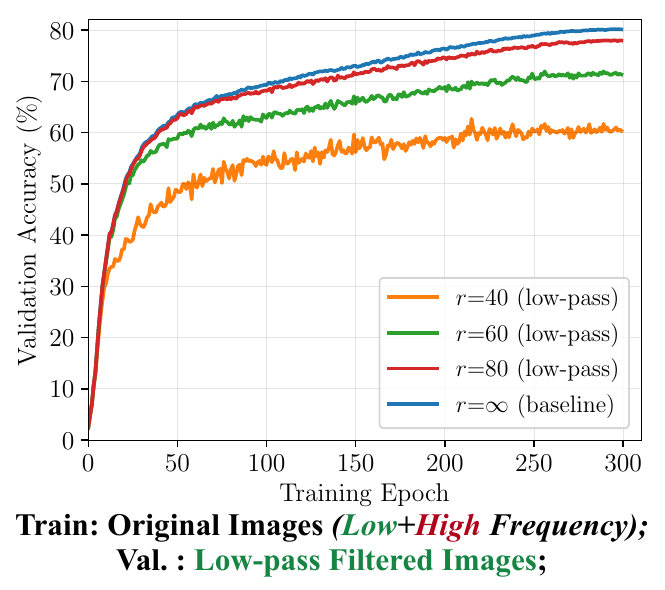}	  
  \end{minipage}
  \hspace{-1ex}
  \begin{minipage}{.47\linewidth} 
    \centering 
    \vskip 0.1in
    \captionsetup{font={footnotesize}}
    \caption{\textbf{Performing low-pass filtering only on the validation inputs} (other setups are the same as Figure \ref{fig:low_pass_training}). We train a model using the original images without any filtering (\emph{i.e.}, containing both lower- and higher-frequency components), and evaluate all the intermediate checkpoints on low-pass filtered validation sets with varying bandwidths. \label{fig:low_pass_training_2}}
  \end{minipage} 
  \end{center}
  \vspace{-2ex}
\end{figure}

\textbf{More evidence.}
Our assumption can be further confirmed by a well-controlled experiment. Consider training a model using original images, where lower/higher-frequency components are simultaneously provided. In Figure \ref{fig:low_pass_training_2}, we evaluate all the intermediate checkpoints on low-pass filtered validation sets with varying bandwidths. Obviously, at earlier epochs, only leveraging the low-pass filtered validation data does not degrade the accuracy. This phenomenon suggests that the learning process starts with a focus on the lower-frequency information, even though the model is always accessible to higher-frequency components during training. Furthermore, in Table \ref{tab:direct_evidence}, we compare the accuracies of the intermediate checkpoints on low/high-pass filtered validation sets. The accuracy on the low-pass filtered validation set grows much faster at earlier training stages, even though the two final accuracies are the same.





\begin{table}[t]
  \centering
  \begin{footnotesize}
  \setlength{\tabcolsep}{1.5mm}{
  \renewcommand\arraystretch{1.3}
  \resizebox{0.91\columnwidth}{!}{
  \begin{tabular}{c|ccccccccc}
  Training Epoch &	20$^{\textnormal{th}}$ &	50$^{\textnormal{th}}$ &	100$^{\textnormal{th}}$ &	200$^{\textnormal{th}}$ &	300$^{\textnormal{th}}$ (end) \\
  \shline
   Low-pass Filtered Val. Set &	\textbf{46.9\%} &	\textbf{58.8\%} &	\textbf{63.1\%} &	\textbf{68.6\%} &	\underline{71.3\%} \\
   High-pass Filtered Val. Set &	23.9\% &	43.5\% &	53.5\% &	64.3\% &	\underline{71.3\%}
   \\
  \end{tabular}}}
  \vskip -0.05in
  \captionsetup{font={footnotesize}}
  \caption{\textbf{Comparisons: evaluating the model in Figure \ref{fig:low_pass_training_2} on low/high-pass filtered validation sets}. Note that the model is trained using the original images without any filtering. The bandwidths of the low/high-pass filters are configured to make the finally trained model (300$^{\textnormal{th}}$ epoch) have the same accuracy on the two validation sets (\emph{i.e.}, \underline{71.3\%}).\label{tab:direct_evidence}}
  \end{footnotesize}
  \vskip -0.05in
\end{table}


\textbf{Frequency-based curricula.}
Returning to our hypothesis in Section \ref{sec:GCL}, we have shown that lower-frequency components are naturally captured earlier. Hence, it would be straightforward to consider them as a type of the `easier-to-learn' patterns. This begs a question: can we design a training curriculum, which starts with providing only the lower-frequency information for the model, while gradually introducing higher-frequency components? We investigate this idea in Table \ref{tab:frequency_based_curriculum}, where we perform low-pass filtering on the training data only in a given number of the beginning epochs. The rest of the training process remains unchanged.

\begin{table}[h!]
  \centering
  \begin{footnotesize}
  \setlength{\tabcolsep}{3.5mm}{
  \renewcommand\arraystretch{1.35}
  \resizebox{0.93\columnwidth}{!}{
    \begin{tabular}{cc|ccc}
       \multicolumn{2}{c}{Curricula (ep: epoch)} & \multicolumn{3}{c}{Final Top-1 Accuracy}   \\[-0.5ex]
       Low-pass Filtered & Original & \multicolumn{3}{c}{($r$: filter bandwidth)}   \\[-0.65ex]
       Training Data & Training Data & $r$=40 & $r$=60 & $r$=80 \\
       \shline
       1$^{\textnormal{st}}$ -- 300$^{\textnormal{th}}$ ep & -- & 78.3\% & 79.6\% & 80.0\%  \\
       1$^{\textnormal{st}}$ -- 225$^{\textnormal{th}}$ ep & 226$^{\textnormal{th}}$ -- 300$^{\textnormal{th}}$ ep & 79.4\% & \underline{80.2\%} 
       & \underline{80.5\%}  \\
       1$^{\textnormal{st}}$ -- 150$^{\textnormal{th}}$ ep & 151$^{\textnormal{th}}$ -- 300$^{\textnormal{th}}$ ep & \underline{80.1\%} & {80.2\%}
       & {80.6\%}  \\
       1$^{\textnormal{st}}$ -- \ 75$^{\textnormal{th}}$\ \  ep & \ \ 76$^{\textnormal{th}}$\  -- 300$^{\textnormal{th}}$ ep &{80.3\%}  & {80.4\%} & {80.6\%}  \\
       \hline
       -- & \ \ \ \ 1$^{\textnormal{st}}$\  -- 300$^{\textnormal{th}}$ ep & \multicolumn{3}{c}{\textit{80.3\% (baseline)}}  \\
      \end{tabular}}}
      \vskip -0.05in
      \captionsetup{font={footnotesize}}
      \captionof{table}{\textbf{Results with the straightforward frequency-based training curricula} (DeiT-Small \cite{touvron2021training} on ImageNet-1K). {{Observation}: one can eliminate the higher-frequency components of the inputs in 50-75\% of the training process without sacrificing the final accuracy} (see: comparisons between the \emph{baseline} and the \underline{underlined} data).}
      \label{tab:frequency_based_curriculum}
  \end{footnotesize}
  \vskip 0.025in
\end{table}

\begin{figure*}[!t]
  \begin{center}
    \begin{minipage}{0.2645\linewidth}
      \includegraphics[width=\textwidth]{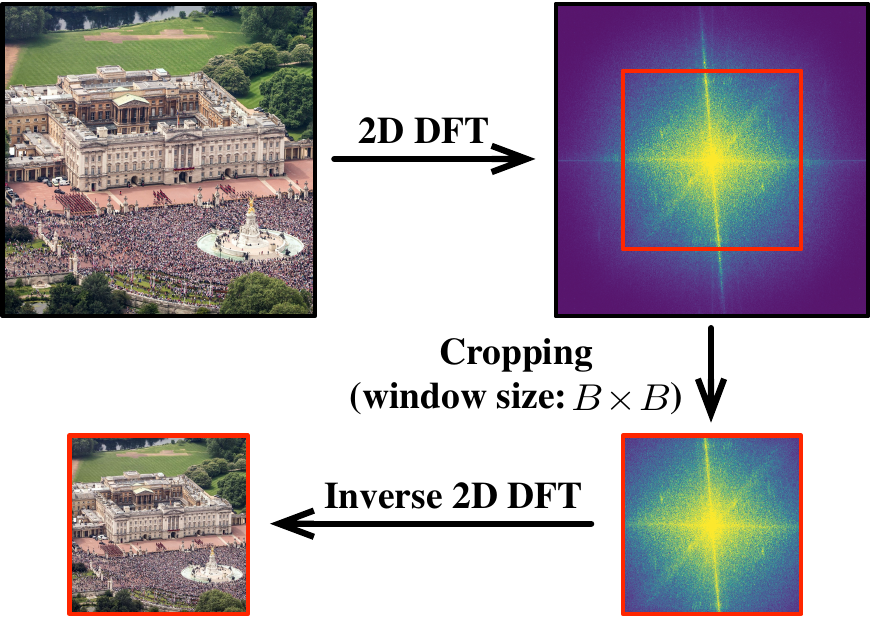}
    \vskip -0.09in
    \captionsetup{font={footnotesize}}
    \caption{\label{fig:freq_crop}\textbf{Low-frequency cropping in the frequency domain} (${B}^2$: bandwidth).}
  \end{minipage}
  \hspace{0.25ex}
  \begin{minipage}{.7235\linewidth}
      \centering
      \begin{footnotesize}
      \setlength{\tabcolsep}{0.75mm}{
      \vspace{4pt}
      \vskip 0.025in
      \renewcommand\arraystretch{1.6}
      \resizebox{\linewidth}{!}{
        \begin{tabular}{cc|cccc|cccc}
           \multicolumn{2}{c}{Curricula (ep: epoch)} & \multicolumn{8}{c}{Final Top-1 Accuracy / Relative Computational Cost for Training (compared to the baseline)}   \\[-0.25ex]
           Low-frequency & Original Training & \multicolumn{4}{c|}{DeiT-Small \cite{touvron2021training}} & \multicolumn{4}{c}{Swin-Tiny \cite{liu2021swin}}   \\[-0.65ex]
           Cropping (${B}^2$) & Data ($B\!=\!224$) & $B\!=\!96$ & $B\!=\!128$ & $B\!=\!160$ & $B\!=\!192$ & $B\!=\!96$ & $B\!=\!128$ & $B\!=\!160$ & $B\!=\!192$ \\
          \shline
           1$^{\textnormal{st}}$ -- 300$^{\textnormal{th}}$ ep & -- & 70.5\% \!/\! 0.18 & 75.3\% \!/\! 0.31 & 77.9\% \!/\! 0.49 & 79.1\% \!/\! 0.72 & 73.3\% \!/\! 0.18 & 76.8\% \!/\! 0.32 & 78.9\% \!/\! 0.50 & 80.5\% \!/\! 0.73 \\
           1$^{\textnormal{st}}$ -- 225$^{\textnormal{th}}$ ep & 226$^{\textnormal{th}}$ -- 300$^{\textnormal{th}}$ ep & 78.7\% \!/\! 0.38 & 79.6\% \!/\! 0.48 & 80.0\% \!/\! 0.62 & \underline{80.3\% \!/\! 0.79} & 80.0\% \!/\! 0.38 & 80.5\% \!/\! 0.49 & 81.0\% \!/\! 0.63 & \underline{81.2\% \!/\! 0.80} \\
           1$^{\textnormal{st}}$ -- 150$^{\textnormal{th}}$ ep & 151$^{\textnormal{th}}$ -- 300$^{\textnormal{th}}$ ep & 79.2\% \!/\! 0.59  & 79.8\% \!/\! 0.66 & \underline{80.3\% \!/\! 0.75} & {80.4\% \!/\! 0.86} & 80.9\% \!/\! 0.59 & 80.9\% \!/\! 0.66 & \underline{81.2\% \!/\! 0.75} & 81.3\% \!/\! 0.86 \\
           1$^{\textnormal{st}}$ -- \ 75$^{\textnormal{th}}$\ \  ep & \ \ 76$^{\textnormal{th}}$\  -- 300$^{\textnormal{th}}$ ep & 79.6\% \!/\! 0.79 & \underline{80.2\% \!/\! 0.83} & {80.4\% \!/\! 0.87} & {80.3\% \!/\! 0.93} & \underline{81.2\% \!/\! 0.79} & \underline{81.2\% \!/\! 0.83} & 81.3\% \!/\! 0.88 & 81.3\% \!/\! 0.93 \\
          \hline
           -- & \ \ \ \ 1$^{\textnormal{st}}$\  -- 300$^{\textnormal{th}}$ ep & \multicolumn{4}{c|}{\textit{80.3\% \!/\! 1.00 (baseline)}}  & \multicolumn{4}{c}{\textit{81.3\% \!/\! 1.00 (baseline)}} \\
          \end{tabular}}}
          \vskip -0.025in
          \captionsetup{font={footnotesize}}
          \captionof{table}{\textbf{Results on ImageNet-1K with the low-pass filtering in Table \ref{tab:frequency_based_curriculum} replaced by the low-frequency cropping}, which \emph{yields competitive accuracy with a significantly reduced training cost} (see: \underline{underlined} data).}
          \label{tab:resolution_based_curriculum}
      \end{footnotesize}
      \vskip -0.35in
  \end{minipage} 
  \end{center}
  \vspace{-4ex}
\end{figure*}


\textbf{Learning from low-frequency information efficiently.}
At the first glance, the results in Table \ref{tab:frequency_based_curriculum} may be less dramatic, \emph{i.e.}, by processing the images with a properly-configured low-pass filter at earlier epochs, the accuracy is moderately improved. However, an important observation is noteworthy: the final accuracy of the model can be largely preserved even with aggressive filtering (\emph{e.g.}, $r\!=\!40, 60$) performed in 50-75\% of the training process. This phenomenon turns our attention to training efficiency.
At earlier learning stages, it is harmless to train the model with only the lower-frequency components. These components incorporate only a selected subset of all the information within the original input images. Hence, \textit{can we enable the model to learn from them efficiently with less computational cost than processing the original inputs?}
As a matter of fact, this idea is feasible, and we may have at least two approaches.




$\bm{\bullet}$\ 1) \textbf{Down-sampling}. Approximating the low-pass filtering in Table \ref{tab:frequency_based_curriculum} with image down-sampling may be a straightforward solution. Down-sampling preserves much of the lower-frequency information, while it quadratically saves the computational cost for a model to process the inputs \cite{yang2020resolution, wang2020glance, wang2021not}. However, it is not an operation tailored for extracting lower-frequency components. Theoretically, it preserves some of the higher-frequency components as well (see: Proposition \ref{prop:downsampling}). Empirically, we observe that this issue degrades the performance (see: Table \ref{tab:ablation} (b)).


$\bm{\bullet}$\ 2) \textbf{Low-frequency cropping} (see: Figure \ref{fig:freq_crop}). We propose a more precise approach that extracts exactly all the lower-frequency information. Consider mapping an $H\!\times\!W$ image $\boldsymbol{X}$ into the frequency domain with the 2D discrete Fourier transform (DFT), obtaining an $H\!\times\!W$ Fourier spectrum, where the value in the centre denotes the strength of the component with the lowest frequency. The positions distant from the centre correspond to higher-frequency. We crop a $B\!\times\!B$ patch from the centre of the spectrum, where $B$ is the window size ($B\!<\!H, W$). Since the patch is still centrosymmetric, we can map it back to the pixel space with the inverse 2D DFT, obtaining a $B\!\times\!B$ new image $\boldsymbol{X}_{\textnormal{c}}$, \emph{i.e.},
\begin{equation}
    \boldsymbol{X}_{\textnormal{c}}=\mathcal{F}^{-1} \circ
    \mathcal{C}_{B, B} \circ \mathcal{F}(\boldsymbol{X}) \in \mathbb{R}^{B\!\times\!B}
    ,\ \  \boldsymbol{X} \in \mathbb{R}^{H\!\times\!W},
  \label{eq:low_freq_crop}
\end{equation}
where $\mathcal{F}$, $\mathcal{F}^{-1}$ and $\mathcal{C}_{B, B}$ denote 2D DFT, inverse 2D DFT and $B^2$ centre-cropping. The computational or the time cost for accomplishing Eq. (\ref{eq:low_freq_crop}) is negligible on GPUs.
  

\begin{figure}[t]
  \begin{center}
  \begin{minipage}{.405\linewidth}
    \centering
    \vskip -0.075in
    \includegraphics[width=\textwidth]{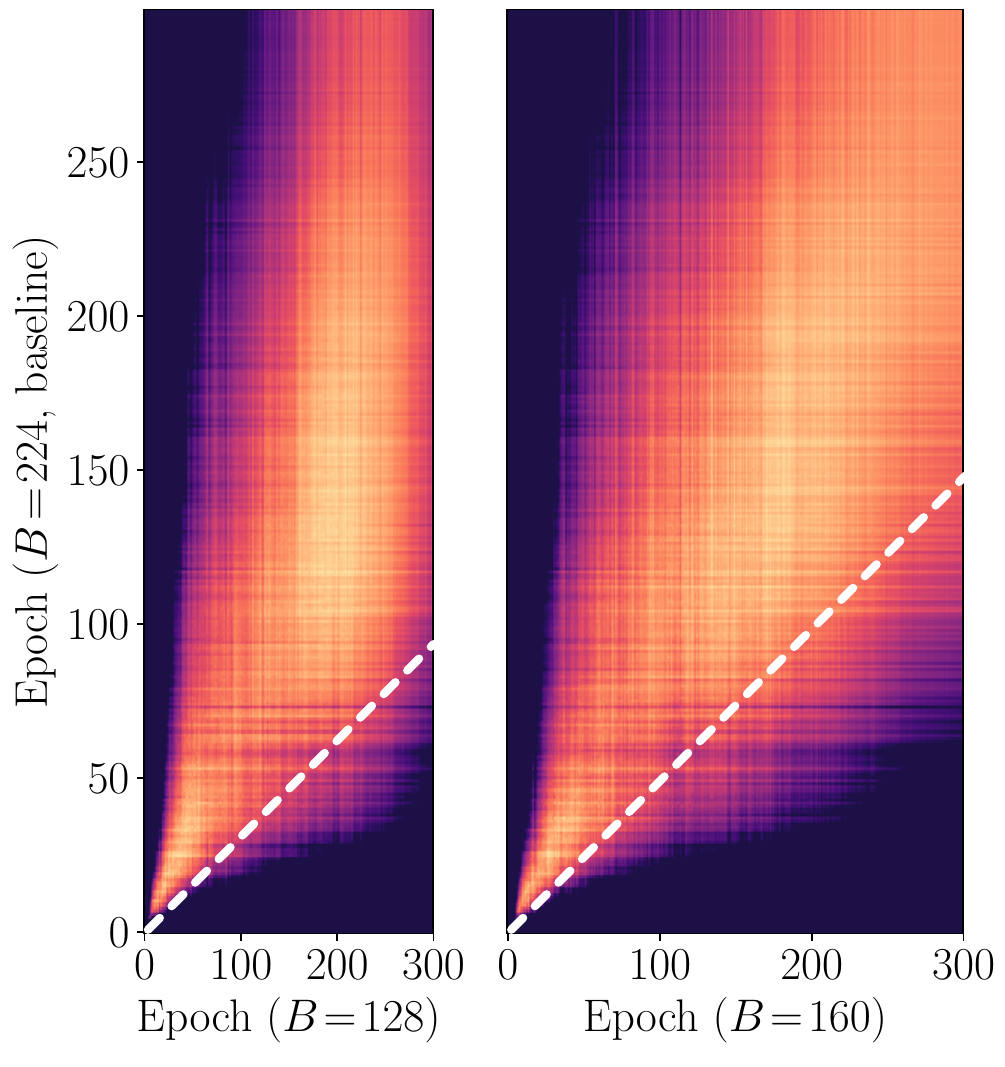}	  
  \end{minipage}
  \hspace{-1ex}
  \begin{minipage}{.593\linewidth} 
    \centering 
    \vskip 0.025in
    \captionsetup{font={footnotesize}}
    \caption{\textbf{CKA feature similarity} heatmaps \cite{cortes2012algorithms, kornblith2019similarity} between the DeiT-S \cite{touvron2021training} trained using the inputs with low-frequency cropping ($B\!\!=\!\!128, 160$) and the original inputs ($B\!\!=\!\!224$). The X and Y axes index training epochs (scaled according to the computational cost of training). 
    Here we feed the same original images into all the models (including the ones trained with $B\!\!=\!\!128, 160$) and take the features from the final layer. The $45^{\circ}$ lines are highlighted in white.
    \label{fig:cka}}
  \end{minipage} 
  \end{center}
  \vspace{-5.25ex}
\end{figure}

Notably, $\boldsymbol{X}_{\textnormal{c}}$ achieves a lossless extraction of lower-frequency components, while the higher-frequency parts are strictly eliminated. Hence, feeding $\boldsymbol{X}_{\textnormal{c}}$ into the model at earlier training stages can provide the vast majority of the useful information, such that the final accuracy will be minimally affected or even not affected. In contrast, importantly, due to the reduced input size of $\boldsymbol{X}_{\textnormal{c}}$, the computational cost for a model to process $\boldsymbol{X}_{\textnormal{c}}$ is able to be dramatically saved, yielding a considerably more efficient training process.

\begin{proposition}
  \label{prop:downsampling}
  Suppose that $\boldsymbol{X}_{\textnormal{c}}\!=\!\mathcal{F}^{-1} \circ \mathcal{C}_{B, B} \circ \mathcal{F}(\boldsymbol{X})$, and that $\boldsymbol{X}_{\textnormal{d}}\!=\!\mathcal{D}_{B, B}(\boldsymbol{X})$, where $B\!\times\!B$ down-sampling $\mathcal{D}_{B, B}(\cdot)$ is realized by a common interpolation algorithm ({e.g.}, nearest, bilinear or bicubic). 
  Then we have two properties.

  $\bm{\bullet}$\ a) $\boldsymbol{X}_{\textnormal{c}}$ is only determined by the lower-frequency spectrum of $\boldsymbol{X}$ (i.e., $\mathcal{C}_{B, B} \!\circ\! \mathcal{F}(\boldsymbol{X})$). In addition, the mapping to $\boldsymbol{X}_{\textnormal{c}}$  is reversible. We can always recover $\mathcal{C}_{B, B} \!\circ\! \mathcal{F}(\boldsymbol{X})$ from $\boldsymbol{X}_{\textnormal{c}}$.

  $\bm{\bullet}$\ b) $\boldsymbol{X}_{\textnormal{d}}$ has a non-zero dependency on the higher-frequency spectrum of $\boldsymbol{X}$ (i.e., the regions outside $\mathcal{C}_{B, B} \!\circ\! \mathcal{F}(\boldsymbol{X})$).

\end{proposition}
\noindent\textbf{\textit{Proof.}}
See: Appendix \textcolor{red}{C}. $\hfill\qedsymbol$
\vspace{1ex}

Our claims can be empirically supported by the results in Table \ref{tab:resolution_based_curriculum}, where we replace the low-pass filtering operation in Table \ref{tab:frequency_based_curriculum} with low-frequency cropping. Even such a straightforward implementation yields favorable results: the training cost can be saved by $\sim$20\% while a competitive final accuracy is preserved. This phenomenon can be interpreted via Figure \ref{fig:cka}: at intermediate training stages, the models trained using the inputs with low-frequency cropping can learn deep representations similar to the baseline with a significantly reduced cost, \emph{i.e.}, the bright parts are clearly above the white lines.


\begin{table}[t]
  \centering
  \vskip -0.05in
  \begin{footnotesize}
  \setlength{\tabcolsep}{0.75mm}{
  \vspace{5pt}
  \renewcommand\arraystretch{1.6}
  \resizebox{\linewidth}{!}{
    \begin{tabular}{cc|cc|ccccc}
       \multicolumn{4}{c}{Curricula (ep: epoch)} & \multicolumn{5}{c}{Final Top-1 Accuracy}   \\[-0.25ex]
       Weaker & {RandAug} & \ Low-frequency\  & Original & \multicolumn{5}{c}{($m$: magnitude of RandAug)}   \\[-0.65ex]
       RandAug & ($m\!=\!9$) & ($B\!=\!128$)  & ($B\!=\!224$) & $m\!=\!1$ & $m\!=\!3$ & $m\!=\!5$ & $m\!=\!7$ & $m\!=\!9$ \\
       \shline
       \multirow{2}{*}{\shortstack{1$^{\textnormal{st}}$ --\\[-0.25ex] 150$^{\textnormal{th}}$ ep}} & \multirow{2}{*}{\shortstack{151$^{\textnormal{th}}$ --\\[-0.25ex] 300$^{\textnormal{th}}$ ep}} & -- & \!\!\!\ \ \ \ \ 1$^{\textnormal{st}}$\  \!--\! 300$^{\textnormal{th}}$ ep & 80.4\% & 80.6\% 
       & \textbf{80.7\%} & 80.5\%  & 80.3\%  \\
       \cline{3-9}
       & & 1$^{\textnormal{st}}$ \!--\! 150$^{\textnormal{th}}$ ep & \!\!\!\ 151$^{\textnormal{th}}$ \!--\! 300$^{\textnormal{th}}$ ep & \textbf{80.2\%} & \textbf{80.2\%}  & \textbf{80.2\%} & 79.9\%  & 79.8\% \\
      \end{tabular}}}
      \vskip -0.05in 
      \captionsetup{font={footnotesize}}
      \captionof{table}{\textbf{Performance of the data-augmentation-based curricula} (DeiT-Small \cite{touvron2021training} on ImageNet-1K). We test reducing the magnitude of RandAug at 1$^{\textnormal{st}}$ \!-\! 150$^{\textnormal{th}}$ training epochs ($m\!=\!9$ refers to the baselines).}
      \label{tab:randaug_based_curriculum}
  \end{footnotesize}
  \vspace{-2ex}
\end{table}

\subsection{Easier-to-learn Patterns: Spatial Domain}
\label{sec:data_aug}

Apart from the frequency domain operations, extracting `easier-to-learn' patterns can also be attained through spatial domain transformations. For example, modern deep networks are typically trained with strong and delicate data augmentation techniques \cite{touvron2021training, liu2021swin, dong2021cswin, liu2022convnet, cubuk2019autoaugment, zhang2019adversarial, wang2021regularizing}. We argue that the augmented training data provides a combination of both the information from original samples and the information introduced by the augmentation operations. The original patterns may be `easier-to-learn' as they are drawn from real-world distributions. This assumption can be supported by the observation that data augmentation is mainly influential at the later stages of training \cite{tian2020improving}.

To this end, following our generalized formulation of curriculum learning in Section \ref{sec:GCL}, a curriculum may adopt a weaker-to-stronger data augmentation strategy during training. We investigate this idea by selecting RandAug \cite{cubuk2020randaugment} as a representative example, which incorporates a family of common spatial-wise data augmentation transformations (rotate, sharpness, shear, solarize, etc.). In Table \ref{tab:randaug_based_curriculum}, the magnitude of RandAug is varied in the first half training process. One can observe that this idea improves the accuracy, and the gains are compatible with low-frequency cropping.

Notably, the observation from Table \ref{tab:randaug_based_curriculum} is consistent with \cite{tan2021efficientnetv2, li2022automated}. We do not introduce weaker-to-stronger data augmentation for the first time. In contrast, our contributions here lie in demonstrating that `easier-to-learn' patterns can be identified and extracted through the lens of both frequency and spatial domains. This finding not only enhances the completeness of our generalized curriculum learning framework, but also offers novel insights to explain why weaker-to-stronger augmentation is effective. However, note that this does not mean that our technical innovations over \cite{tan2021efficientnetv2, li2022automated} are limited. Our contributions also incorporate many other important aspects (see: Section \ref{sec:related}).



\subsection{A Unified Training Curriculum}
\label{sec:EfficientTrain}

Finally, we integrate the techniques proposed in Sections \ref{sec:freq_inspired} and \ref{sec:data_aug}, and design a unified efficient training curriculum. In specific, we first set the magnitude $m$ of RandAug to be a linear function of the training epoch $t$: $m\!=\!(t/T)\!\times\!m_0$, with other data augmentation techniques unchanged. Despite the simplicity, this setting yields consistent and significant empirical improvements. Note that we adopt $m_0\!=\!9$ following the common practice \cite{touvron2021training, wang2021pyramid, liu2021swin, dong2021cswin, liu2022convnet}. 

Then we leverage a greedy-search algorithm to determine the schedule of $B$ during training (\emph{i.e.}, the bandwidth for low-frequency cropping). As shown in Algorithm \ref{alg:greedy_search}, we divide the training process into several stages and solve for a value of $B$ for each stage. The algorithm starts from the last stage, minimizing $B$ under the constraint of not degrading the performance compared to the baseline (trained with a fixed $B=224$). In our implementation, we only execute Algorithm \ref{alg:greedy_search} \emph{for a single time}. We obtain a schedule on top of Swin-Tiny \cite{liu2021swin} on ImageNet-1K under the standard 300-epoch training setting \cite{liu2021swin}, and directly adopt this schedule for other models or other training settings.

\begin{figure}[!h]
  \begin{center}
    \vskip -0.21in
    \resizebox{0.85\linewidth}{!}{
  \begin{minipage}{\linewidth}
      \begin{center}
              \begin{algorithm}[H]
                  \caption{The Greedy-search Algorithm.}
                  \label{alg:greedy_search}
              \begin{algorithmic}[1]
                  \STATE {\bfseries Input:} Number of training epochs $T$ and training stages $N$ (\emph{i.e.}, $T/N$ epochs for each stage).
                  \STATE {\bfseries Input:} Baseline accuracy $a_0$ (with 224$^2$ images).
                  \STATE {\bfseries To solve for:} The value of $B$ for $i^{\textnormal{th}}$ training stage: $\hat{B}_i$.
                  \STATE {\bfseries Initialize:}  $\hat{B}_1=\ldots=\hat{B}_N=224$
                  \FOR{$i=N-1$ {\bfseries to} $1$}
                  \STATE $\hat{B}_i = \mathop{\textnormal{minimize}}\limits_{B_1=\ldots=B_i=B,\ B_j=\hat{B}_j,\ i<j\leq N} B$, \\ s.t. $\textnormal{ValidationAccuracy}(B_1, \ldots, B_N) \geq a_0$
                  \ENDFOR
                  \STATE {\bfseries Output:} $\hat{B}_1, \ldots, \hat{B}_N$
              \end{algorithmic}
              \end{algorithm}
              \vskip -0.2in
      \end{center}
  \vspace{-5.5ex}
  \end{minipage}}
  \end{center}
\end{figure}

\begin{figure}[!h]
  \begin{center}
      \begin{minipage}{\linewidth}
          \centering
  \vskip -0.15in
  \begin{footnotesize}
  \setlength{\tabcolsep}{2mm}{
  \renewcommand\arraystretch{1.35}
  \resizebox{0.85\linewidth}{!}{
    \begin{tabular}{r|cc}
      \multicolumn{1}{c|}{Epochs} & Low-frequency Cropping & RandAug  \\
       \shline
       1$^{\textnormal{st}}$\ \ \  -- 180$^{\textnormal{th}}$ & $B=160$ &  \multirow{3}{*}{\shortstack{$m=0 \to 9$\\ Increase linearly.}}\\  
       181$^{\textnormal{th}}$ -- 240$^{\textnormal{th}}$  & $B=192$  &  \\
       241$^{\textnormal{th}}$ -- 300$^{\textnormal{th}}$  & $B=224$ &   \\
      \end{tabular}}}
      \vskip -0.075in 
      \captionsetup{font={footnotesize}}
      \captionof{table}{\textbf{The EfficientTrain curriculum} obtained from Algorithm \ref{alg:greedy_search}. We only execute Algorithm \ref{alg:greedy_search} once, and directly adopt this resulting curriculum for different models and training settings. The standard 300-epoch training pipeline \cite{liu2021swin} is considered when executing Algorithm \ref{alg:greedy_search}, and hence the curriculum presented here adopts this configuration. However, EfficientTrain can easily adapt to varying epochs and final input sizes utilizing simple linear scaling (see: Table \ref{tab:vary_res} \& Appendix \textcolor{red}{B}).}
      \label{tab:EfficientTrain}
  \end{footnotesize}
\end{minipage}
\end{center}
\vskip -0.075in
\end{figure}


Derived from the aforementioned procedure, our finally proposed learning curriculum is presented in Table \ref{tab:EfficientTrain}. We refer to it as \emph{EfficientTrain}. Notably, although being simple, EfficientTrain is well-generalizable and surprisingly effective. In the context of training or pre-training visual backbones on large-scale natural image databases (\emph{e.g.}, on ImageNet-1K/22K \cite{deng2009imagenet}), EfficientTrain can be directly applied to state-of-the-art deep networks \emph{without additional hyper-parameter tuning}, and improves their training efficiency significantly. The gains are consistent across various network architectures, different computational budgets for training, supervised/self-supervised learning algorithms, and varying amounts of training data.

\section{EfficientTrain++}

In this section, we propose an enhanced \emph{EfficientTrain++} approach. EfficientTrain++ improves the vanilla EfficientTrain from two aspects, \emph{i.e.}, 1) saving the non-trivial computational cost for executing Algorithm \ref{alg:greedy_search}, and 2) reducing the increased CPU-GPU input/output (I/O) cost caused by the low-frequency cropping operation. We address these issues by introducing a low-cost algorithm to determine the schedule of $B$ (Section \ref{sec:ET_plus_1}), as well as an efficient low-frequency down-sampling operation (Section \ref{sec:ET_plus_2}). 

In addition to these methodological innovations, we further present two simple but effective implementation techniques for EfficientTrain++ (Section \ref{sec:ET_plus_3}). Neither technique is an indispensable component of our method, but adopting them enables EfficientTrain++ to achieve a more significant practical training speedup, \emph{e.g.}, by 1) taking full advantage of more GPUs for large-scale parallel training and 2) reducing the data pre-processing loads for CPUs/memory.

\subsection{Computational-constrained Sequential Searching}
\label{sec:ET_plus_1}

The underlying logic behind Algorithm \ref{alg:greedy_search} is straightforward. A constrained optimization problem is considered. The constraint is not to degrade the final accuracy of the model, while the optimization objective is to minimize the training cost. We solve this problem in multiple steps (Algorithm \ref{alg:greedy_search}, lines 5-7), where we always try to reduce as much training cost as possible at each step.

\textbf{Efficiency of Algorithm \ref{alg:greedy_search}.}
However, Algorithm \ref{alg:greedy_search} suffers from a high computational cost. First, the final accuracy is considered as the constraint. Hence, given any possible combination of $\{B_1, \ldots, B_N\}$, we need to train the model to convergence, obtain the final validation accuracy, and verify whether $\{B_1, \ldots, B_N\}$ satisfies the constraint. Second, such verification always needs to train the model from scratch, \emph{i.e.}, since we solve for $B$ values from the last training stage to the first one, we need to alter $B$ for certain beginning stages of training whenever we execute line 6 of Algorithm \ref{alg:greedy_search} (\emph{i.e.}, changing $B_1, \ldots, B_i$). In general, assume that we have $N$ training stages and $M$ candidate values for $B$. Executing Algorithm \ref{alg:greedy_search} will train the model from scratch to convergence for $\sim\!(N\!+\!M\!-\!1)$ times, which may introduce non-trivial computational and time costs.
Although this drawback does not interfere with directly utilizing the resulting curriculum from Algorithm \ref{alg:greedy_search} as an off-the-shelf broadly-applicable efficient training technique, the limited efficiency of Algorithm \ref{alg:greedy_search} may hinder future works, \emph{e.g.}, to design a tailored curriculum for new generalized curriculum learning approaches or for other related methods.



\textbf{A more efficient algorithm to solve for the curriculum.}
Driven by the aforementioned analysis, we propose a new algorithm to determine the schedule of $B$ in the curriculum, as shown in Algorithm \ref{alg:greedy_search_v2}. This new approach is not only more efficient than Algorithm \ref{alg:greedy_search}, but also outperforms Algorithm \ref{alg:greedy_search} in terms of its resulting curriculum. To be specific, Algorithm \ref{alg:greedy_search_v2} introduces two major innovations:
\begin{itemize}
  \item \textbf{New formulation: computational-constrained searching.} We consider a different constrained optimization problem from Algorithm \ref{alg:greedy_search}. The constraint is to save the training cost by a pre-defined ratio compared with the baseline (specified by $\beta$ in Algorithm \ref{alg:greedy_search_v2}), while the optimization objective is to maximize the final accuracy. Similar to Algorithm \ref{alg:greedy_search}, we solve this problem by dividing the training process into $N$ stages, and solving for a $B$ value for each stage. However, here we allocate $1/N$ of the total training cost to each stage, and then simultaneously vary $B$ and the number of training steps at each stage with this training budget fixed. That is to say, given a pre-defined batch size, when $B$ is relatively small, the cost for a model to process a mini-batch of training data is small, and we update the model for more times using more mini-batches of data, yet the overall amount of computation is unchanged. In contrast, when $B$ is large at a training stage, the number of model updating times needs to be relatively less since the training budget is fixed. Note that the learning rate schedule for each training stage always remains unchanged.
  \item \textbf{Performing sequential searching.} We solve for the $B$ values from the first training stage to the last one, where the solving of $B$ at any stage is based on the checkpoint from the previous stage. Specifically, given a training stage, we execute it from the previous checkpoint with several candidate $B$ values respectively, and fine-tune the obtained models with the original images ($B\!=\!224$) for a small number of epochs ($T_{\textnormal{ft}}$ in Algorithm \ref{alg:greedy_search_v2}). We select the best $B$ by comparing the validation accuracy of these fine-tuned models, which we find is an effective proxy to reflect the influence of different $B$ on the final accuracy. 
\end{itemize}

In our implementation, Algorithm \ref{alg:greedy_search_v2} is executed on top of Swin-Tiny \cite{liu2021swin} on ImageNet-1K, where we set $\beta\!=\!2/3$, $T_0\!=\!300$ and $T_{\textnormal{ft}}\!=\!10$. The resulting curriculum is presented in Table \ref{tab:EfficientTrain_plus}, which is named as \emph{EfficientTrain++}. We directly adopt the curriculum in Table \ref{tab:EfficientTrain_plus} for other models and other training settings without executing Algorithm \ref{alg:greedy_search_v2} again.

\begin{figure}[!h]
  \begin{center}
    \vskip -0.225in
    \resizebox{0.95\linewidth}{!}{
  \begin{minipage}{1.15\linewidth}
      \begin{center}
              \begin{algorithm}[H]
                  \caption{The Computational-constrained Sequential Searching Algorithm.}
                  \label{alg:greedy_search_v2}
              \begin{algorithmic}[1]
                  \STATE {\bfseries Input:} Training epochs $T\!=\!\beta T_0$ ($T_0$: baseline training epochs; $0\!<\!\beta\!<\!1$: pre-defined ratio of saving training cost). Number of training stages $N$. Proxy fine-tuning epochs $T_{\textnormal{ft}}$.
                    \STATE {\bfseries Input:} Learning rate schedule $\bm{\alpha}^{\textnormal{lr}}$, where $\bm{\alpha}^{\textnormal{lr}}_{t_1:t_2} (0 \!\leq\! t_1 \!<\! t_2 \!\leq\! 1)$ denotes a continuous segment.
                    \STATE {\bfseries Input:} Random initialized model parameterized by $\bm{\Theta}$. 
                  \STATE {\bfseries To solve for:} The value of $B$ for $i^{\textnormal{th}}$ training stage: $\hat{B}_i$.
                  \FOR{$i=1$ {\bfseries to} $N-1$}
                  \STATE $\hat{B}_i = \mathop{\textnormal{argmax}}\limits_{B}
                  \ \ \textnormal{ValidationAccuracy}(\bm{\widehat{\Theta}}^{i}_{B})$, 
                  \\ s.t. \ $\bm{\widehat{\Theta}}^{i}_{B}\!=\!\textnormal{Fine-tune}(
                    \bm{\Theta}^i_{B}\ |\  {
                      224^2,\ \!T_{\textnormal{ft}}\ \!\textnormal{epochs},\ \!\bm{\alpha}^{\textnormal{lr}}_{i/N:(i/N + T_{\textnormal{ft}}/T)}
                      }
                    )$,
                  \\ \ \ \ \ \ \ \ \!$\bm{\Theta}^i_{B}\!=\!\textnormal{Train}(
                    \bm{\Theta}\ |\  {
                    B^2,\ \!
                    \frac{T}{N}\!\cdot\!\frac{\textnormal{FLOPs}(224^2)}{\textnormal{FLOPs}(B^2)}\ \!\textnormal{epochs},\ \!
                    \bm{\alpha}^{\textnormal{lr}}_{(i-1)/N:i/N}
                    }
                  )$
                  \STATE $\bm{\Theta}\leftarrow\bm{\Theta}^i_{\hat{B}_i}$
                  \ENDFOR
                  \STATE {\bfseries Output:} $\hat{B}_1, \ldots, \hat{B}_N$
              \end{algorithmic}
              \end{algorithm}
              \vskip -0.2in
      \end{center}
  \vspace{-5.5ex}
  \end{minipage}}
  \end{center}
\end{figure}

Importantly, different from Table \ref{tab:EfficientTrain}, \emph{EfficientTrain++} is based on the computational-constrained formulation introduced by Algorithm \ref{alg:greedy_search_v2}. In other words, the values of $B$, $m$, and the learning rate are determined conditioned on how much of the total computational budget for training has been consumed, \emph{i.e.}, their schedules are defined on top of the computational cost. Built upon this characteristic, our EfficientTrain++ curriculum can flexibly adapt to a varying number of total training budgets. Notably, to measure the total amount of training cost of our method, we convert the required computation for training into the number of the standard training epochs in baselines (\emph{e.g.}, with the original $224^2$ inputs), and directly report equivalent epoch numbers. This setup is adopted for the ease of understanding.




\begin{figure}[!t]
  \begin{center}
      \begin{minipage}{\linewidth}
          \centering
  \begin{footnotesize}
  \setlength{\tabcolsep}{3mm}{
  \renewcommand\arraystretch{1.35}
  \resizebox{0.9\linewidth}{!}{
    \begin{tabular}{c|cc}
      \multicolumn{1}{c|}{Computational Budget} & Low-frequency & \multirow{2}{*}{RandAug} \\[-0.6ex]
      \multicolumn{1}{c|}{for Training} & Cropping &   \\
       \shline
       0\% -- 20\% & $B=96$ \hspace{0.25em} &  \multirow{3}{*}{\shortstack{$m=0 \to 9$\\ Increase linearly.}}\\  
       \hspace{-0.5em}20\% -- 60\%  & $B=160$  &  \\
       60\% -- 100\%  & $B=224$ &   \\
      \end{tabular}}}
      \vskip -0.06in 
      \captionsetup{font={footnotesize}}
      \captionof{table}{\textbf{The EfficientTrain++ curriculum}. Notably, the computation-constrained formulation of Algorithm \ref{alg:greedy_search_v2} is adopted here, \emph{i.e.}, the values of $B$, $m$, and the learning rate are configured conditioned on how much of the total training cost has been consumed. 
      Besides, to measure the total length of the training process, we convert the required computation for training into the number of the standard training epochs in baselines (\emph{e.g.}, with the original $224^2$ inputs), and report this equivalent number of epochs for the ease of understanding. 
      }
      \label{tab:EfficientTrain_plus}
  \end{footnotesize}
\end{minipage}
\end{center}
\vskip -0.225in
\end{figure}

\subsection{Efficient Low-frequency Down-sampling}
\label{sec:ET_plus_2}

As aforementioned, the low-frequency cropping operation in Figure \ref{fig:freq_crop} is able to extract all the lower-frequency components while eliminating the rest higher-frequency information. However, it suffers from a high CPU-GPU input/output (I/O) cost. This issue is caused by the 2D DFT and inverse 2D DFT in Figure \ref{fig:freq_crop}, which are unaffordable for CPUs and have to be performed on GPUs. Hence, we need to transfer the original images (\emph{e.g.}, 224$^2$) from CPUs to GPUs. Consequently, the demand for CPU-to-GPU I/O throughput will increase by the same ratio as the training speedup, \emph{e.g.}, by $>\!5\times$ with $B\!=\!96$. We empirically find that this high I/O cost is an important bottleneck that inhibits the efficient implementation of our method.

\textbf{Low-pass filtering + image down-sampling.}
Inspired by this issue, we propose an efficient approximation of low-frequency cropping. Our method is derived by improving the image down-sampling operation. As demonstrated by Proposition \ref{prop:downsampling}, the drawback of down-sampling is that it cannot strictly filter out all the higher-frequency information. To address this problem, we propose a two-step procedure. First, we perform low-pass filtering on the original images with a $B\!\times\!B$ square filter (without changing the image size). Second, we perform $B\!\times\!B$ down-sampling. Since low-pass filtering extracts exactly the same information as the $B\!\times\!B$ low-frequency cropping, down-sampling will no longer introduce the undesirable higher-frequency components, and thus the final accuracy of the model will not be affected. Moreover, $B\!\times\!B$ low-pass filtering is mathematically equivalent to 2D sinc convolution (see: Appendix \textcolor{red}{D}), which can be efficiently implemented on CPUs with the Lanczos algorithm \cite{turkowski1990filters, burger2010principles}. As a result, our two-step procedure can be accomplished using CPUs, while only the resulting $B\!\times\!B$ model inputs need to be transferred to GPUs. Therefore, the CPU-GPU I/O cost is effectively reduced.

\begin{table*}[!t]
  \centering
  \begin{footnotesize}
  \setlength{\tabcolsep}{3mm}{
  \renewcommand\arraystretch{1.375}
  \resizebox{0.925\linewidth}{!}{
  \begin{tabular}{clcccccccc}
  \multicolumn{2}{c}{\multirow{2.25}{*}{Models}} & {\!\!\!Input Size\!\!} & \multirow{2.25}{*}{\!\!\#Param.\!\!}  & \multirow{2.25}{*}{\!\!\#FLOPs\!\!}  & \multicolumn{2}{c}{Top-1 Accuracy\ \ \ } & \multicolumn{2}{c}{Training Speedup} & \!\!\!\!\!\!\!\!\! \\[-0.2ex]
  && \!\!\!(inference)\!\!&& & Baseline\ \ \   & \!\!\!\!\baseline{}\textbf{EfficientTrain++}\!\!\!\! & Computation & \baseline{}\textbf{\ Wall-time\ } \\
  \shline
  \multirow{4}{*}{\textit{ConvNets}} & ResNet-50 \cite{He_2016_CVPR} 
  & 224$^2$\ \  & 26M & 4.1G & 78.8\%\ \ \   & \baseline{}\textbf{79.6\%}  & $1.49\times$ & \baseline{}$\bm{1.45\times}$  \\
  \hhline{|~---------|}
  & ConvNeXt-Tiny \cite{liu2022convnet} & 224$^2$\ \  & 29M & 4.5G &  82.1\%\ \ \    & \baseline{}\textbf{82.2\%} & $1.49\times$ & \baseline{}$\bm{1.49\times}$  \\
  & ConvNeXt-Small \cite{liu2022convnet}\!\!\!\!\!\! & 224$^2$\ \  & 50M & 8.7G  &  83.1\%\ \ \  & \baseline{}\textbf{83.2\%}  & $1.49\times$ & \baseline{}$\bm{1.52\times}$  \\
  & ConvNeXt-Base \cite{liu2022convnet} & 224$^2$\ \  & 89M & 15.4G  &  {83.8\%}\ \ \   & \baseline{}\textbf{83.8\%}  & $1.49\times$ & \baseline{}$\bm{1.49\times}$  \\
   \hline
   \multirow{2}{*}{\shortstack{\textit{Isotropic} \\[-0.4ex] \textit{ViTs}}} 
   & DeiT-Tiny \cite{touvron2021training} & 224$^2$\ \  & 5M &  1.3G &  72.5\%\ \ \    & \baseline{}\textbf{73.7\%}  & $1.56\times$ & \baseline{}$\bm{1.64\times}$   \\
   & DeiT-Small \cite{touvron2021training} & 224$^2$\ \  & 22M & 4.6G  &  80.3\%\ \ \    & \baseline{}\textbf{81.0\%} & $1.52\times$ & \baseline{}$\bm{1.60\times}$  \\
   \hline
   \multirow{10}{*}{\shortstack{\textit{Multi-stage} \\[-0.4ex] \textit{ViTs}}} 
   & PVT-Tiny \cite{wang2021pyramid} & 224$^2$\ \  & 13M &  1.9G   &  75.5\%\ \ \   & \baseline{}\textbf{75.9\%}  & $1.51\times$ & \baseline{}$\bm{1.51\times}$  \\
   & PVT-Small \cite{wang2021pyramid} & 224$^2$\ \  & 25M & 3.8G   &  79.9\%\ \ \   & \baseline{}\textbf{80.4\%}  & $1.51\times$ & \baseline{}$\bm{1.49\times}$  \\
   & PVT-Medium \cite{wang2021pyramid} & 224$^2$\ \  & 44M & 6.7G  &  81.8\%\ \ \    & \baseline{}\textbf{81.9\%}  & $1.51\times$ & \baseline{}$\bm{1.49\times}$  \\
   & PVT-Large \cite{wang2021pyramid} & 224$^2$\ \  & 61M & 9.8G  &  82.3\%\ \ \   & \baseline{}\textbf{82.4\%}  & $1.51\times$ & \baseline{}$\bm{1.48\times}$  \\
  \hhline{|~---------|}
   & Swin-Tiny \cite{liu2021swin} & 224$^2$\ \  & 28M & 4.5G   &  81.3\%\ \ \   & \baseline{}\textbf{81.6\%}  & $1.51\times$ & \baseline{}$\bm{1.49\times}$  \\
   & Swin-Small \cite{liu2021swin} & 224$^2$\ \  & 50M & 8.7G   &  83.1\%\ \ \    & \baseline{}\textbf{83.2\%}  & $1.51\times$ & \baseline{}$\bm{1.51\times}$   \\
   & Swin-Base \cite{liu2021swin} & 224$^2$\ \  & 88M & 15.4G   &  83.4\%\ \ \    & \baseline{}\textbf{83.6\%}  & $1.50\times$ & \baseline{}$\bm{1.47\times}$  \\
  \hhline{|~---------|}
   & CSWin-Tiny \cite{dong2021cswin} & 224$^2$\ \  & 23M & 4.3G   &  82.7\%\ \ \    & \baseline{}\textbf{82.9\%} & $1.52\times$ & \baseline{}$\bm{1.50\times}$  \\
   & CSWin-Small \cite{dong2021cswin} & 224$^2$\ \  & 35M & 6.9G  &  83.4\%\ \ \     & \baseline{}\textbf{83.6\%} & $1.52\times$ & \baseline{}$\bm{1.52\times}$  \\
   & CSWin-Base \cite{dong2021cswin} & 224$^2$\ \  & 78M & 15.0G  &   84.3\%\ \ \    & \baseline{}\textbf{84.3\%} & $1.51\times$ & \baseline{}$\bm{1.52\times}$  \\
  \hline
  \multirow{3}{*}{\shortstack{\textit{ConvNet} \\[-0.2ex] \textit{ + ViT}}} 
  & CAFormer-S18 \cite{yu2022metaformer} & 224$^2$\ \  & 26M & 4.1G  & \textbf{83.5\%}\ \ \  & \baseline{}83.4\%  & $1.52\times$ & \baseline{}$\bm{1.59\times}$  \\
  & CAFormer-S36 \cite{yu2022metaformer} & 224$^2$\ \  & 39M & 8.0G  & 84.3\%\ \ \    & \baseline{}\textbf{84.3\%} & $1.52\times$ & \baseline{}$\bm{1.59\times}$  \\
  & CAFormer-M36 \cite{yu2022metaformer} & 224$^2$\ \  & 56M & 13.2G & 84.9\%\ \ \    & \baseline{}\textbf{85.0\%} & $1.51\times$  & \baseline{}$\bm{1.58\times}$ \\
  \end{tabular}}}
  \end{footnotesize}
  \vskip -0.085in
  \captionsetup{font={footnotesize}}
  \captionof{table}{\textbf{Results on ImageNet-1K (IN-1K)}. We train the models with or without EfficientTrain++ on the IN-1K training set, and report the accuracy on the IN-1K validation set. The number of equivalent training epochs for EfficientTrain++ is set to 200, which is sufficient to achieve a competitive or better performance compared with the baselines. The wall-time training speedup is benchmarked on NVIDIA 3090 GPUs. \label{tab:img_1k_main_result}}
  \vskip -0.15in
\end{table*}

\subsection{Implementation Techniques for EfficientTrain++}
\label{sec:ET_plus_3}

\textbf{Facilitating large-scale parallel training: early large batch.}
In Table \ref{tab:EfficientTrain_plus}, the batch size is fixed during the whole training process. Nevertheless, our method always starts the training with a small $B$ (\emph{e.g.}, $B\!=\!96$), where the computational cost for each mini-batch is dramatically reduced compared to a larger $B$ (\emph{e.g.}, $B\!=\!224$). This property is harmless on typical settings of the training hardware (\emph{e.g.}, training a model on $\leq\!16$ GPUs). However, if we consider an increasing number of GPUs (\emph{e.g.}, 32/64 GPUs or more), the smaller $B$ at earlier training stages usually yields a bottleneck that inhibits the efficient implementation of EfficientTrain++.

This issue can be addressed by simultaneously increasing the batch size and the learning rate when $B$ is small. For one thing, such a modification has been shown to be an effective approximation of the learning process with the original smaller batch size \cite{goyal2017accurate, you2019large}, and hence the final performance of the model will not be degraded (see: Table \ref{tab:results_early_large_bs}). For another, the per-batch training cost for small $B$ is increased, such that the scaling efficiency of EfficientTrain++ with respect to the GPU number is significantly improved. In our implementation, we use an upper-bounded square root learning rate scaling rule:
\begin{equation}
  \textnormal{LR}^{\textnormal{max}}_{B} = \textnormal{min}\!\left(
    \textnormal{LR}^{\textnormal{max}}_{224}\!\times\!\sqrt{\frac{\textnormal{BS}_{B}}{\textnormal{BS}_{224}}}, \quad
    \overline{\textnormal{LR}^{\textnormal{max}}}
  \right),
\end{equation}
where $\textnormal{LR}^{\textnormal{max}}_{B}$ denotes the maximum value of the learning rate schedule (in this paper, we adopt the cosine annealing schedule with a linear warm-up \cite{liu2021swin, liu2022convnet}) corresponding to $B$, and $\textnormal{BS}_{B}$ denotes the batch size corresponding to $B$. We set $\overline{\textnormal{LR}^{\textnormal{max}}}$ to be the upper-bound of $\textnormal{LR}^{\textnormal{max}}_{B}$, since an excessively large learning rate usually leads to an unstable training process.

\textbf{Reducing data pre-processing loads: replay buffer.}
In our method, the computational cost for learning from each training sample is quadratically saved when $B$ is small. As a matter of fact, this reduction in learning cost comes with the demand for a higher data pre-processing throughput, \emph{i.e.}, the speed of preparing training inputs needs to keep up with the model training speed on GPUs. Therefore, EfficientTrain++ may increase the loads for the hardware like memory or CPUs, compared with the baseline, which may affect the practical training efficiency of our method. To alleviate this problem, we propose to leverage a replay buffer. Once a newly pre-processed mini-batch of data is produced, we put it into the buffer, while if a pre-defined maximum size is reached, we remove the oldest data. We observe that this maximum size can typically be set to reasonably large on common hardware. During training, to obtain the model inputs, we alternately execute the following two steps: 1) sampling mini-batches from the replay buffer for $n_{\textnormal{buffer}}$ times, and 2) producing a new mini-batch of training data, feeding it into the model for training, and updating the replay buffer with it. With this technique, the data pre-processing loads are reduced by $n_{\textnormal{buffer}}$+1 times, while the generalization performance of the model can be effectively preserved (see: Table \ref{tab:replay_buffer}). 
Notably, only the number of pre-processed mini-batches is reduced. Both the total number of training iterations and the computational cost for training models remain unchanged.







\section{Experiments}
\label{sec:experiment}




\textbf{Overview.}
Section \ref{sec:supervised_learning} evaluates our proposed method in standard supervised learning scenarios. Our experiments incorporate training a variety of visual backbones (\emph{e.g.}, ConvNets, isotropic/multi-stage ViTs, and ConvNet-ViT hybrid models) under diverse training settings (\emph{e.g.}, varying training budgets, different test input sizes, and various amounts of training data). We also present a comprehensive comparison of our method and state-of-the-art efficient training methods. In section \ref{sec:self_supervised_learning}, we demonstrate that our method can be conveniently applied to self-supervised learning (\emph{e.g.}, MAE) and yields notable improvements in training efficiency. In Section \ref{sec:transfer_learning}, we study the transferability of the models pre-trained using our method to downstream tasks, \emph{e.g.}, image classification, object detection, and dense prediction. In Section \ref{sec:discussion}, thorough ablation studies and analytical results are provided for a better understanding.

 \textbf{Datasets.}  
 Our main experiments are based on the large-scale ImageNet-1K/22K \cite{deng2009imagenet} datasets, which consist of $\sim$1.28M/$\sim$14.2M images in 1K/$\sim$22K classes. We also verify the transferability of our method on ADE20K \cite{zhou2019semantic}, MS COCO \cite{lin2014microsoft}, CIFAR \cite{krizhevsky2009learning}, Flowers-102 \cite{nilsback2008automated}, Stanford Dogs \cite{khosla2011novel}, NABirds \cite{van2015building}, and Food-101 \cite{bossard2014food}.

\textbf{Models \& Setups.}  
 A wide variety of visual backbone models are considered in our experiments, including ResNet \cite{He_2016_CVPR}, ConvNeXt \cite{liu2022convnet}, DeiT \cite{touvron2021training}, PVT \cite{wang2021pyramid}, Swin \cite{liu2021swin} Transformers, CSWin \cite{dong2021cswin} Transformers, and CAFormer \cite{yu2022metaformer}. We adopt the 300-epoch training pipeline in \cite{liu2021swin, liu2022convnet} as our baseline, on top of which EfficientTrain and EfficientTrain++ only modify the terms mentioned in Tables \ref{tab:EfficientTrain} and \ref{tab:EfficientTrain_plus}, respectively. In particular, the default version of EfficientTrain++ improves EfficientTrain by leveraging the advancements proposed in Sections \ref{sec:ET_plus_1} and \ref{sec:ET_plus_2}. The two implementation techniques proposed in Section \ref{sec:ET_plus_3} are treated as optional components, and their effectiveness is separately investigated in Section \ref{sec:validate_implementation_technique}. Moreover, to measure the total length of the training process, for the baselines and the vanilla EfficientTrain, we directly report the real training epochs. For EfficientTrain++, as we adopt the new computational-constrained formulation, we convert the total amount of computation consumed by training models into the number of the standard training epochs in baselines, and report the equivalent epoch numbers for the ease of understanding (see: Section \ref{sec:ET_plus_1}). For example, the computational cost of $M$-epoch training for EfficientTrain++ is equal to the $M$-epoch trained baselines. Besides, unless otherwise specified, we report the results of our implementation for both our method and the baselines. More implementation details can be found in Appendix \textcolor{red}{A}.


 

\begin{figure*}[!t]
  \begin{center}
  \centerline{
    \includegraphics[width=0.755\linewidth]{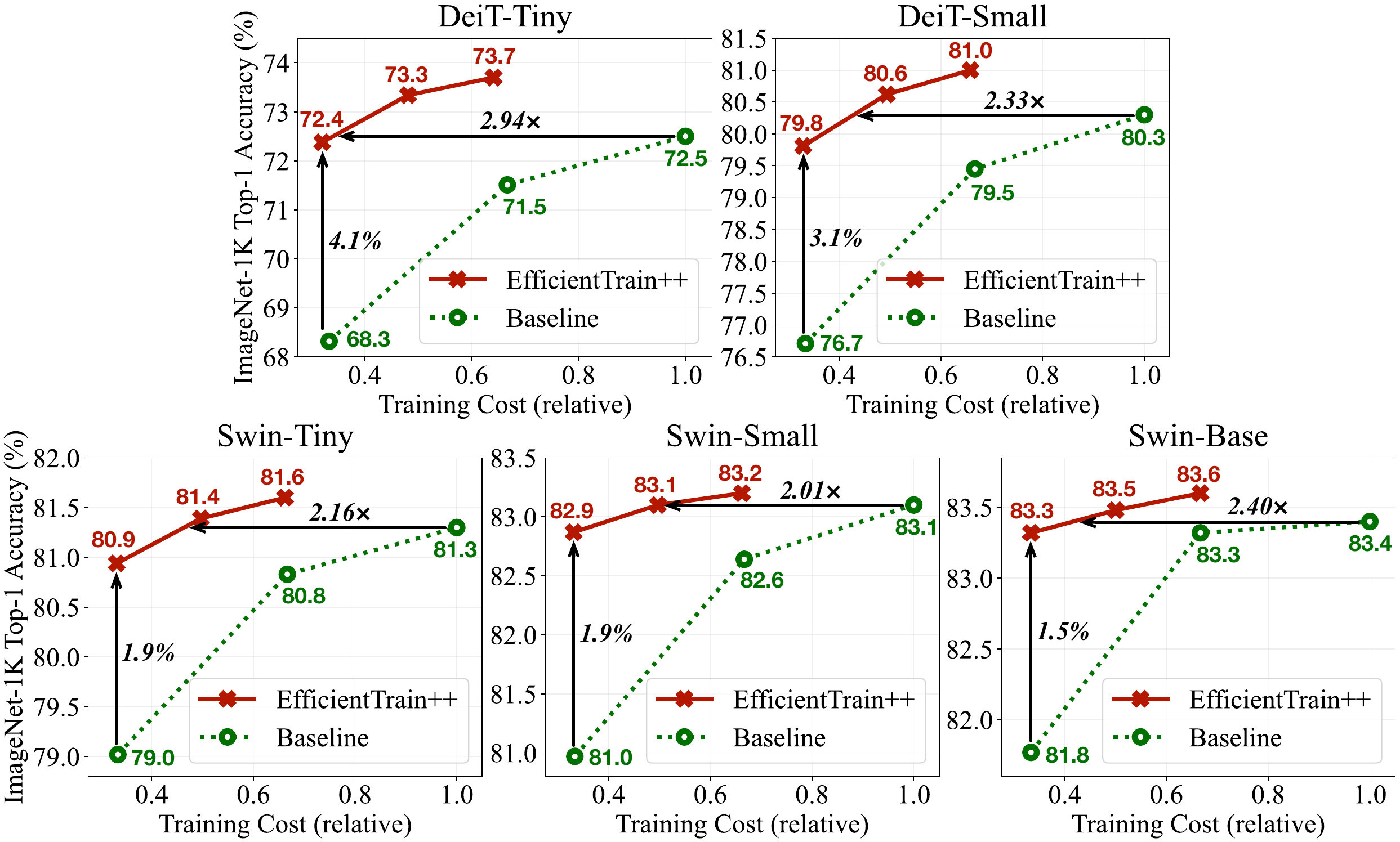}
    }
  \vskip -0.1in
  \captionsetup{font={footnotesize}}
  \caption{\textbf{Top-1 accuracy v.s. total training cost on ImageNet-1K}. For the baselines and EfficientTrain++, we vary the total number of training epochs and equivalent training epochs within $[100, 300]$ and $[100, 200]$, respectively. \label{fig:acc_vs_cost}
  }
  \end{center}
  \vskip -0.385in
\end{figure*}

 \subsection{Supervised Learning}
 \label{sec:supervised_learning}

\subsubsection{Main Results on ImageNet-1K}

 \textbf{Training various visual backbones on ImageNet-1K.}
 Table \ref{tab:img_1k_main_result} presents the results of applying EfficientTrain++ to train representative deep networks on ImageNet-1K, which is one of the most widely-used settings for evaluating deep learning algorithms. It is clear that our method achieves a competitive or better validation accuracy compared to the baselines (\emph{e.g.}, 85.0\% v.s. 84.9\% on CAFormer-M36), while saving the computational cost for training by $1.5\!-\!1.6\times$. Moreover, an important observation is that the gains of EfficientTrain++ are consistent across different types of deep networks, which demonstrates the strong generalizability of our method. In addition, Table \ref{tab:img_1k_main_result} also reports the wall-time training speedup on GPU devices. It can be observed that the practical efficiency of EfficientTrain++ is inline with the theoretical results. Visualization of the curves of accuracy during training v.s. time can be found in Appendix \textcolor{red}{B}.

\textbf{EfficientTrain v.s. EfficientTrain++.}
In Table \ref{tab:ET_vs_ETplus}, we present a comprehensive comparison of EfficientTrain, EfficientTrain++, and the baselines. Several observations can be obtained from the results. First, both the two versions of our methods effectively improve the training efficiency. The vanilla EfficientTrain is able to reduce the training cost of a variety of deep networks by $1.5\times$ compared to the baselines, while effectively enhancing the accuracy at the same time. Second, the EfficientTrain++ curriculum further outperforms EfficientTrain consistently. For example, on ImageNet-1K, EfficientTrain++ boosts the accuracy by 0.6\% for DeiT-small (81.0\% v.s. 80.4\%). In the context of pre-training larger models like CSWin-Base/Large, EfficientTrain++ achieves a significantly higher training speedup than EfficientTrain ($2.0\times$ v.s. $1.5\times$), and yields a better generalization performance. Third, one can observe that Algorithm \ref{alg:greedy_search_v2} utilized in EfficientTrain++ reduces the solving cost for obtaining the curriculum by $2.56\times$ compared to Algorithm \ref{alg:greedy_search} used for acquiring EfficientTrain. Overall, these aforementioned observations indicate that EfficientTrain++ improves EfficientTrain holistically, as we expect.

\begin{table}[!t]
  \vskip -0.075in
  \centering
  \begin{footnotesize}
  \setlength{\tabcolsep}{1mm}{
  \renewcommand\arraystretch{1.375}
  \resizebox{0.98\linewidth}{!}{
  \begin{tabular}{ccccccc}
  & & & \!Baseline\! & \!\textbf{EfficientTrain}\! & \ \ \ \ \baseline{}\textbf{ET++}\ \ \ \ & \!\!   \\[0.2ex]
  \shline
  \multicolumn{3}{c}{Cost for Solving for the} &  & &  \baseline{}  \\[-0.5ex]
  \multicolumn{3}{c}{\!\!\!\!\!\!Curriculum (in GPU-hours)\!\!\!\!\!\!} & \multirow{-1.75}{*}{--}  & \multirow{-1.75}{*}{1088.1} & \baseline{}\multirow{-1.6}{*}{\shortstack{\textbf{425.8}\\\textcolor{blue}{\scriptsize{\textbf{($\bm{\downarrow\!2.56\times}$)}}}}} \\[0.4ex]
  \hline
  \multirow{9}{*}{\!\!\textit{\shortstack{Results on\\ImageNet-1K}}} &
  \multicolumn{2}{c}{\multirow{2}{*}{\shortstack{Wall-time\\Training Speedup}}} &  \multirow{2}{*}{${1.0\times}$} & \multirow{2}{*}{${\sim\!1.5\times}$}  &  \baseline{}  \\
  & & & & & \baseline{}\multirow{-2}{*}{$\bm{\sim\!1.5\times}$}  \\
  \hhline{|~------|} 
  & \multirow{7}{*}{\shortstack{Top-1\\[-0.5ex]Acc.}\!\!} & \multicolumn{1}{l}{ResNet-50 \cite{He_2016_CVPR}}  & 78.8\%  & 79.4\% & \baseline{}\textbf{79.6\%}  \\
  & & \multicolumn{1}{l}{DeiT-Tiny \cite{touvron2021training}} &  72.5\%  & 73.3\% & \baseline{}\textbf{73.7\%}  \\
  & & \multicolumn{1}{l}{DeiT-Small \cite{touvron2021training}}  &  80.3\%  & 80.4\% & \baseline{}\textbf{81.0\%} \\
  & & \multicolumn{1}{l}{Swin-Tiny \cite{liu2021swin}}   &  81.3\%  & 81.4\% & \baseline{}\textbf{81.6\%} \\
  & & \multicolumn{1}{l}{Swin-Small \cite{liu2021swin}}   &  83.1\%  & 83.2\% & \baseline{}\textbf{83.2\%}  \\
  & & \multicolumn{1}{l}{CSWin-Tiny \cite{dong2021cswin}}  &  82.7\%  & 82.8\% & \baseline{}\textbf{82.9\%} \\
  & & \multicolumn{1}{l}{CSWin-Small \cite{dong2021cswin}}  &  83.4\%  & 83.6\% & \baseline{}\textbf{83.6\%}  \\
  \hline
  \multirow{4}{*}{\!\!\textit{\shortstack{Results with\\ImageNet-22K\\[-0.75ex]Pre-training$^\dagger$}}} &
  \multicolumn{2}{c}{\multirow{2}{*}{\shortstack{Wall-time\\Pre-training Speedup}}} &  \multirow{2}{*}{${1.0\times}$} & \multirow{2}{*}{${\sim\!1.5\times}$}  &  \baseline{}  \\
  & & & & & \baseline{}\multirow{-2}{*}{$\bm{\sim\!2.0\times}$}  \\
  \hhline{|~------|} 
  & \multirow{2.2}{*}{\shortstack{Top-1\\[-0.5ex]Acc.}\!\!} & \multicolumn{1}{l}{CSWin-Base \cite{dong2021cswin}}  &  86.0\%  & 86.2\% & \baseline{}\textbf{86.3\%}  \\
  & & \multicolumn{1}{l}{CSWin-Large \cite{dong2021cswin}}  &  86.8\%  & 86.9\% & \baseline{}\textbf{87.0\%}  \\
  \end{tabular}}}
  \end{footnotesize}
  \captionsetup{font={footnotesize}}
  \vskip -0.1in
  \caption{\textbf{Comparisons of EfficientTrain and EfficientTrain++ (ET++)}. $\dagger$: these results are presented here for a comprehensive comparison (for the detailed results on ImageNet-22K, please refer to Table \ref{tab:img_22K_main_result}).
  \label{tab:ET_vs_ETplus}
  }
  \vskip -0.075in
\end{table}

\begin{table}[!t]
  \centering
  \begin{footnotesize}
  \setlength{\tabcolsep}{1.75mm}{
    \renewcommand\arraystretch{1.325}
    \resizebox{0.975\linewidth}{!}{
  \begin{tabular}{lcccccc}
  \multicolumn{1}{c}{\multirow{2}{*}{Models}} & Input Size & \multicolumn{3}{c}{Top-1 Accuracy} & \!\!\!  \\[-0.15ex]
  & (inference) & Baseline &  \textbf{EfficientTrain} & \baseline{}\textbf{ET++} \\
  \shline
  DeiT-Tiny \cite{touvron2021training} & 224$^2$ & 72.5\% & {74.3\%}\textcolor{blue}{\ \scriptsize{\textbf{(+1.8)}}}  & \baseline{}\textbf{74.4\%}\textcolor{blue}{\ \scriptsize{\textbf{(+1.9)}}} \\
  DeiT-Small \cite{touvron2021training} & 224$^2$ &  80.3\% & {80.9\%}\textcolor{blue}{\ \scriptsize{\textbf{(+0.6)}}}  & \baseline{}\textbf{81.3\%}\textcolor{blue}{\ \scriptsize{\textbf{(+1.0)}}} \\
  \end{tabular}}}
  \end{footnotesize}
  \captionsetup{font={footnotesize}}
  \vskip -0.1in
  \caption{\textbf{Higher accuracy with the same training cost} (ET++: EfficientTrain++). Here our method is configured to have the same training cost as the baselines (\emph{e.g.}, for ET++, we set equivalent training epochs = 300).
  \label{tab:high_acc}
  }
  \vskip -0.25in
\end{table}

\textbf{Adapted to varying training budgets.}
Our method can conveniently adapt to varying computational budgets for training, \emph{i.e.}, by simply modulating the number of total training epochs on top of Tables \ref{tab:EfficientTrain} and \ref{tab:EfficientTrain_plus}. As representative examples, we report the curves of validation accuracy v.s. computational training cost in Figure \ref{fig:acc_vs_cost} for both EfficientTrain++ and the baselines. The advantage of our method is even more significant under the constraint of a relatively smaller training cost, \emph{e.g.}, it outperforms the baseline by 3.1\% (79.8\% v.s. 76.7\%) on top of the 100-epoch trained DeiT-Small. We attribute this to the greater importance of efficient training algorithms in the scenarios of limited training resources. In addition, we observe that, if we mainly hope to achieve the same performance as the baselines, EfficientTrain++ is able to save the training cost by $2-3\times$ at most, which is more significant than the results in Table \ref{tab:img_1k_main_result}. These aforementioned observations can also be easily obtained on top of EfficientTrain (see: Appendix \textcolor{red}{B}). Furthermore, in Table \ref{tab:high_acc}, we deploy our proposed training curricula by allowing them to utilize the same training cost as the standard 300-epoch training pipeline \cite{liu2021swin, liu2022convnet}. The results show that our method significantly improves the accuracy (\emph{e.g.}, by 1.9\% for DeiT-Tiny). Interestingly, when properly trained using EfficientTrain++, the vanilla DeiT-Small network performs on par with the baseline Swin-Tiny model (81.3\%), without increasing the training wall-time.

\textbf{Adapted to any final input size $\gamma$.}
Our method can flexibly adapt to an arbitrary final input size $\gamma$. To attain this goal, the value of $B$ for the three stages of EfficientTrain or EfficientTrain++ can be simply adjusted to $160, (160+\gamma)/2,\gamma$ or $96, (96+\gamma)/2,\gamma$. As shown in Table \ref{tab:vary_res}, our method outperforms the baselines by large margins for $\gamma>224$ in terms of training efficiency.

\begin{table}[!t]
  \centering
  \begin{footnotesize}
  \setlength{\tabcolsep}{1mm}{
    \renewcommand\arraystretch{1.325}
    \resizebox{0.973\linewidth}{!}{
  \begin{tabular}{lcccccc}
    \multicolumn{1}{c}{\multirow{2}{*}{Models}} & \multirow{2}{*}{Method} & \multicolumn{3}{c}{Top-1 Accuracy / Wall-time Training Speedup} & \!\!\!  \\[-0.2ex]
     & & \multicolumn{1}{c}{224$^2$} & \multicolumn{1}{c}{384$^2$} & \multicolumn{1}{c}{512$^2$} \\
     \shline
     \multirow{2}{*}{\shortstack{Swin-Base\\[-0.2ex]\cite{liu2021swin}}} & Baseline &  83.4\% \!/\! $1.00\times$ & 84.5\% \!/\! $1.00\times$ & 84.7\%  \!/\! $1.00\times$   \\
     & \baseline{}\textbf{EfficientTrain} & \baseline{}\ \textbf{83.6\%} \!/\! $\bm{1.50\times}$  &  \baseline{}\ \textbf{84.7\%} \!/\! \textbf{$\bm{2.91\times}$}  &  \baseline{}\ \textbf{85.1\%} \!/\! \textbf{$\bm{3.37\times$}}  \\
  \end{tabular}}}
  \end{footnotesize}
  \captionsetup{font={footnotesize}}
  \vskip -0.09in
  \caption{\textbf{Adapted to different final input sizes.} Swin-Base is selected as a representative example since larger models typically benefit more from larger input sizes.
  \label{tab:vary_res}
  }
  \vskip -0.1in
\end{table}

\textbf{Orthogonal to $224^2$ pre-training + $\gamma^2$ fine-tuning.}
In particular, in some cases, existing works find it efficient to fine-tune $224^2$ pre-trained models to a target test input size $\gamma^2$  \cite{liu2021swin, liu2022convnet, dong2021cswin}. Here our method can be directly leveraged for more efficient pre-training (\emph{e.g.}, $\gamma\!=\!384$ in Tables \ref{tab:img_22K_main_result}).

\subsubsection{Comparisons with Existing Efficient Training Methods}

\textbf{Comparisons with state-of-the-art efficient training algorithms}
are summarized in Table \ref{tab:img1k_vs_baseline}. Our method is comprehensively compared with 1) the recently proposed sample-wise \cite{kumar2010self, zhou2018minimax, guo2018curriculumnet, NEURIPS2020_62000dee, qin2024infobatch} or regularization-wise \cite{sinha2020curriculum, dogan2020label} curriculum learning approaches; 2) the progressive learning algorithms \cite{tan2021efficientnetv2, li2022automated}; and 3) the efficient training methods for vision Transformers \cite{Li2020Budgeted, Touvron2022DeiTIR, xia2023budgeted}. Notably, some of these methods are not developed on top of the state-of-the-art training pipeline we consider (\emph{i.e.}, `AugRegs' in Table \ref{tab:img1k_vs_baseline}, see details in Appendix \textcolor{red}{A}). When comparing our method with them, we also implement our method without AugRegs (`RandAug' in our method is removed in this scenario), and adopt the same training settings as them (\emph{e.g.}, training ResNet-18/50 for 90/200 epochs). The results in Table \ref{tab:img1k_vs_baseline} indicate that EfficientTrain/EfficientTrain++ outperforms all the competitive baselines in terms of both accuracy and training efficiency. Moreover, the simplicity of our method enables it to be conveniently applied to different models and training settings by only modifying a few lines of code, which is an important advantage over baselines.

\textbf{Orthogonal to FixRes.}
FixRes \cite{touvron2019FixRes} reveals that there exists a discrepancy in the scale of images between the training and test inputs, and thus the inference with a larger resolution will yield a better test accuracy. However, our method does not leverage the gains of FixRes. We adopt the original inputs (\emph{e.g.}, $224^2$) at the final training stage, and hence the finally-trained model resembles the $224^2$-trained networks, while FixRes is orthogonal to our method. This fact can be confirmed by both the direct empirical evidence in Table \ref{tab:img1k_vs_baseline} (see: FixRes v.s. EfficientTrain/EfficientTrain++ + FixRes on top of the state-of-the-art CSWin Transformers \cite{dong2021cswin}) and the results in Table \ref{tab:img_22K_main_result} (see: Input Size=$384^2$).

\begin{table}[!t]
  \centering
  \begin{footnotesize}
  \setlength{\tabcolsep}{0.3mm}{
  \renewcommand\arraystretch{1.325}
  \resizebox{\linewidth}{!}{
  \begin{tabular}{clcccccc}
  \multicolumn{1}{c}{\multirow{2}{*}{Models}}  & \multicolumn{1}{c}{\multirow{2}{*}{\!\!\!\!\!\!\!\!Training Approach}}  & Training & \ \!Aug-\ \!  & {Top-1 } & Training \\[-0.2ex]
       &  & Epochs & Regs & Accuracy & Speedup \\
  \shline
  \multirow{2}{*}{\shortstack{ResNet-18\\\cite{He_2016_CVPR}}}
   & Smoothing Curriculum \cite{sinha2020curriculum} \textcolor{gray}{\tiny{(\textit{NeurIPS'20})}}\!\!\!\!\!\!  & 90 & \xmark & 71.0\% & $1.00\times$ \\
    & \baseline{}\textbf{EfficientTrain}  & \baseline{}\textbf{90} & \baseline{}\xmark & \baseline{}\textbf{71.0\%} & \baseline{}\ $\bm{1.48\times}$ \\
  \hline
  \multirow{11}{*}{\shortstack{\ ResNet-50\ \\\cite{He_2016_CVPR}}}  & 
  Self-paced Learning \cite{kumar2010self} \textcolor{gray}{\tiny{(\textit{NeurIPS'10})}}  & 200 & \xmark & 73.2\% & $1.15\times$ \\
   & Minimax Curriculum \cite{zhou2018minimax} \textcolor{gray}{\tiny{(\textit{ICLR'18})}}\!\!\!\!\!  & 200 & \xmark & 75.1\% & $1.97\times$ \\
   & DIH Curriculum \cite{NEURIPS2020_62000dee} \textcolor{gray}{\tiny{(\textit{NeurIPS'20})}}  & 200 & \xmark & 76.3\% & $2.45\times$ \\
   &  \baseline{}\textbf{EfficientTrain}  & \baseline{}\textbf{200} & \baseline{}\xmark & \baseline{}\textbf{77.5\%} & \baseline{}${1.44\times}$ \\
  &CurriculumNet \cite{guo2018curriculumnet} \textcolor{gray}{\tiny{(\textit{ECCV'18})}}  & 90 & \xmark & 76.1\% & $<\!2.22\times$ \\
   & Label-sim. Curriculum \cite{dogan2020label} \textcolor{gray}{\tiny{(\textit{ECCV'20})}}\!\!\!\!\!\!\!  & 90 & \xmark & 76.9\% & $2.22\times$ \\
   & \baseline{}\textbf{EfficientTrain}  & \baseline{}\textbf{90} & \baseline{}\xmark & \baseline{}{77.0\%} & \baseline{}\ $\bm{3.21\times}$ \\
  \cline{2-6}
   & Progressive Learning \cite{tan2021efficientnetv2} \textcolor{gray}{\tiny{(\textit{ICML'21})}}  & 350 & \cmark & 78.4\% & $1.21\times$ \\
   & InfoBatch$^\dagger$ \cite{qin2024infobatch} \textcolor{gray}{\tiny{(\textit{ICLR'24})}}  & 300 & \cmark & 78.6\% & $1.39\times$ \\
    & \baseline{}\textbf{EfficientTrain}  & \baseline{}\textbf{300} & \baseline{}\cmark & \baseline{}{79.4\%} & \baseline{}${1.44\times}$ \\
    & \baseline{}\textbf{EfficientTrain++}  & \baseline{}\textbf{200} & \baseline{}\cmark & \baseline{}\textbf{79.6\%} & \baseline{}\ $\bm{1.45\times}$ \\
  \hline
  \multirow{3}{*}{\shortstack{\ DeiT-\\[-0.2ex]Tiny\ \\[-0.4ex]\cite{touvron2021training}}}
   & Auto Progressive Learning \cite{li2022automated} \textcolor{gray}{\tiny{(\textit{CVPR'22})}}\!\!\!\!  & 300 & \cmark & 72.4\% & $1.51\times$ \\
    & \baseline{}\textbf{EfficientTrain}  & \baseline{}\textbf{300} & \baseline{}\cmark & \baseline{}{73.3\%} & \baseline{}${1.55\times}$ \\
    & \baseline{}\textbf{EfficientTrain++}  & \baseline{}\textbf{200} & \baseline{}\cmark & \baseline{}\textbf{73.7\%} & \baseline{}\ $\bm{1.64\times}$ \\
  \hline
  \multirow{11}{*}{\shortstack{\ DeiT-\\[-0.2ex]Small\ \\\cite{touvron2021training}}}  & Progressive Learning \cite{tan2021efficientnetv2} \textcolor{gray}{\tiny{(\textit{ICML'21})}}  & 100 & \cmark & 72.6\% & \ $\bm{1.54\times}$ \\
  & Auto Progressive Learning \cite{li2022automated} \textcolor{gray}{\tiny{(\textit{CVPR'22})}}\!\!\!\!  & 100 & \cmark & 74.4\% & $1.41\times$ \\
  & Budgeted ViT \cite{xia2023budgeted} \textcolor{gray}{\tiny{(\textit{ICLR'23})}}\!\!\!\!  & 128 & \cmark & 74.5\% & $1.34\times$ \\
  & \baseline{}\textbf{EfficientTrain}  & \baseline{}\textbf{100} & \baseline{}\cmark & \baseline{}\textbf{76.4\%} & \baseline{}${1.51\times}$ \\
  \cline{2-6}
  & Budgeted Training$^\dagger$ \cite{Li2020Budgeted} \textcolor{gray}{\tiny{(\textit{ICLR'20})}}  & 225 & \cmark & 79.6\% & $1.33\times$ \\
  & Progressive Learning$^\dagger$ \cite{tan2021efficientnetv2} \textcolor{gray}{\tiny{(\textit{ICML'21})}}  & 300 & \cmark & 79.5\% & $1.49\times$ \\
  & Auto Progressive Learning \cite{li2022automated} \textcolor{gray}{\tiny{(\textit{CVPR'22})}}\!\!\!\!  & 300 & \cmark & 79.8\% & $1.42\times$ \\
  & DeiT III \cite{Touvron2022DeiTIR} \textcolor{gray}{\tiny{(\textit{ECCV'22})}}\!\!\!\!  & 300 & \cmark & 79.9\% & $1.00\times$ \\
  & Budgeted ViT \cite{xia2023budgeted} \textcolor{gray}{\tiny{(\textit{ICLR'23})}}\!\!\!\!  & 303 & \cmark & 80.1\% & $1.34\times$ \\
  & \baseline{}\textbf{EfficientTrain}  & \baseline{}\textbf{300} & \baseline{}\cmark & \baseline{}{80.4\%} & \baseline{}${1.51\times}$ \\
  & \baseline{}\textbf{EfficientTrain++} & \baseline{}\textbf{200} & \baseline{}\cmark & \baseline{}\textbf{81.0\%} & \baseline{}\ $\bm{1.60\times}$ \\
  \hline
   \multirow{6}{*}{\shortstack{CSWin-\\[-0.2ex]Tiny\\[-0.2ex]\cite{dong2021cswin}}}   
   & Progressive Learning$^\dagger$ \cite{tan2021efficientnetv2} \textcolor{gray}{\tiny{(\textit{ICML'21})}}  & 300 & \cmark & 82.3\% & $1.51\times$ \\
    & \baseline{}\textbf{EfficientTrain}  & \baseline{}\textbf{300} & \baseline{}\cmark & \baseline{}{82.8\%} & \baseline{}\ $\bm{1.55\times}$ \\
    & \baseline{}\textbf{EfficientTrain++}  & \baseline{}\textbf{200} & \baseline{}\cmark & \baseline{}\textbf{82.9\%} & \baseline{}${1.50\times}$ \\
   \cline{2-6}
    & FixRes$^\dagger$ \cite{touvron2019FixRes} \textcolor{gray}{\tiny{(\textit{NeurIPS'19})}}  & 300 & \cmark & 82.9\% & $1.00\times$ \\
     & \baseline{}\textbf{EfficientTrain} + FixRes  & \baseline{}\textbf{300} & \baseline{}\cmark & \baseline{}{83.1\%} & \baseline{}\ $\bm{1.55\times}$ \\
     & \baseline{}\textbf{EfficientTrain++} + FixRes  & \baseline{}\textbf{200} & \baseline{}\cmark & \baseline{}\textbf{83.3\%} & \baseline{}${1.50\times}$ \\

  \hline
  \multirow{6}{*}{\shortstack{CSWin-\\[-0.2ex]Small\\[-0.2ex]\cite{dong2021cswin}}}   
  & Progressive Learning$^\dagger$ \cite{tan2021efficientnetv2} \textcolor{gray}{\tiny{(\textit{ICML'21})}}  & 300 & \cmark & 83.3\% & $1.48\times$ \\
    & \baseline{}\textbf{EfficientTrain}  & \baseline{}\textbf{300} & \baseline{}\cmark & \baseline{}\textbf{83.6\%} & \baseline{}${1.51\times}$ \\
    & \baseline{}\textbf{EfficientTrain++}  & \baseline{}\textbf{200} & \baseline{}\cmark & \baseline{}\textbf{83.6\%} & \baseline{}\ $\bm{1.52\times}$ \\
  \cline{2-6}
    & FixRes$^\dagger$ \cite{touvron2019FixRes} \textcolor{gray}{\tiny{(\textit{NeurIPS'19})}}  & 300 & \cmark & 83.7\% & $1.00\times$ \\
    & \baseline{}\textbf{EfficientTrain} + FixRes & \baseline{}\textbf{300} & \baseline{}\cmark & \baseline{}{83.8\%} & \baseline{}${1.51\times}$ \\
    & \baseline{}\textbf{EfficientTrain++} + FixRes & \baseline{}\textbf{200} & \baseline{}\cmark & \baseline{}\textbf{83.9\%} & \baseline{}\ $\bm{1.52\times}$ \\
  \end{tabular}}}
  \vskip -0.09in
  \captionsetup{font={footnotesize}}
  \caption{\textbf{EfficientTrain v.s. state-of-the-art efficient training algorithms on ImageNet-1K.} 
  Here `AugRegs' denotes the widely-used holistic combination of various model regularization and data augmentation techniques \cite{touvron2021training, wang2021pyramid, liu2021swin, dong2021cswin, liu2022convnet}. For EfficientTrain++, we report the number of equivalent training epochs.
  $\dagger$: our reproduced results.
  \label{tab:img1k_vs_baseline}}
  \end{footnotesize}
  \vskip -0.1in
\end{table}

 \begin{table*}[!t]
  \centering
  \begin{footnotesize}
  \setlength{\tabcolsep}{2mm}{
  \renewcommand\arraystretch{1.365}
  \resizebox{0.985\linewidth}{!}{
  \begin{tabular}{lccccccc}
  \multicolumn{1}{c}{\multirow{3.3}{*}{Models}} 
  & \multirow{3.3}{*}{\!\!\!\#Param.\!\!\!}  
  & \multirow{3.3}{*}{\shortstack{\!\#FLOPs\\(224$^2$/384$^2$)}\!}  
  & \multirow{3.3}{*}{\shortstack{Pre-training\\Approach}} 
  & \multicolumn{2}{c}{Top-1 Accuracy} 
  & \multirow{3.3}{*}{\shortstack{Wall-time Pre-training\\Cost (in GPU-days)}}
  & \multirow{3.3}{*}{\shortstack{Time Saving\\[0.4ex](for an 8-GPU node)}}  \\[-0.5ex]
  & & & & \multicolumn{2}{c}{(fine-tuned to ImageNet-1K)}   &  & \\[-0.5ex]
  & & & & \!Input size: 224$^2$\! & \!Input size: 384$^2$\! & &  \\  
  \shline

  \multirow{3}{*}{ConvNeXt-Base \cite{liu2022convnet}} 
  & \multirow{3}{*}{89M} & \multirow{3}{*}{15.4G/45.1G} & Baseline &  85.8\% & 86.7\% & 181.9  & -- \\
  & & & \textbf{EfficientTrain}  &  {85.8\%} & {86.8\%} & 122.4\textcolor{blue}{\ \scriptsize{\textbf{($\bm{\downarrow\!1.49\times}$)}}} & \textit{22.7$\ \!\to\!\ $15.3 Days} \\
  & & & \baseline{}\textbf{EfficientTrain++}  &  \baseline{}\textbf{85.9\%} & \baseline{}\textbf{86.9\%} & \baseline{}\textbf{122.3}\textcolor{blue}{\ \scriptsize{\textbf{($\bm{\downarrow\!1.49\times}$)}}} & \baseline{}\textit{\textbf{22.7$\ \!\to\!\ $15.3 Days}} \\
  \hhline{|~-------|}
  \multirow{3}{*}{ConvNeXt-Large \cite{liu2022convnet}} 
  & \multirow{3}{*}{198M} & \multirow{3}{*}{34.4G/101.0G} & Baseline &  86.4\% & 87.3\% & 375.8 & -- \\
  & & & \textbf{EfficientTrain}  &  {86.4\%} & {87.3\%} & \textbf{243.8}\textcolor{blue}{\ \scriptsize{\textbf{($\bm{\downarrow\!1.54\times}$)}}} & \textit{\textbf{47.0$\ \!\to\!\ $30.5 Days}} \\
  & & & \baseline{}\textbf{EfficientTrain++}  &  \baseline{}\textbf{86.5\%} & \baseline{}\textbf{87.3\%} & \baseline{}249.3\textcolor{blue}{\ \scriptsize{\textbf{($\bm{\downarrow\!1.51\times}$)}}} & \baseline{}\textit{47.0$\ \!\to\!\ $31.2 Days} \\
  \hline
  \multirow{4}{*}{CSWin-Base \cite{dong2021cswin}} 
  & \multirow{4}{*}{78M} & \multirow{4}{*}{15.0G/47.0G} & Baseline &  86.0\% & 87.2\% & 126.9 & -- \\
  & & & \textbf{EfficientTrain}  &  {86.2\%} & {87.4\%} & \hspace{0.5em}83.7\textcolor{blue}{\ \scriptsize{\textbf{($\bm{\downarrow\!1.52\times}$)}}} & \textit{15.9$\ \!\to\!\ $10.5 Days} \\
  & & & \baseline{}  &  \baseline{}\textbf{86.3\%} & \baseline{}\textbf{87.4\%} & \baseline{}\hspace{0.5em}62.9\textcolor{blue}{\ \scriptsize{\textbf{($\bm{\downarrow\!2.02\times}$)}}} & \baseline{}\textit{15.9$\ \!\to\!\ $7.9 Days} \\
  & & & \baseline{}\multirow{-2}{*}{\textbf{EfficientTrain++}}  &  \baseline{}86.1\% & \baseline{}87.1\% & \baseline{}\hspace{0.51em}\textbf{41.9}\textcolor{blue}{\ \scriptsize{\textbf{($\bm{\downarrow\!3.03\times}$)}}} & \baseline{}\textit{\textbf{15.9$\ \!\to\!\ $5.2 Days}} \\
   \hhline{|~-------|}
   \multirow{4}{*}{CSWin-Large \cite{dong2021cswin}} 
   & \multirow{4}{*}{173M} & \multirow{4}{*}{31.5G/96.8G} & Baseline &  86.8\% & 87.9\% & 237.8 & -- \\
   & & & \textbf{EfficientTrain}  &  {86.9\%} & {87.9\%} & 155.8\textcolor{blue}{\ \scriptsize{\textbf{($\bm{\downarrow\!1.53\times}$)}}} & \textit{29.7$\ \!\to\!\ $19.5 Days} \\
   & & & \baseline{}  &  \baseline{}\textbf{87.0\%} & \baseline{}\textbf{87.9\%} & \baseline{}119.0\textcolor{blue}{\ \scriptsize{\textbf{($\bm{\downarrow\!2.00\times}$)}}} & \baseline{}\textit{29.7$\ \!\to\!\ $14.9 Days} \\
   & & & \baseline{}\multirow{-2}{*}{\textbf{EfficientTrain++}}  &  \baseline{}86.8\% & \baseline{}87.8\% & \baseline{}\hspace{0.51em}\textbf{79.3}\textcolor{blue}{\ \scriptsize{\textbf{($\bm{\downarrow\!3.00\times}$)}}} & \baseline{}\textit{\textbf{29.7$\ \!\to\!\ $9.9 Days}} \\
  \end{tabular}}}
  \vskip -0.08in
  \captionsetup{font={footnotesize}}
  \caption{\textbf{Results with ImageNet-22K (IN-22K) pre-training.} The models are pre-trained on IN-22K with or without our method, fine-tuned on the ImageNet-1K (IN-1K) training set, and evaluated on the IN-1K validation set. We present the results by setting the total number of equivalent pre-training epochs of EfficientTrain++ to 2/3, 1/2, and 1/3 of the baselines, within which range our method generally performs on par with the baselines in terms of accuracy. The wall-time pre-training cost is benchmarked on NVIDIA 3090 GPUs.}
  \label{tab:img_22K_main_result}
  \end{footnotesize}
  \vskip -0.1in
\end{table*}

\begin{table*}[!t]
  \centering
  \begin{footnotesize}
  \setlength{\tabcolsep}{2.5mm}{
  \renewcommand\arraystretch{1.365}
  \resizebox{0.81\linewidth}{!}{
  \begin{tabular}{lcccccc}
  \multicolumn{1}{c}{\multirow{3.3}{*}{Models}} 
  & \multirow{3.3}{*}{\shortstack{Pre-training Approach}} 
  & \multirow{3.3}{*}{\shortstack{Early\\Large Batch}}
  & \multicolumn{2}{c}{Top-1 Accuracy} 
  & \multicolumn{2}{c}{Wall-time Pre-training} \\[-0.5ex]
  & & & \multicolumn{2}{c}{(fine-tuned to ImageNet-1K)}   &  \multicolumn{2}{c}{Speedup} \\[-0.5ex]
  & & & \!Input size: 224$^2$\! & \!Input size: 384$^2$\! & 16 GPUs & 64 GPUs \\  
  \shline
  \multirow{5}{*}{CSWin-Base \cite{dong2021cswin}} 
  & Baseline & -- &  86.0\% & 87.2\% & ${1.0\times}$ & ${1.0\times}$ \\
  \hhline{|~~-----|}
  & \multirow{2}{*}{\textbf{EfficientTrain++}} &  \xmark  &  86.3\% & 87.4\% & ${2.0\times}$ & ${1.4\times}$ \\
  & & \baseline{}\cmark  & \baseline{}\textbf{86.4\%} & \baseline{}\textbf{87.4\%} & \baseline{}\ \!$\bm{2.0\times}$ & \baseline{}\ \!$\bm{2.0\times}$ \\
  \hhline{|~~-----|}
  & \multirow{2}{*}{\textbf{EfficientTrain++}} &  \xmark  &  \textbf{86.1\%} & 87.1\% & ${3.0\times}$ & ${2.1\times}$ \\
  & & \baseline{}\cmark  & \baseline{}86.0\% & \baseline{}\textbf{87.1\%} & \baseline{}\ \!$\bm{3.0\times}$ & \baseline{}\ \!$\bm{2.9\times}$ \\
   \hhline{|~------|}
   \multirow{5}{*}{CSWin-Large \cite{dong2021cswin}} 
   & Baseline & -- &  86.8\% & 87.9\% & ${1.0\times}$ & ${1.0\times}$ \\
   \hhline{|~~-----|}
   &  \multirow{2}{*}{\textbf{EfficientTrain++}} &  \xmark  &  87.0\% & 87.9\% & ${2.0\times}$  & ${1.4\times}$ \\
   & & \baseline{}\cmark  & \baseline{}\textbf{87.1\%} & \baseline{}\textbf{88.1\%} & \baseline{}\ \!$\bm{2.0\times}$ & \baseline{}\ \!$\bm{1.9\times}$ \\
   \hhline{|~~-----|}
   & \multirow{2}{*}{\textbf{EfficientTrain++}} &  \xmark  &  86.8\% & 87.8\% & ${3.0\times}$ & ${2.1\times}$ \\
   & & \baseline{}\cmark  & \baseline{}\textbf{86.8\%} & \baseline{}\textbf{87.8\%} & \baseline{}\ \!$\bm{3.0\times}$ & \baseline{}\ \!$\bm{2.9\times}$ \\
  \end{tabular}}}
  \vskip -0.08in
  \captionsetup{font={footnotesize}}
  \caption{\textbf{Effects of introducing the early large batch mechanism (see: Section \ref{sec:ET_plus_3}) in EfficientTrain++.} The scenario of ImageNet-22K pre-training + ImageNet-1K fine-tuning is considered here. The experimental settings are the same as Table \ref{tab:img_22K_main_result}.
  \label{tab:results_early_large_bs}}
  \end{footnotesize}
  \vskip -0.075in
\end{table*}


\subsubsection{Main Results on ImageNet-22K}

\textbf{ImageNet-22K pre-training.}
One of the important advantages of modern visual backbones is their excellent scalability with a growing amount of training data \cite{dong2021cswin, liu2022convnet}. To this end, we further verify the effectiveness of our method on the larger ImageNet-22K benchmark dataset. The results are summarized in Table \ref{tab:img_22K_main_result}, where the models are pre-trained on ImageNet-22K, and evaluated by being fine-tuned to ImageNet-1K. We implement EfficientTrain/EfficientTrain++ at the pre-training stage, which accounts for the vast majority of the total computation/time cost. One can observe that, similar to ImageNet-1K, our method performs at least on par with the baselines on top of both ConvNets and vision Transformers, while achieving a significant training speedup of up to $2\!-\!3\times$. A highlight from the results is that EfficientTrain++ saves a considerable amount of real training time, \emph{e.g.}, 158.5 GPU-days (237.8 v.s. 79.3) for CSWin-Large, which correspond to $\sim$20 days for a standard computational node with 8 GPUs. 



\subsubsection{Implementation Techniques for EfficientTrain++}
\label{sec:validate_implementation_technique}

In the following, we investigate the effectiveness of the two implementation techniques for EfficientTrain++ proposed in Section \ref{sec:ET_plus_3}. We demonstrate that these techniques effectively lower the thresholds of implementing EfficientTrain++, and enable our method to acquire significant practical training speedup in a broader range of scenarios.

\textbf{Large-scale parallel training: early large batch.}
Since this mechanism is designed to enable EfficientTrain++ to make full use of more GPUs, we evaluate it in the scenario of pre-training CSWin-Base/Large on ImageNet-22K. CSWin Transformers are representative examples of state-of-the-art deep networks. Due to their large model size and the considerable amount of ImageNet-22K training data, there is a practical demand of pre-training the models in parallel using a number of GPUs (\emph{e.g.}, 64). In Table \ref{tab:results_early_large_bs}, we study the effects of introducing larger batches at earlier learning stages (named as `early large batch'), where our implementation follows from the statements in Section \ref{sec:ET_plus_3}. In general, the vanilla EfficientTrain++ is able to achieve significant training speedup when leveraging 16 GPUs. However, its practical efficiency notably decreases when the number of GPUs grows to 64. In contrast, the wall-time pre-training speedup of EfficientTrain++ equipped with the early large batch mechanism on 64 GPUs is approximately identical to exploiting 16 GPUs. This observation demonstrates that enlarging batch sizes for small $B$ effectively enhances the scalability of EfficientTrain++ with the growing number of GPUs. Besides, it is worth noting that this mechanism does not affect the performance of the models.

\begin{table}[!t]
  \centering
  \vskip -0.075in
  \begin{footnotesize}
  \setlength{\tabcolsep}{0.6mm}{
  \renewcommand\arraystretch{1.38}
  \resizebox{\linewidth}{!}{
  \begin{tabular}{lcccccc}
  \multicolumn{1}{c}{\multirow{2}{*}{Models}} 
  & Training
  & \!\!Replay\!\!
  & Top-1 
  & Computational 
  & Peak Data Pre- \\[-0.5ex]
  & Approach & Buffer & Accuracy & Training Speedup & processing Loads \\
  \shline
  \multirow{2}{*}{\shortstack{DeiT-Tiny\\\cite{touvron2021training}}} 
  & \multirow{2}{*}{\textbf{ET++}} &  \xmark  &  {73.7\%}  & ${1.56\times}$ & 12630 images/s \\
  & & \baseline{}\cmark  & \baseline{}73.6\% & \baseline{}${1.56\times}$ & \baseline{}\textbf{6315 images/s} \\
  \hhline{|~------|}
  \multirow{2}{*}{\shortstack{DeiT-Small\\\cite{touvron2021training}}} 
  & \multirow{2}{*}{\textbf{ET++}} &  \xmark  &  {81.0\%}  & ${1.52\times}$ & 5329 images/s \\
  & & \baseline{}\cmark  & \baseline{}80.8\% & \baseline{}${1.52\times}$ & \baseline{}\textbf{2665 images/s} \\
  \hline
  \multirow{2}{*}{\shortstack{CSWin-Tiny\!\!\\\cite{dong2021cswin}}} 
  & \multirow{2}{*}{\textbf{ET++}} &  \xmark  &  {82.9\%}  & ${1.52\times}$ & 2294 images/s \\
  & & \baseline{}\cmark  & \baseline{}82.9\% & \baseline{}${1.52\times}$ & \baseline{}\textbf{1147 images/s} \\
  \hhline{|~------|}
  \multirow{2}{*}{\shortstack{CSWin-Small\!\\\cite{dong2021cswin}}} 
  & \multirow{2}{*}{\textbf{ET++}} &  \xmark  &  {83.6\%}  & ${1.52\times}$ & 1517 images/s \\
  & & \baseline{}\cmark  & \baseline{}83.6\% & \baseline{}${1.52\times}$ & \baseline{}\textbf{759 images/s} \\
  \end{tabular}}}
  \vskip -0.08in
  \captionsetup{font={footnotesize}}
  \caption{\textbf{Effects of introducing the replay buffer (see: Section \ref{sec:ET_plus_3}) in EfficientTrain++ (ET++).} The scenario of training relatively small models on ImageNet-1K is considered here, where high data pre-processing loads tend to be a practical bottleneck for efficient training. The experimental settings are the same as Table \ref{tab:img_1k_main_result}. The value of `peak data pre-processing loads' refers to the peak loads introduced by saturating all the computational cores of a single NVIDIA 3090 GPU.
  \label{tab:replay_buffer}}
  \end{footnotesize}
  \vskip -0.2in
\end{table}

\textbf{Reducing data pre-processing loads: replay buffer.}
To investigate this mechanism, we take training relatively light-weighted models (\emph{e.g.}, DeiT-T/S and CSWin-T/S) as representative examples. Due to the smaller size of these networks, their training speed on GPUs is usually fast, such that they put higher demands on CPU and memory in terms of the throughput of preparing training inputs. Hence, the data pre-processing load is a notable potential bottleneck for training them efficiently. We train these models by introducing the replay buffer as stated in Section \ref{sec:ET_plus_3}, and compare the results with the original EfficientTrain++ in Table \ref{tab:replay_buffer}. One can observe that the replay buffer contributes to the significantly reduced peak data pre-processing loads, and maintains a competitive performance with the baselines.

\begin{table}[!t]
  \centering
  \begin{footnotesize}
  \setlength{\tabcolsep}{1.1mm}{
    \renewcommand\arraystretch{1.37}
    \resizebox{0.992\linewidth}{!}{
  \begin{tabular}{lcccccccc}
  \multicolumn{1}{c}{\multirow{2.05}{*}{Methods}} & \multirow{2.05}{*}{\!\!Backbone\!}  & \multirow{2.05}{*}{\!\#Param.\!\!\!}   & \multirow{2.05}{*}{\shortstack{\!Top-1 Accuracy\!\\(fine-tuning)}} & \multirow{2.05}{*}{\shortstack{\!Pre-training\!\\Speedup}} \\[-0.2ex]
  & &  & & \\
  \shline
    DINO \cite{caron2021emerging} \textcolor{gray}{\tiny{(\textit{ICCV'21})}} & ViT-B & 86M &82.8\% & -- \\
    MoCo V3 \cite{chen412empirical} \textcolor{gray}{\tiny{(\textit{ICCV'21})}} & ViT-B & 86M &83.2\% & -- \\
    BEiT \cite{bao2022beit} \textcolor{gray}{\tiny{(\textit{ICLR'22})}} & ViT-B & 86M &\hspace{0.49em}83.2\%$^\dagger$ & ${0.92\times}$ \\
    MaskFeat \cite{wei2022masked}  \textcolor{gray}{\tiny{(\textit{CVPR'22})}} & ViT-B & 86M & 83.6\% & ${1.50\times}$ \\
    LoMaR \cite{chen2022efficient}  \textcolor{gray}{\tiny{(\textit{arXiv'22})}} & \!ViT-B \!+\! RPE$^\ddagger$\! & 86M & 83.6\% & ${3.52\times}$ \\
    CAE \cite{ContextAutoencoder2022}  \textcolor{gray}{\tiny{(\textit{IJCV'23})}} & ViT-B & 86M & \hspace{0.49em}83.6\%$^\dagger$ & ${2.10\times}$ \\
    \multirow{2}{*}{\shortstack{LocalMIM \cite{wang2023masked} \textcolor{gray}{\tiny{(\textit{CVPR'23})}}\\ + HOG target \cite{wei2022masked} \textcolor{gray}{\tiny{(\textit{CVPR'22})}}}} 
    & \multirow{2}{*}{ViT-B} & \multirow{2}{*}{86M} & \multirow{2}{*}{\textbf{83.7\%}} & \multirow{2}{*}{${3.14\times}$} \\
    \\
    \hline
    MAE$_{\textnormal{ 1600-epoch}}$ \cite{he2022masked} \textcolor{gray}{\tiny{(\textit{CVPR'22})}}\!\!\!\! & ViT-B & 86M &83.6\% & ${1.00\times}$ \\
    MAE$_{\textnormal{ 400-epoch}}$ \cite{he2022masked} \textcolor{gray}{\tiny{(\textit{CVPR'22})}}\!\!\!\! & ViT-B & 86M &83.0\% & $\bm{4.00\times}$ \\
    MAE$_{\textnormal{ 1600-epoch}}$ + HOG & ViT-B & 86M &\textbf{83.7\%} & ${1.00\times}$ \\
    MAE$_{\textnormal{ 400-epoch}}$ + HOG & ViT-B & 86M &83.2\% & $\bm{4.00\times}$ \\
    \baseline{}\textbf{MAE (ET++)$_{\textnormal{ 400-epoch}}$ + HOG}\!\! & \baseline{}ViT-B & \baseline{}86M &\baseline{}\textbf{83.7\%} & \baseline{}${3.98\times}$ \\
  \end{tabular}}}
  \end{footnotesize}
  \captionsetup{font={footnotesize}}
  \vskip -0.085in
  \caption{\textbf{Self-supervised learning results on top of MAE \cite{he2022masked}}. The model is pre-trained on ImageNet-1K with EfficientTrain++ (ET++), and evaluated by end-to-end fine-tuning \cite{he2022masked}. $\dagger$: using additional data (the DALL-E tokenizer \cite{ramesh2021zero}). $\ddagger$: using relative position encoding (RPE).
  }
  \label{tab:mae_result}
  \vskip -0.1in
\end{table}


\vspace{-0.75ex}
\subsection{Self-supervised Learning}
\label{sec:self_supervised_learning}
\vspace{-0.25ex}

\textbf{Results on top of Masked Autoencoders (MAE)}.
In addition to supervised learning, our method can also be conveniently applied to self-supervised learning algorithms since it only modifies the training inputs. Table \ref{tab:mae_result} presents a representative example, where we deploy EfficientTrain++ on top of MAE \cite{he2022masked} equipped with HOG reconstruction targets \cite{wei2022masked}. We also present the results of several recently proposed self-supervised learning approaches that use the same backbone and have comparable pre-training costs with us. One can observe that our method reduces the pre-training cost of MAE significantly while preserving the accuracy, outperforming all the competitive baselines.

\vspace{-0.75ex}
\subsection{Transfer Learning}
\label{sec:transfer_learning}
\vspace{-0.25ex}

\textbf{Downstream image recognition tasks.}
We first evaluate the transferability of the models trained with EfficientTrain/EfficientTrain++ by fine-tuning them on downstream classification datasets. The results are reported in Table \ref{tab:transferability}. Notably, following \cite{touvron2021training}, the 32$\times$32 images in CIFAR-10/100 \cite{krizhevsky2009learning} are resized to 224$\times$224 for pre-processing, and thus the discriminative patterns are mainly distributed within the lower-frequency components. On the contrary, Flowers-102 \cite{nilsback2008automated} and Stanford Dogs \cite{khosla2011novel} are fine-grained visual recognition datasets where the high-frequency clues contain important discriminative information (\emph{e.g.}, the detailed facial/skin characteristics to distinguish between the species of dogs). One can observe that our method yields competitive or better transfer learning performance than the baselines on both types of datasets. In other words, although our method learns to exploit the lower/higher-frequency information via an ordered curriculum, the finally obtained models can leverage both types of information effectively.

\textbf{Object detection \& instance segmentation.}
To investigate transferring our pre-trained models to more complex computer vision tasks, we initialize the backbones of representative detection and instance segmentation frameworks with the models pre-trained using EfficientTrain/EfficientTrain++.  Results on MS COCO are reported in Table \ref{tab:coco}. When reducing the pre-training cost by $\bm{\sim\!1.5\!\times}$, our method significantly outperforms the baselines in terms of detection/segmentation performance in all scenarios. Our implementation follows from MMDetection \cite{mmdetection}.

\textbf{Semantic segmentation.}
Table \ref{tab:ade20k} further evaluates the performance of our method on downstream semantic segmentation tasks. We implement several representative backbones pre-trained with EfficientTrain++ as the initialization of encoders in UperNet \cite{xiao2018unified}, and fine-tune UperNet on ADE20K using MMSegmentation \cite{mmseg2020}. The observation from Table \ref{tab:ade20k} is similar to Tables \ref{tab:transferability} and \ref{tab:coco}. Our method can save pre-training cost by at least $\bm{1.5\!\times}$, while effectively enhancing downstream performance. For example, on top of DeiT-Small, EfficientTrain++ improves mIoU/$\textnormal{mIoU}^{\textnormal{MS+Flip}}$ by 1.1/1.2 under the 160k-step-training setting.

\begin{table}[t!]
  \centering
    \begin{footnotesize}
    \captionsetup{font={footnotesize}}
    \setlength{\tabcolsep}{1.0mm}{
    \renewcommand\arraystretch{1.38}
    \resizebox{\linewidth}{!}{
    \begin{tabular}{cccccccccc}
    \multicolumn{1}{c}{\multirow{3}{*}{\!\!Backbone\!\!\!\!}} & \multicolumn{2}{c}{\multirow{2}{*}{\ \ \ \ \ Pre-training}} & \multicolumn{4}{c}{Top-1 Accuracy} \\[-0.5ex]
     & & & \multicolumn{4}{c}{(fine-tuned to downstream datasets)} \\[-0.5ex]
    & Method & \!\!\!\!Speedup\! & C10 & C100 & \!Flowers-102\! & Stanford Dogs \\
    \shline
    \multirow{3}{*}{\shortstack{DeiT-S\\\cite{touvron2021training}}} & Baseline & $1.0\times$ & 98.39\% & 88.65\% & 96.57\% &  90.72\% \\ 
    & \baseline{}\!\textbf{EfficientTrain}\! & \baseline{}${1.5\times}$ & \baseline{}{98.47\%}  & \baseline{}{88.93\%} & \baseline{}{96.62\%} & \baseline{}{91.12\%} \\ 
    & \baseline{}\!\textbf{EfficientTrain++}\! & \baseline{}\ \!$\bm{1.6\times}$ & \baseline{}\textbf{98.51\%} & \baseline{}\textbf{89.55\%} & \baseline{}\textbf{97.38\%} & \baseline{}\textbf{91.49\%} \\ 
    \end{tabular}}}
    \vskip -0.085in
    \captionof{table}{\textbf{Transferability to downstream image recognition tasks.} The models are pre-trained on ImageNet-1K w/ or w/o EfficientTrain or EfficientTrain++, and fine-tuned to the downstream datasets to report the accuracy. C10/C100 refers to the CIFAR-10/100 datasets. \label{tab:transferability}}
    \end{footnotesize}
  \vskip -0.08in
  \end{table}

\begin{table}[t!]
  \centering
    \begin{footnotesize}
    \captionsetup{font={footnotesize}}
    \setlength{\tabcolsep}{0.6mm}{
    \renewcommand\arraystretch{1.38}
    \resizebox{\linewidth}{!}{
    \begin{tabular}{ccccccccccc}
    \multicolumn{1}{c}{\multirow{2}{*}{Backbone}} & \multicolumn{2}{c}{\ \ \ \ \ Pre-training} &  \multirow{2}{*}{$\textnormal{AP}^{\textnormal{box}}$} &  \multirow{2}{*}{$\textnormal{AP}^{\textnormal{box}}_{\textnormal{50}}$} &  \multirow{2}{*}{$\textnormal{AP}^{\textnormal{box}}_{\textnormal{75}}$} &  \multirow{2}{*}{$\textnormal{AP}^{\textnormal{mask}}$} &  \multirow{2}{*}{$\textnormal{AP}^{\textnormal{mask}}_{\textnormal{50}}$} &  \multirow{2}{*}{$\textnormal{AP}^{\textnormal{mask}}_{\textnormal{75}}$} \\[-0.5ex]
     &  Method & \!\!Speedup &&&&&& \\
    \shline
    \\[-3.6ex]
    \multicolumn{9}{c}{\textbf{RetinaNet \cite{lin2017focal} ($1\times$ schedule)}} \\[-0.5ex]
    \multirow{2}{*}{\shortstack{Swin-T\\[-0.4ex]\cite{liu2021swin}}} & Baseline & $1.0\times$ & 41.7 & 62.9 & 44.5 & -- & -- & -- \\ 
    & \baseline{}\!\textbf{EfficientTrain}\!& \baseline{}$\bm{\ 1.5\times}$ & \baseline{}\textbf{41.8} & \baseline{}63.3 & \baseline{}44.6 & \baseline{}-- & \baseline{}-- & \baseline{}-- \\ 
    \shline
    \\[-3.6ex]
    \multicolumn{9}{c}{\textbf{Cascade Mask-RCNN \cite{cai2019cascade} ($1\times$ schedule)}} \\[-0.5ex]
    \multirow{3}{*}{\shortstack{Swin-T\\[-0.4ex]\cite{liu2021swin}}} & Baseline& $1.0\times$ & 48.1 & 67.0 & 52.1 & 41.5 & 64.2 & 44.9 \\ 
    & \baseline{}\!\textbf{EfficientTrain}\!& \baseline{}$\bm{\ 1.5\times}$ & \baseline{}{48.2} & \baseline{}67.5 & \baseline{}52.3 & \baseline{}{41.8} & \baseline{}64.6 & \baseline{}45.0  \\ 
    & \baseline{}\textbf{EfficientTrain++}\!& \baseline{}$\bm{\ 1.5\times}$ & \baseline{}\textbf{48.4} & \baseline{}67.6 & \baseline{}52.6 & \baseline{}\textbf{41.9} & \baseline{}64.8 & \baseline{}45.2  \\ 
    \hhline{|~---------|}
    \multirow{2}{*}{\shortstack{Swin-S\\[-0.4ex]\cite{liu2021swin}}} & Baseline& $1.0\times$ & 50.0 & 69.1 & 54.4 & 43.1 & 66.3 & 46.3 \\  
    & \baseline{}\textbf{EfficientTrain++}\!& \baseline{}$\bm{\ 1.5\times}$ & \baseline{}\textbf{50.7} & \baseline{}69.9 & \baseline{}55.0 & \baseline{}\textbf{43.7} & \baseline{}67.1 & \baseline{}47.3  \\ 
    \hhline{|~---------|}
    \multirow{2}{*}{\shortstack{Swin-B\\[-0.4ex]\cite{liu2021swin}}} & Baseline& $1.0\times$ & 50.9 & 70.2 & 55.5 & 44.0 & 67.4 & 47.4 \\ 
    & \baseline{}\textbf{EfficientTrain++}\!& \baseline{}$\bm{\ 1.5\times}$ & \baseline{}\textbf{51.3} & \baseline{}70.5 & \baseline{}55.9 & \baseline{}\textbf{44.3} & \baseline{67.6} & \baseline{}48.0  \\  

    \end{tabular}}}
    \vskip -0.085in
    \captionof{table}{\textbf{Object detection and instance segmentation on COCO.} We implement representative detection/segmentation algorithms on top of the backbones pre-trained w/ or w/o our method on ImageNet-1K. \label{tab:coco}}
    \end{footnotesize}
    \vskip -0.0925in
  \end{table}

\begin{table}[t!]
  \centering
    \begin{footnotesize}
    \captionsetup{font={footnotesize}}
    \setlength{\tabcolsep}{1.5mm}{
    \renewcommand\arraystretch{1.38}
    \resizebox{0.84\linewidth}{!}{
    \begin{tabular}{ccccc}
    \multicolumn{1}{c}{\multirow{2}{*}{Backbone}} & \multicolumn{2}{c}{Pre-training} &  \multirow{2}{*}{mIoU} &  \multirow{2}{*}{$\textnormal{mIoU}^{\textnormal{MS+Flip}}$$^\dagger$}  \\[-0.5ex]
     &  Method & Speedup && \\
    \shline
    \\[-3.6ex]
    \multicolumn{5}{c}{\textbf{UperNet \cite{xiao2018unified} (80k steps)}} \\[-0.5ex]
    \multirow{2}{*}{DeiT-S \cite{touvron2021training}} & Baseline & $1.0\times$ & 43.0 & 43.8 \\ 
    & \baseline{}\textbf{EfficientTrain++}& \baseline{}$\bm{1.6\times}$ & \baseline{}\textbf{43.8} & \baseline{}\textbf{44.9} \\ 
    \hhline{|~----|}
    \multirow{2}{*}{Swin-T \cite{liu2021swin}} & Baseline & $1.0\times$ & 43.3 & 44.5 \\ 
    & \baseline{}\textbf{EfficientTrain++}& \baseline{}$\bm{1.5\times}$ & \baseline{}\textbf{44.1} & \baseline{}\textbf{44.9} \\ 
    \hhline{|~----|}
    \multirow{2}{*}{Swin-S \cite{liu2021swin}} & Baseline & $1.0\times$ & 47.5 & 48.9 \\ 
    & \baseline{}\textbf{EfficientTrain++}& \baseline{}$\bm{1.5\times}$ & \baseline{}\textbf{47.8} & \baseline{}\textbf{49.2} \\ 
    \shline
    \\[-3.6ex]
    \multicolumn{5}{c}{\textbf{UperNet \cite{xiao2018unified} (160k steps)}} \\[-0.5ex]
    \multirow{2}{*}{DeiT-S \cite{touvron2021training}} & Baseline & $1.0\times$ & 43.1 & 44.0 \\ 
    & \baseline{}\textbf{EfficientTrain++}& \baseline{}$\bm{1.6\times}$ & \baseline{}\textbf{44.2} & \baseline{}\textbf{45.2} \\ 
    \hhline{|~----|}
    \multirow{2}{*}{Swin-T \cite{liu2021swin}} & Baseline & $1.0\times$ & 44.4 & 45.8 \\ 
    & \baseline{}\textbf{EfficientTrain++}& \baseline{}$\bm{1.5\times}$ & \baseline{}\textbf{44.8} & \baseline{}\textbf{46.1} \\ 
    \hhline{|~----|}
    \multirow{2}{*}{Swin-S \cite{liu2021swin}} & Baseline & $1.0\times$ & 47.7 & 49.2 \\ 
    & \baseline{}\textbf{EfficientTrain++}& \baseline{}$\bm{1.5\times}$ & \baseline{}\textbf{48.2} & \baseline{}\textbf{49.7} \\ 
    \hhline{|~----|}
    \multirow{2}{*}{Swin-B \cite{liu2021swin}} & Baseline & $1.0\times$ & 48.0 & 49.6 \\ 
    & \baseline{}\textbf{EfficientTrain++}& \baseline{}$\bm{1.5\times}$ & \baseline{}\textbf{48.5} & \baseline{}\textbf{50.0} \\ 

    \end{tabular}}}
    \vskip -0.085in
    \captionof{table}{\textbf{Semantic segmentation on ADE20K.} \label{tab:ade20k} We initialize the backbone networks in UperNet \cite{xiao2018unified} with the models pre-trained w/ or w/o our method on ImageNet-1K. $\dagger$: `MS+Flip' indicates the ensemble results of multi-scale and flipped input images.} 
    \end{footnotesize}
    \vskip -0.099in
\end{table}

\begin{table*}[!t]
  \centering
  \begin{footnotesize}
  \begin{subtable}[t]{\linewidth}
    \centering
    \captionsetup{font={footnotesize}}
    \setlength{\tabcolsep}{1.5mm}{
      \renewcommand\arraystretch{1.38}
      \resizebox{0.738\linewidth}{!}{
      \begin{tabular}{cc|ccccccc}
      Low-frequency & Linear\ \  & \ \ Training  & \multicolumn{6}{c}{Top-1 Accuracy{ \textit{(100ep: 100-epoch; Others: 300-epoch)}}} \\[-0.3ex]
       Cropping & RandAug\ \  & \ \ Speedup  & \ DeiT-Tiny\ & DeiT-Small$_{\textnormal{100ep}}$ & DeiT-Small\  & \ Swin-Tiny\ & \ Swin-Small\ & \ CSWin-Tiny\  \\
      \shline
       &  & $1.0\times$                         & 72.5\% & 75.5\% & 80.3\% & 81.3\% & 83.1\% & 82.7\% \\
      \cmark &  & $\bm{\sim\!1.5\times}$        & 72.4\% & 75.5\% & 80.0\%  & 81.1\% & 83.0\% & 82.6\% \\
      \baseline{}\cmark & \baseline{}\cmark & \baseline{}$\bm{\sim\!1.5\times}$    & \baseline{}\textbf{73.3\%} & \baseline{}\textbf{76.4\%} & \baseline{}\textbf{80.4\%} & \baseline{}\textbf{81.4\%} & \baseline{}\textbf{83.2\%} & \baseline{}\hspace{0.3ex}\textbf{82.8\%} \\
      \end{tabular}
    }}
    \vskip -0.05in 
    \caption{\textbf{Ablating low-frequency cropping and linearly increased RandAug.}}
    \end{subtable}
  \begin{subtable}[t]{0.9\linewidth}
    \centering
    \vskip 0.06in 
    \captionsetup{font={footnotesize}}
    \setlength{\tabcolsep}{1.5mm}{
      \renewcommand\arraystretch{1.38}
      \resizebox{0.846\linewidth}{!}{
      \begin{tabular}{c|cccccc}
        \multicolumn{1}{c|}{\multirow{2}{*}{\shortstack{Low-frequency Information Extraction\\[-0.25ex]in \textbf{EfficientTrain}}}} & \ \ Training & \multicolumn{5}{c}{Top-1 Accuracy{ \textit{(100ep: 100-epoch; Others: 300-epoch)}}} \\[-0.3ex]
        & \ \ Speedup  & \ DeiT-Small$_{\textnormal{100ep}}$ & DeiT-Small\  & \ Swin-Tiny\ & \ Swin-Small\ & \ CSWin-Large$^\dagger$\ \\
        \shline
        Standard Image Down-sampling & $\bm{\sim\!1.5\times}$ & 75.9\% & 80.3\% & 81.0\% & 83.0\% & 86.4\%\\
        \baseline{}Low-frequency Cropping\ & \baseline{}$\bm{\sim\!1.5\times}$ & \baseline{}{76.4\%} & \baseline{}\textbf{80.4\%} & \baseline{}\textbf{81.4\%} & \baseline{}\textbf{83.2\%} & \baseline{}{86.6\%} \\
        Efficient Low-frequency Down-sampling$^\ddagger$\ & $\bm{\sim\!1.5\times}$ & \textbf{76.5\%}& \textbf{80.4\%} & \textbf{81.4\%} & {83.1\%} & \textbf{86.7\%} \\
      \end{tabular}
    }}
    \vskip -0.05in 
    \captionsetup{width=0.87\linewidth}
    \caption{\textbf{Design choices of the operation for extracting low-frequency information.} $\dagger$: pre-trained on ImageNet-22K following the configurations in \cite{dong2021cswin, wang2023efficienttrain}. $\ddagger$: mainly proposed to alleviate high CPU-GPU I/O Costs (see: Table \ref{tab:ET_plus_abl}). }
    \end{subtable}
  \begin{subtable}[t]{0.6\linewidth}
    \centering
    \vskip 0.06in
    \captionsetup{font={footnotesize}}
    \setlength{\tabcolsep}{1.5mm}{
      \renewcommand\arraystretch{1.38}
      \resizebox{0.825\linewidth}{!}{
      \begin{tabular}{c|cccc}
        \multicolumn{1}{c|}{\multirow{2}{*}{\shortstack{Schedule of $B$ in \textbf{EfficientTrain}}}} & \ \ Training & \multicolumn{3}{c}{Top-1 Accuracy} \\[-0.3ex]
        & \ \ Speedup  & \ DeiT-Tiny\  & \ DeiT-Small\  & \ Swin-Tiny\ \\
        \shline
        Linear Increasing \cite{tan2021efficientnetv2} & $\bm{\sim\!1.5\times}$ & 72.8\% & 79.9\% & 81.0\% \\
        \baseline{}Obtained from Algorithm \ref{alg:greedy_search}\ & \baseline{}$\bm{\sim\!1.5\times}$ & \baseline{}\textbf{73.3\%} & \baseline{}\textbf{80.4\%} & \baseline{}\textbf{81.4\%} \\
      \end{tabular}
    }}
    \vskip -0.05in 
    \captionsetup{width=0.95\linewidth}
    \caption{\textbf{Schedule of $B$}. For fair comparisons, the linear increasing schedule is configured to have the same training cost as the schedule of EfficientTrain.}
    \end{subtable}
  \end{footnotesize}
  \vskip -0.1in 
  \captionsetup{font={footnotesize}}
  \setcounter{table}{18}
  \captionof{table}{\textbf{Ablation studies of the techniques proposed in \colorbox{baselinecolor}{\!\!{EfficientTrain}\!\!}.} Unless otherwise specified, the results are reported on ImageNet-1K. \label{tab:ablation}}
  \vskip -0.075in 
\end{table*}

\begin{table*}[!t]
  \centering
  \begin{footnotesize}
    \centering
    \captionsetup{font={footnotesize}}
    \setlength{\tabcolsep}{1mm}{
      \renewcommand\arraystretch{1.38}
      \resizebox{0.8865\linewidth}{!}{
      \begin{tabular}{c|cc|cccccc}
      \multicolumn{1}{c|}{\multirow{2}{*}{Method}} & 
      \multirow{2}{*}{Algorithm \ref{alg:greedy_search_v2}} & Efficient Low-frequency & Cost for Solving & Peak CPU-GPU & Training  & \multicolumn{3}{c}{Top-1 Accuracy} \\[-0.4ex]
      & & Down-sampling & for the Curriculum & I/O Cost$^\dagger$ & Speedup  & \ DeiT-Small\  & \ Swin-Tiny\ & \ Swin-Small\ \\
      \shline
      \textbf{EfficientTrain} & & & 1088.1 GPU-hours & 1078.1 M/s & $\bm{\sim\!1.5\times}$                         & 80.4\%  & 81.4\% & 83.2\% \\
      \textbf{EfficientTrain+} &  & \cmark  & 1088.1 GPU-hours & \textbf{550.1 M/s} & $\bm{\sim\!1.5\times}$      & 80.4\%  & 81.4\% & 83.1\% \\
      \textbf{EfficientTrain+} & \cmark &   & \textbf{425.8 GPU-hours} & 3060.1 M/s & $\bm{<\!1.5\times}^\ddagger$        & \textbf{81.0\%}  & \textbf{81.6\%} & \textbf{83.3\%} \\
      \baseline{}\textbf{EfficientTrain++} & \baseline{}\cmark & \baseline{}\cmark  & \baseline{}\textbf{425.8 GPU-hours}\textcolor{blue}{\ \scriptsize{\textbf{($\bm{\downarrow\!2.6\times}$)}}}  & \baseline{}562.1 M/s\textcolor{blue}{\ \scriptsize{\textbf{($\bm{\downarrow\!5.4\times}$)}}} & \baseline{}$\bm{\sim\!1.5\times}$   & \baseline{}\textbf{81.0\%} & \baseline{}\textbf{81.6\%} & \baseline{}{83.2\%}  \\
      \end{tabular}
    }}
  \vskip -0.075in
  \captionsetup{font={footnotesize}}
  \caption{\textbf{Ablation studies of the improvements of \colorbox{baselinecolor}{\!\!{EfficientTrain++}\!\!} over EfficientTrain} (on ImageNet-1K). $\dagger$: as a representative example, we report the results corresponding to training DeiT-Small, where the value of I/O is obtained by saturating all the computational cores of a single NVIDIA 3090 GPU. $\ddagger$: the practical training efficiency is usually lower than the theoretical results due to high CPU-GPU I/O costs.
  \label{tab:ET_plus_abl}}
  \end{footnotesize}
  \vskip -0.075in 
\end{table*}

\begin{table*}[!t]
  \centering
  \begin{footnotesize}
    \centering
    \captionsetup{font={footnotesize}}
    \setlength{\tabcolsep}{3mm}{
      \renewcommand\arraystretch{1.38}
      \resizebox{0.87\linewidth}{!}{
      \begin{tabular}{c|ccc|ccc}
      \multicolumn{1}{c|}{\multirow{3}{*}{\shortstack{Searching Configurations$^\dagger$}}} & \multicolumn{2}{c}{Variables} & \multirow{2}{*}{Optimization Objective\ \ \ \ } & \multicolumn{3}{c}{Effectiveness of the Resulting Curriculum} \\[-0.5ex]
      & Constant & Changeable & & Training  & \multicolumn{2}{c}{Top-1 Accuracy} \\[-0.5ex]
      & \multicolumn{3}{c|}{\scriptsize\textit{(see caption for the meaning of notations)}} & Speedup$^\ddagger$ & \ DeiT-Small\  & \ Swin-Tiny\ \\
      \shline
      Following AutoProg \cite{li2022automated} & $I_i$ & $B_i$, $\mathcal{T}_i$ & \multicolumn{1}{l|}{minimize\ \ $\mathcal{L}_i \cdot \mathcal{T}_i^{\alpha}${\tiny ($\alpha$: hyper-parameter)}} & $\bm{\sim\!1.5\times}$ & 80.2\% & 81.1\% \\
      -- & $I_i$ & $B_i$, $\mathcal{T}_i$ & \multicolumn{1}{l|}{minimize\ \ $(1 - \textnormal{Acc}_i) \cdot \mathcal{T}_i^{\alpha}${\tiny ($\alpha$: hyper-parameter)}\!\!\!\!} & $\bm{\sim\!1.5\times}$  & 80.3\% & 81.3\% \\
      \baseline{}Algorithm \ref{alg:greedy_search_v2} & \baseline{}$\mathcal{T}_i$ & \baseline{}$B_i$, $I_i$ & \multicolumn{1}{l|}{\baseline{}maximize\ \ $\textnormal{Acc}_i$} & \baseline{}$\bm{\sim\!1.5\times}$  & \baseline{}\textbf{81.0\%} & \baseline{}\textbf{81.6\%}  \\
      \end{tabular}
    }}
  \vskip -0.05in
  \captionsetup{font={footnotesize}}
  \caption{\textbf{Ablation studies of the design choices of Algorithm \ref{alg:greedy_search_v2}}. We modify the searching configurations of Algorithm \ref{alg:greedy_search_v2} (\emph{e.g.}, configs of variables and optimization objectives), and solve for curricula by leveraging the same procedure as obtaining EfficientTrain++. Then we compare the resulting curricula to investigate the effectiveness of different searching configurations. In `Variables', $B_i$, $I_i$, and $\mathcal{T}_i$ refer to the bandwidth of efficient low-frequency down-sampling, number of training iterations, and computational training cost corresponding to $i^{\textnormal{th}}$ training stage, while $i$ can be an arbitrary index. In `Optimization Objective', $\mathcal{L}_i$ and $\textnormal{Acc}_i$ denote training loss and validation accuracy, respectively. $\dagger$: all variants considered here have similar searching cost. $\ddagger$: for clarity, different curricula are compared under the same training cost.
  \label{tab:design_of_alg2}}
  \end{footnotesize}
  \vskip -0.15in
\end{table*}

 \subsection{Discussions}
 \label{sec:discussion}

\subsubsection{Ablation Study}

To get a better understanding of our proposed approach, we conduct a series of ablation studies. In specific, following the organization of this paper, we first investigate the effectiveness of the components of EfficientTrain, and then study whether the new techniques proposed in EfficientTrain++ improve EfficientTrain as we expect.

\textbf{Ablation study results with EfficientTrain}
are summarized in Table \ref{tab:ablation}.
In Table \ref{tab:ablation} (a), we show that the major gain of training efficiency comes from low-frequency cropping, which effectively reduces the training cost at the price of a slight drop of accuracy. On top of this, linear RandAug further improves the accuracy. Moreover, replacing low-frequency cropping with image down-sampling consistently degrades the accuracy (see: Table \ref{tab:ablation} (b)), since down-sampling cannot strictly filter out all the higher-frequency information (see: Proposition \ref{prop:downsampling}), yielding a sub-optimal implementation of our idea. In addition, as shown in Table \ref{tab:ablation} (c), the schedule of $B$ found by Algorithm \ref{alg:greedy_search} outperforms the heuristic design choices (\emph{e.g.}, the linear schedule in \cite{tan2021efficientnetv2}).

\begin{table*}[!t]
  \centering
  \begin{footnotesize}
  \begin{subtable}[t]{0.312\linewidth}
    \centering
    \captionsetup{font={footnotesize}}
    \setlength{\tabcolsep}{0.5mm}{
      \renewcommand\arraystretch{1.38}
      \resizebox{\linewidth}{!}{
      \begin{tabular}{l|cc}
      \multicolumn{1}{c|}{\multirow{2}{*}{Method}} & Training  & IN-1K \\[-0.6ex]
       & Speedup  & Accuracy \\
      \shline
      Baseline, DeiT-S$_{\textnormal{IN-1K}}$ & $1.0\times$ & 80.3\% \\
      \baseline{}\textbf{ET++}, Swin-T$_{\textnormal{IN-1K}}$ $\to$ DeiT-S$_{\textnormal{IN-1K}}$ & \baseline{}\ \!$\bm{1.6\times}$ & \baseline{}81.0\% \\
      \textbf{ET++}, DeiT-S$_{\textnormal{IN-1K}}$ $\to$ DeiT-S$_{\textnormal{IN-1K}}$ & \ \!$\bm{1.6\times}$ & \textbf{81.2\%} \\
      \end{tabular}
    }}
    \vskip -0.05in 
    \caption{\textbf{Multi-stage ViTs $\to$ isotropic ViTs.}}
    \end{subtable}
    \begin{subtable}[t]{0.34\linewidth}
      \centering
      \captionsetup{font={footnotesize}}
      \setlength{\tabcolsep}{0.5mm}{
        \renewcommand\arraystretch{1.38}
        \resizebox{\linewidth}{!}{
        \begin{tabular}{l|cc}
        \multicolumn{1}{c|}{\multirow{2}{*}{Method}} & Training  & IN-1K \\[-0.6ex]
         & Speedup  & Accuracy \\
        \shline
        Baseline, CSWin-T$_{\textnormal{IN-1K}}$ & $1.0\times$ & 82.7\% \\
        \baseline{}\textbf{ET++}, Swin-T$_{\textnormal{IN-1K}}$ $\to$ CSWin-T$_{\textnormal{IN-1K}}$ & \baseline{}\ \!$\bm{1.5\times}$ & \baseline{}\textbf{82.9\%} \\
        \textbf{ET++}, CSWin-T$_{\textnormal{IN-1K}}$ $\to$ CSWin-T$_{\textnormal{IN-1K}}$ & \ \!$\bm{1.5\times}$ & \textbf{82.9\%} \\
        \end{tabular}
      }}
      \vskip -0.05in 
      \caption{\textbf{Swin $\to$ advanced multi-stage ViTs.}}
      \end{subtable}
      \begin{subtable}[t]{0.34\linewidth}
        \centering
        \captionsetup{font={footnotesize}}
        \setlength{\tabcolsep}{0.5mm}{
          \renewcommand\arraystretch{1.38}
          \resizebox{\linewidth}{!}{
          \begin{tabular}{l|cc}
          \multicolumn{1}{c|}{\multirow{2}{*}{Method}} & Training  & IN-1K \\[-0.6ex]
           & Speedup  & Accuracy \\
          \shline
          Baseline, Swin-S$_{\textnormal{IN-1K}}$ & $1.0\times$ & 83.1\% \\
          \baseline{}\textbf{ET++}, Swin-T$_{\textnormal{IN-1K}}$ $\to$ Swin-S$_{\textnormal{IN-1K}}$ & \baseline{}\ \!$\bm{1.5\times}$ & \baseline{}\textbf{83.2\%} \\
          \textbf{ET++}, Swin-S$_{\textnormal{IN-1K}}$ $\to$ Swin-S$_{\textnormal{IN-1K}}$ & \ \!$\bm{1.5\times}$ & \textbf{83.2\%} \\
          \hline
          Baseline, CSWin-S$_{\textnormal{IN-1K}}$ & $1.0\times$ & 83.4\% \\
          \baseline{}\textbf{ET++}, CSWin-T$_{\textnormal{IN-1K}}$ $\to$ CSWin-S$_{\textnormal{IN-1K}}$ & \baseline{}\ \!$\bm{1.5\times}$ & \baseline{}83.6\% \\
          \textbf{ET++}, CSWin-S$_{\textnormal{IN-1K}}$ $\to$ CSWin-S$_{\textnormal{IN-1K}}$ & \ \!$\bm{1.5\times}$ & \textbf{83.7\%} \\
          \end{tabular}
        }}
        \vskip -0.05in 
        \caption{\textbf{Model size: smaller $\to$ larger.}}
        \end{subtable}

        \vskip -0.07in 
      \begin{subtable}[t]{0.39\linewidth}
        \centering
        \captionsetup{font={footnotesize}}
        \setlength{\tabcolsep}{0.5mm}{
          \renewcommand\arraystretch{1.38}
          \resizebox{0.96\linewidth}{!}{
          \begin{tabular}{l|cc}
          \multicolumn{1}{c|}{\multirow{2}{*}{Method}} & Pre-training  & IN-1K \\[-0.6ex]
            & Speedup  & Accuracy \\
          \shline
          Baseline, CSWin-B (IN-22K pre-training) & $1.0\times$ & 86.0\% \\
          \baseline{} 
          & \baseline{}${2.0\times}$ & \baseline{}\textbf{86.3\%} \\
          \baseline{}\multirow{-2}{*}{\textbf{ET++}, Swin-T$_{\textnormal{IN-1K}}$ $\to$ CSWin-B$_{\textnormal{IN-22K}}$} 
          & \baseline{}${3.0\times}$ & \baseline{}86.1\% \\
          \multirow{2}{*}{\textbf{ET++}, CSWin-B$_{\textnormal{IN-22K}}$ $\to$ CSWin-B$_{\textnormal{IN-22K}}$} 
          & ${3.0\times}$ & \textbf{86.3\%} \\
          & \ \!$\bm{3.8\times}$ & 86.0\% \\
          \end{tabular}
        }}
        \vskip -0.05in 
        \caption{\textbf{Small models + IN-1K$\to$ larger models + IN-22K.}}
        \end{subtable}
      \hskip -0.085in
      \begin{subtable}[t]{0.6175\linewidth}
        \centering
        \captionsetup{font={footnotesize}}
        \setlength{\tabcolsep}{0.5mm}{
          \renewcommand\arraystretch{1.38}
          \resizebox{\linewidth}{!}{
          \begin{tabular}{l|ccccc}
          \multicolumn{1}{c|}{\multirow{2}{*}{Method}} & Training  & \multicolumn{4}{c}{Accuracy on Different Dataset} \\[-0.6ex]
            & Speedup  & Food-101 & NABirds & CIFAR-100 & Stanford Dogs \\
          \shline
          Baseline, Swin-T (trained on specific dataset) & $1.0\times$ & 87.9\% & 49.7\% & 74.4\% & 54.3\% \\
          \baseline{}\textbf{ET++}, Swin-T (IN-1K $\to$ specific dataset) & \baseline{}\ \!$\bm{1.5\times}$ & \baseline{}89.1\% & \baseline{}59.4\% & \baseline{}81.8\% & \baseline{}61.0\% \\
          \textbf{ET++}, Swin-T (specific dataset $\to$ specific dataset) & \ \!$\bm{1.5\times}$ & \textbf{89.3\%} & \textbf{59.6\%} & \textbf{81.9\%} & \textbf{61.3\%} \\
          \end{tabular}
        }}
        \vskip -0.075in 
        \caption{\textbf{ImageNet-1K $\to$ different datasets.}}
        \end{subtable}

  \end{footnotesize}

  \vskip -0.08in 
  \captionsetup{font={footnotesize}}
  \setcounter{table}{21}
  \captionof{table}{\textbf{Investigation on the transferability of EfficientTrain++ (ET++).} The left side of `$\to$' indicates the model and dataset with which we solve for the learning curriculum, while the right side indicates the configurations with which we implement the curriculum to train the model. All the curricula are obtained using Algorithm \ref{alg:greedy_search_v2}. IN-1K/22K denotes ImageNet-1K/22K. Our default settings are marked in \colorbox{baselinecolor}{\!\!{gray}\!\!}. \label{tab:alg2_generalizability}}
  \vskip -0.165in 
\end{table*}

\textbf{Improvements of EfficientTrain++ over EfficientTrain.}
In Table \ref{tab:ET_plus_abl}, we study the effectiveness of the improved techniques proposed in EfficientTrain++ (\emph{i.e.}, the components introduced in Sections \ref{sec:ET_plus_1} and \ref{sec:ET_plus_2}). One can observe that Algorithm \ref{alg:greedy_search_v2} dramatically reduces the cost for obtaining the training curriculum. Moreover, the models trained using the curriculum found by Algorithm \ref{alg:greedy_search_v2} are able to acquire a stronger generalization performance with the same training cost. On its basis, further introducing the efficient low-frequency down-sampling operation alleviates the problem of high CPU-GPU I/O costs, and thus effectively improves the practical training efficiency of our method.

\textbf{Design of Algorithm \ref{alg:greedy_search_v2}.}
We further study whether the design of Algorithm \ref{alg:greedy_search_v2} is reasonable. Specifically, we select the searching configurations of AutoProg \cite{li2022automated} as a competitive baseline. AutoProg is a recently proposed efficient training approach for ViTs. It is similar to Algorithm \ref{alg:greedy_search_v2} in the paradigm of fine-tuning the model at each training stage to search for the proper progressive learning strategy. However, both its optimization objectives and its configurations of variables are different from our method. In our experiments, we first implement the searching configurations of AutoProg in our problem. Then, we introduce modifications on top of AutoProg step by step and finally obtain Algorithm \ref{alg:greedy_search_v2}, aiming to demonstrate our advantages over the current state-of-the-art design. The results are summarized in Table \ref{tab:design_of_alg2}, where two observations can be obtained: 1) validation accuracy is a better searching objective than training loss in our problem; and 2) our proposed formulation of computational-constrained searching (\emph{i.e.}, maximizing validation accuracy under a fixed ratio of training cost saving; see Section \ref{sec:ET_plus_1} for details) results in a significantly better training curriculum, while it can simplify the searching objective by eliminating the hyper-parameter for balancing the effectiveness-efficiency trade-off.

\subsubsection{Transferability of EfficientTrain++}

Table \ref{tab:alg2_generalizability} systematically discusses the generalization ability of our EfficientTrain++ curriculum (\emph{i.e.}, Table \ref{tab:EfficientTrain_plus}, obtained on Swin-Tiny and ImageNet-1K). We execute Algorithm \ref{alg:greedy_search_v2} on top of different visual backbone architectures and diverse datasets. Then, in each corresponding scenario, we compare the resulting specialized curriculum with EfficientTrain++. 


\textbf{Different networks.}
Like many other useful training techniques (\emph{e.g.}, cosine learning rate schedule, AdamW optimizer, and random erasing), the effectiveness of EfficientTrain++ is generally not tied to a specific network architecture. For example, Table \ref{tab:alg2_generalizability} (b) \& (c) reveal that the curriculum optimized on Swin/CSWin-Tiny is still close to the optimal design (found by Algorithm \ref{alg:greedy_search_v2}) if we adopt advanced architectures within the family of multi-stage ViTs or enlarge model size. Besides, as shown in Table \ref{tab:alg2_generalizability} (a), when applying standard EfficientTrain++ to another family of networks (\emph{e.g.}, isotropic ViTs), model-specific curriculum searching can yield minor improvements in training efficiency (\emph{e.g.}, 81.0\% $\!\to\!$ 81.2\% for DeiT-Small), while this comes at the notable price of executing Algorithm \ref{alg:greedy_search_v2} once again. In contrast, directly utilizing standard EfficientTrain++ introduces no additional cost, and it can already acquire the vast majority of gains for efficient training.



\textbf{Different datasets.}
When applying our method to a new dataset, two facts are noteworthy. 
\emph{First}, based on the aforementioned robustness across different networks, one can solve for a generalizable curriculum on top of a common, small model (\emph{e.g.}, Swin-Tiny). According to Table \ref{tab:ET_vs_ETplus}, the searching cost is approximately equal to training this small model for 2$\sim$3 times.
\emph{Second}, in many cases, it may not be necessary to conduct searching on the new dataset (or on the full new dataset).
For example, Table \ref{tab:alg2_generalizability} (d) shows that when increasing training data scale from ImageNet-1K to ImageNet-22K (by $\sim\!10\times$), EfficientTrain++ can already acquire the majority of gains for efficient training (lossless training speedup: ${3.0\times}$ v.s. ${3.8\times}$). Hence, it is reasonable to solve for the curriculum on a smaller subset of the full large-scale dataset (\emph{e.g.}, with $\sim$1/10 data), unless one wants to maximize training efficacy regardless of cost. 
Moreover, as shown in Table \ref{tab:alg2_generalizability} (e), on four widely-used datasets (\emph{i.e.}, Food-101 \cite{bossard2014food}, NABirds \cite{van2015building}, CIFAR \cite{krizhevsky2009learning}, and Stanford Dogs \cite{khosla2011novel}), the ImageNet-1K-based curriculum performs comparably with dataset-specific curriculum optimization. This may be attributed to the fact that accomplishing the task of learning representations on a challenging large-scale natural image database like ImageNet has already covered the needs of many specialized visual scenarios. Therefore, on many common datasets, one may simply adopt an off-the-shelf curriculum obtained on comprehensive natural image datasets like ImageNet-1K (\emph{e.g.}, EfficientTrain++).




\begin{table}[!t]
  \centering
  \vskip -0.05in
  \begin{footnotesize}
  \setlength{\tabcolsep}{1.5mm}{
  \renewcommand\arraystretch{1.33}
  \resizebox{0.875\linewidth}{!}{
  \begin{tabular}{clcc}
  \multicolumn{1}{c}{\multirow{2}{*}{Models}}  & \multicolumn{1}{c}{\multirow{2}{*}{Training Approach}} & Training  & {ImageNet-1K} \\[-0.6ex]
       &  & Speedup & Accuracy  \\
  \shline
  \multirow{4}{*}{ResNet-50}  
  & Baseline & $1.0\times$ & 78.8\%  \\
  & InfoBatch \cite{qin2024infobatch} \textcolor{gray}{\tiny{(\textit{ICLR'24})}} & $1.4\times$ & 78.6\%  \\
  & \baseline{}\textbf{EfficientTrain++} & \baseline{}${1.5\times}$ & \baseline{}\textbf{79.6\%}  \\
  & \baseline{}\textbf{EfficientTrain++} + InfoBatch & \baseline{}\ \!$\bm{1.7\times}$ & \baseline{}\textbf{79.6\%}  \\
  \end{tabular}}}
  \vskip -0.075in
  \captionsetup{font={footnotesize}}
  \caption{\textbf{Compatibility with sample-selection-based methods.} 
  \label{tab:compatibility_with_sample_selection}}
  \end{footnotesize}
  \vskip -0.15in
\end{table}

\subsubsection{Incorporating Sample-selection-based Methods}

As a flexible framework, our proposed generalized curriculum learning can easily incorporate previous sample-selection curriculum learning methods. In this paper, these methods are mainly considered as the baselines to compare with, aiming to elaborate on our new contributions. 
However, the idea of data-selection is actually compatible with us. It is straightforward to consider a curriculum learning function that simultaneously uncovers progressively more difficult patterns within each example and determines whether each example should be leveraged for training. To make this clear, we provide a representative example in Table \ref{tab:compatibility_with_sample_selection}, where sample-selection is attained with the recently proposed InfoBatch algorithm \cite{qin2024infobatch}. It can be observed that the combination of EfficientTrain++ and InfoBatch further improves training efficiency compared to either individual method, while the effect of EfficientTrain++ is more significant, which is consistent with the findings of our paper.

\section{Conclusion}

This paper aimed to develop a simple and broadly applicable efficient learning algorithm for modern deep networks with the potential for widespread implementation. 
Under this goal, we proposed \emph{EfficientTrain++}, a novel generalized curriculum learning approach that always leverages all the data at every learning stage, but only exposes the `easier-to-learn' patterns of each example at the beginning of training (\emph{e.g.}, lower-frequency components of images and original information before data augmentation), and gradually introduces more difficult patterns as learning progresses. To design a proper curriculum learning schedule, we proposed a tailored computational-constrained sequential searching algorithm, yielding a simple, well-generalizable, yet surprisingly effective training curriculum. The effectiveness of \emph{EfficientTrain++} is extensively validated on the large-scale ImageNet-1K/22K datasets, on top of various state-of-the-art deep networks (\emph{e.g.}, ConvNets, ViTs, and ConvNet-ViT hybrid models), under diverse training settings (\emph{e.g.}, different training budgets, varying test input sizes, and supervised/self-supervised learning scenarios). Without sacrificing model performance or introducing new tunable hyper-parameters, \emph{EfficientTrain++} reduces the wall-time training cost of visual backbones by $\bm{1.5\!-\!3.0\times}$. 

Potentially, our work may open new avenues for developing efficient curriculum learning approaches for vision models. For example, future work may explore more general forms of the transformation function for extracting `easier-to-learn' patterns, \emph{e.g.}, formulating a learnable neural network under principles like meta-learning, or considering differentiating between the patterns within `easier-to-learn' and `harder-to-learn' spatial-temporal regions of vision data. Moreover, it would be interesting to investigate how to uncover progressively more difficult patterns during training for more general data categories, \emph{e.g.}, for tabular data, we may consider masking out some features (columns) to only expose certain `easier-to-learn' relationships between different features at earlier training stages.

\ifCLASSOPTIONcompsoc
  \section*{Acknowledgments}
\else
  \section*{Acknowledgment}
\fi

This work was supported in part by the National Key R\&D Program of China under Grant 2021ZD0140407, in part by the National Natural Science Foundation of China under Grants 62022048 and 62276150, and in part by the Guoqiang Institute of Tsinghua University.
\ifCLASSOPTIONcaptionsoff
  \newpage
\fi


\bibliographystyle{IEEEtran}
\bibliography{IEEEtran}


\clearpage
\onecolumn

\appendices

{\centering\section*{Appendix for\\``EfficientTrain++: Generalized Curriculum\\Learning for Efficient Visual Backbone Training''}}

\section{Implementation Details}
\label{app:implementation_details}

\subsection{Training Models on ImageNet-1K}

\textbf{Dataset.}
We use the data provided by ILSVRC2012\footnote{\url{https://image-net.org/index.php}} \cite{deng2009imagenet}. The dataset includes 1.2 million images for training and 50,000 images for validation, both of which are categorized in 1,000 classes.

\textbf{Training.}
Our approach is developed on top of a state-of-the-art training pipeline of deep networks, which incorporates a holistic combination of various model regularization \& data augmentation techniques, and is widely applied to train recently proposed models \cite{touvron2021training, wang2021pyramid, liu2021swin, dong2021cswin, liu2022convnet, yu2022metaformer}. Our training settings for the standard case (300-epoch training for the baselines and EfficientTrain; 200-epoch training for EfficientTrain++) are summarized in Table \ref{tab:img_1k_details}, which generally follow from \cite{liu2022convnet}, except for the following differences. 
We modify the configurations of weight decay, stochastic depth and exponential moving average (EMA) according to the recommendation in the original papers of different models (\emph{i.e.}, ConvNeXt \cite{liu2022convnet}, DeiT \cite{touvron2021training}, PVT \cite{wang2021pyramid}, Swin Transformer \cite{liu2021swin}, CSWin Transformer \cite{dong2021cswin}, and CAFormer \cite{yu2022metaformer})\footnote{The training of ResNet \cite{He_2016_CVPR} follows the recipe provided in \cite{liu2022convnet}.}. In addition, following \cite{yu2022metaformer}, the training of CSWin-Base \cite{dong2021cswin} and CAFormer \cite{yu2022metaformer} adopts the LAMB optimizer \cite{You2020Large} with an initial learning rate of 8e-3, which contributes to a stable learning procedure for these models with the batch size of 4096. 

When we reduce the training cost on the basis of the standard case (300-epoch training for the baselines; 200-epoch training for EfficientTrain++), we linearly reduce the maximum value of the increasing stochastic depth regularization simultaneously. This configuration is motivated by that relatively less computational budgets for training require weaker regularization. Notably, both baselines and our method adopt this setting, which consistently improves the accuracy.





On top of the baselines, EfficientTrain and EfficientTrain++ only modify the terms mentioned in Tables \ref{tab:EfficientTrain} and \ref{tab:EfficientTrain_plus}, respectively. The only exception is, when training ConvNeXts, we replace $B=96$ with $B=160$ in EfficientTrain++, as $B=96$ will make some parameters of the $7\!\times\!7$ convolution kernels within the last network stage of ConvNeXts have zero gradients. We believe that this straightforward rule for adjustment does not degrade the effectiveness of EfficientTrain++ or affect the major contributions of our method. In addition, note that the low-frequency cropping operation in our method leads to a varying input size during training. On this issue, visual backbones can naturally process different sizes of inputs with no or minimal modifications. Specifically, once the input size varies, ResNets and ConvNeXts do not need any change, while vision Transformers (\emph{i.e.}, DeiT, PVT, Swin, CSWin, an CAFormer) only need to resize their position bias correspondingly, as suggested in their papers. Our method starts the training with small-size inputs and the reduced computational cost. The input size is switched midway in the training process, where we resize the position bias for ViTs (do nothing for ConvNets). Finally, the learning ends up with full-size inputs, as used at test time. As a consequence, the overall computational/time cost to obtain the final trained models is effectively saved.

\begin{table}[h]
  \vskip -0.1in
  \centering
  \begin{footnotesize}
  \setlength{\tabcolsep}{4mm}{
  \vspace{5pt}
  \renewcommand\arraystretch{1.175}
  \resizebox{0.8\columnwidth}{!}{
  \begin{tabular}{l|c}
  Training Config & Values / Setups \\
  \shline
  Input size & 224$^2$ \\
  Weight init. & Truncated normal (0.2) \\
  Optimizer & AdamW \\
  Optimizer hyper-parameters & $\beta_1, \beta_2$=0.9, 0.999 \\
  Initial learning rate & 4e-3 \\
  Learning rate schedule & Cosine annealing \\
  Weight decay & 0.05 \\
  Batch size & 4,096 \\
  Training epochs & 300 \\
  Warmup epochs & 20 \\
  Warmup schedule & linear \\
  RandAug \cite{cubuk2020randaugment} &  (9, 0.5) \\
  Mixup \cite{zhang2018mixup} & 0.8 \\
  Cutmix \cite{yun2019cutmix} & 1.0 \\
  Random erasing \cite{zhong2020random} & 0.25 \\
  Label smoothing \cite{szegedy2016rethinking} & 0.1 \\
  Stochastic depth \cite{huang2016deep} & Following the configurations in original papers \cite{liu2022convnet, touvron2021training, wang2021pyramid, liu2021swin, dong2021cswin, yu2022metaformer}. \\
  Layer scale \cite{touvron2021going} & 1e-6 \scriptsize (ConvNeXt \cite{liu2022convnet}) \footnotesize/ None \scriptsize (others) \footnotesize\\
  Gradient clip & 5.0 \scriptsize (DeiT \cite{touvron2021training}, PVT \cite{wang2021pyramid} and Swin \cite{liu2021swin}) \footnotesize / None \scriptsize (others) \footnotesize \\
  Exp. mov. avg. (EMA) \cite{polyak1992acceleration} & Following the configurations in original papers \cite{liu2022convnet, touvron2021training, wang2021pyramid, liu2021swin, dong2021cswin, yu2022metaformer}. \\
  Auto. mix. prec. (AMP) \cite{micikevicius2018mixed} & Inactivated \scriptsize (ConvNeXt, following \cite{liu2022convnet}) \footnotesize / Activated \scriptsize (others) \footnotesize \\
  \end{tabular}}}
  \end{footnotesize}
  \vskip -0.075in
  \caption{\textbf{Basic hyper-parameters and configurations for training models on ImageNet-1K}}
  \label{tab:img_1k_details}
  \vskip -0.1in
\end{table}

\textbf{Inference.}
Following \cite{touvron2021training, wang2021pyramid, liu2021swin, dong2021cswin, liu2022convnet}, we use a crop ratio of 0.875 and 1.0 for the inference input size of 224$^2$ and 384$^2$, respectively.

\subsection{ImageNet-22K Pre-training}

\textbf{Dataset.} 
The officially released ImageNet-22K dataset \cite{deng2009imagenet, ridnik2021imagenet} from \url{https://www.image-net.org/} is used. Our experiments are based on the latest `Winter 2021 release' version.


\textbf{Pre-training.} 
We pre-train ConvNeXt-Base/Large and CSWin-Base/Large on ImageNet-22K. The pre-training process basically follows the training configurations of ImageNet-1K (\emph{i.e.}, Table \ref{tab:img_1k_details}), except for the differences presented in the following. The basic number of training epochs is set to 150 with a 5-epoch learning rate warm-up. The maximum value of the increasing stochastic depth regularization \cite{huang2016deep} is set to 0.1/0.1 for ConvNeXt-Base/Large and 0.2/0.5 for CSWin-Base/Large. Following \cite{dong2021cswin, yu2022metaformer}, the initial learning rate for CSWin-Base/Large is set to 2e-3, while the weight-decay coefficient for CSWin-Base/Large is set to 0.05/0.1. Following \cite{liu2022convnet}, we do not leverage the exponential moving average (EMA) mechanism. To ensure a fair comparison, we report the results of our implementation for both baselines and our method, where they adopt exactly the same training settings (apart from the configurations modified by EfficientTrain/EfficientTrain++ itself).


\textbf{Fine-tuning.} 
We evaluate the ImageNet-22K pre-trained models by fine-tuning them and reporting the corresponding accuracy on ImageNet-1K. The fine-tuning process follows from \cite{liu2022convnet}. In specific, we directly utilize the basic configurations in \cite{liu2022convnet}, and tune the rate of layer-wise lr decay, stochastic depth regularization, and exponential moving average (EMA) conditioned on each model, as suggested by \cite{liu2022convnet}. It is worth noting that the baselines and our method adopt exactly the same fine-tuning settings.



\subsection{Object Detection and Segmentation on COCO}

Our implementation of RetinaNet \cite{lin2017focal} follows from \cite{xia2022vision}. Our implementation of Cascade Mask-RCNN \cite{cai2019cascade} is the same as \cite{liu2022convnet}.

\subsection{Experiments in Section \ref{sec:EfficientTrain_sec4}}

In particular, the experimental results provided in Section \ref{sec:EfficientTrain_sec4} are based on the training settings listed in Table \ref{tab:img_1k_details} as well, expect for the specified modifications (\emph{e.g.}, with the low-passed filtered inputs). The computing of CKA feature similarity follows from \cite{raghu2021vision}.

\newpage

\section{Additional Results}
\label{app:additional_result}

\subsection{Training Curves}


\begin{wrapfigure}{r}{0.25\linewidth} 
  \begin{center}
    \vspace{-13ex}
    \hskip -0.125in
    \begin{minipage}{1\linewidth}
      \includegraphics[width=\linewidth]{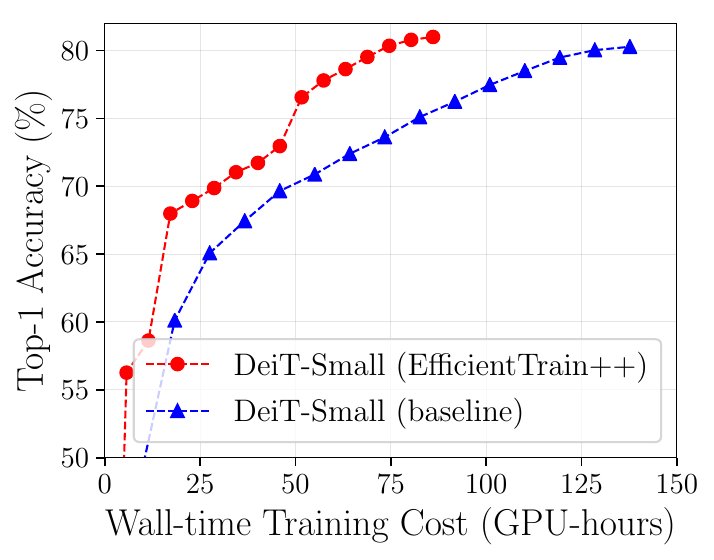}
    \vskip -0.15in
    \captionsetup{font={footnotesize}}
    \caption{\label{fig:training_curve}\textbf{Val. accuracy during training}.}
    \end{minipage}
    \vspace{-6ex}
  \end{center}
\end{wrapfigure}
\textbf{Curves of accuracy during training}
are presented in Figure \ref{fig:training_curve}. The horizontal axis denotes the wall-time training cost. The low-frequency cropping in our method is performed on both the training and test inputs. It is clear that our method learns discriminative representations more efficiently at earlier epochs.

\subsection{Results of EfficientTrain with Varying Epochs}

EfficientTrain can conveniently adapt to a varying number of computational training budgets,  \emph{i.e.}, by simply scaling the indices of epochs in Table \ref{tab:EfficientTrain}. As shown in Table \ref{tab:ETv1_vary_epoch}, the advantage of EfficientTrain is even more significant with fewer training epochs, \emph{e.g.}, it outperforms the baseline by 0.9\% (76.4\% v.s. 75.5\%) for the 100-epoch trained DeiT-Small (the speedup is the same as 300-epoch). We attribute this to the greater importance of efficient training algorithms in the scenarios of limited training resources.

Notably, these results are originally reported in the conference version of this paper. We do not directly present them in Figure \ref{fig:acc_vs_cost}, since the experimental settings here are slightly different from Figure \ref{fig:acc_vs_cost}: here the configuration of stochastic depth regularization remains unchanged for all training budgets. In contrast, in Figure \ref{fig:acc_vs_cost}, we replace this straightforward setting with a more appropriate one: the maximum value of the increasing stochastic depth regularization is linearly decreased together with the reduction of training budgets, motivated by that fewer training budgets may require weaker regularization. Our new setting significantly improves the validation accuracy \emph{for both baselines and our method}, making our experimental results more reasonable. However, we believe that Table \ref{tab:ETv1_vary_epoch} may be sufficient to support the following conclusion: EfficientTrain can conveniently adapt to different numbers of training epochs (\emph{i.e.}, varying training budgets) utilizing simple linear scaling, which is the major aim of Table \ref{tab:ETv1_vary_epoch}.

\begin{table}[!h]
  \centering
  \begin{footnotesize}
    \setlength{\tabcolsep}{0.2mm}{
    \renewcommand\arraystretch{1.4}
    \resizebox{0.6\linewidth}{!}{
    \begin{tabular}{lcccccccc}
    \multicolumn{1}{c}{\multirow{2}{*}{Models}} & {\!\!Input Size\!\!} & \multicolumn{6}{c}{Top-1 Accuracy (baseline / \colorbox{baselinecolor}{\!\!\textbf{EfficientTrain}\!\!})} & \!\!Wall-time Tra- \\[-0.55ex]
    & \!\!(inference)\!\! & \multicolumn{2}{c}{\ \ 100 epochs} & \multicolumn{2}{c}{\ \ 200 epochs} & \multicolumn{2}{c}{\ \ 300 epochs} & \!\!ining Speedup  \\
    \shline
    DeiT-Tiny \cite{touvron2021training} & 224$^2$ & 65.8\% \!/&\baseline{}\!\textbf{68.1\%}\! & \ \ \ 70.5\% \!/&\baseline{}\!\ \textbf{71.8\%} & \ \ \ 72.5\% \!/&\baseline{}\!\ \textbf{73.3\%} & ${1.55\times}$  \\
    DeiT-Small \cite{touvron2021training} & 224$^2$ & 75.5\% \!/&\baseline{}\!\ \textbf{76.4\%} & \ \ \ 79.0\% \!/&\baseline{}\!\ \textbf{79.1\%} & \ \ \ 80.3\% \!/&\baseline{}\!\ \textbf{80.4\%} & ${1.51\times}$  \\
    \hline
    Swin-Tiny \cite{liu2021swin} & 224$^2$ & 78.4\% \!/&\baseline{}\!\ \textbf{78.5\%} & \ \ \ 80.6\% \!/&\baseline{}\!\ \textbf{80.6\%} & \ \ \ 81.3\% \!/&\baseline{}\!\ \textbf{81.4\%} & ${1.49\times}$ \\
    Swin-Small \cite{liu2021swin} & 224$^2$ & 80.6\% \!/&\baseline{}\!\ \textbf{80.7\%} & \ \ \ \textbf{82.7\%} \!/&\baseline{}\!\ 82.6\% & \ \ \ 83.1\% \!/&\baseline{}\!\ \textbf{83.2\%} & ${1.50\times}$ \\
    Swin-Base \cite{liu2021swin} & 224$^2$ & 80.7\% \!/&\baseline{}\!\ \textbf{81.1\%} & \ \ \ 83.2\% \!/&\baseline{}\!\ \textbf{83.2\%} & \ \ \ 83.4\% \!/&\baseline{}\!\ \textbf{83.6\%} & ${1.50\times}$ \\
    \end{tabular}}}
  \end{footnotesize}
  \captionsetup{font={footnotesize}}
  \vskip -0.05in
  \caption{\textbf{Comparison of EfficientTrain and the baselines on top of varying training epochs}. EfficientTrain can easily adapt to different training epochs utilizing simple linear scaling. Compared to the baselines, our method reduces the training cost effectively with the same number of epochs, while achieving competitive or better accuracy.
  \label{tab:ETv1_vary_epoch}
  }
\end{table}

\newpage

\section{Proof of Proposition \ref{prop:downsampling}}
\label{app:proof}
In this section, we theoretically demonstrate the difference between two transformations, namely low-frequency cropping and image down-sampling. In specific, we will show that from the perspective of signal processing, the former perfectly preserves the lower-frequency signals within a square region in the frequency domain and discards the rest, while the image obtained from pixel-space down-sampling contains the signals mixed from both lower- and higher- frequencies. 

\subsection{Preliminaries}
\label{app:preliminary}
An image can be seen as a high-dimensional data point $\boldsymbol{X} \in\mathbb{R}^{C_0 \times H_0 \times W_0}$, where $C_0, H_0, W_0$ represent the number of channels, height and width. Since each channel's signals are regarded as independent, for the sake of simplicity, we can focus on a single-channel image with even edge length $\boldsymbol{X} \in\mathbb{R}^{2H \times 2W}$. Denote the 2D discrete Fourier transform as $\mathcal{F}(\cdot)$. Without loss of generality, we assume that the coordinate ranges are $\{-H, -H+1, \ldots, H-1\}$ and  $\{-W, -W+1, \ldots, W-1\}$. The value of the pixel at the position $[u,v]$ in the frequency map $\boldsymbol{F}=\mathcal{F}(\boldsymbol{X})$ is computed by
\begin{align*}
    \boldsymbol{F}[u,v]= \sum_{x=-H}^{H-1} \sum_{y=-W}^{W-1} \boldsymbol{X}[x,y]\cdot \exp \left( -j2\pi \left(\frac{ux}{2H}+\frac{vy}{2W} \right) \right).
\end{align*}
Similarly, the inverse 2D discrete Fourier transform $\boldsymbol{X} = \mathcal{F}^{-1}(\boldsymbol{F})$ is defined by
\begin{align*}
    \boldsymbol{X}[x,y]= \frac{1}{4HW} \sum_{u=-H}^{H-1} \sum_{v=-W}^{W-1} \boldsymbol{F}[u,v]\cdot \exp \left(j2\pi \left(\frac{ux}{2H}+\frac{vy}{2W} \right) \right).
\end{align*}
Denote the low-frequency cropping operation parametrized by the output size $(2H',2W')$ as $\mathcal{C}_{H',W'}(\cdot)$, which gives outputs by simple cropping:
$$\mathcal{C}_{H',W'}(\boldsymbol{F})[u,v] = \frac{H'W'}{HW}\cdot \boldsymbol{F}[u,v].$$
Note that here $u\in\{-H, -H+1, \ldots, H-1\}, v\in\{-W, -W+1, \ldots, W-1\}$, and this operation simply copies the central area of $\boldsymbol{F}$ with a scaling ratio. The scaling ratio $\frac{H'W'}{HW}$ is a natural term from the change of total energy in the pixels, since the number of pixels shrinks by the ratio of $\frac{H'W'}{HW}$.

Also, denote the down-sampling operation parametrized by the ratio $r\in(0,1]$ as $\mathcal{D}_r(\cdot)$. For simplicity, we first consider the case where $r=\frac{1}{k}$ for an integer $k\in\mathbb{N}^+$, and then extend our conclusions to the general cases where $r\in(0,1]$.
In real applications, there are many different down-sampling strategies using different interpolation methods, \emph{i.e.}, nearest, bilinear, bicubic, etc. When $k$ is an integer, these operations can be modeled as \textit{using a constant convolution kernel to aggregate the neighborhood pixels}. Denote this kernel's parameter as
$\boldsymbol{w}_{s,t}$ where $s,t\in\{0, 1, \ldots, k-1\}$ and $\sum_{s=0}^{k-1} \sum_{t=0}^{k-1} \boldsymbol{w}_{s,t}=\frac{1}{k^2}$. Then the down-sampling operation can be represented as
\begin{align*}
    \mathcal{D}_{1/k}(\boldsymbol{X})[x',y'] = \sum_{s=0}^{k-1} \sum_{t=0}^{k-1} \boldsymbol{w}_{s,t} \cdot \boldsymbol{X}[kx'+s, ky'+t].
\end{align*}


\subsection{Propositions}
Now we are ready to demonstrate the difference between the two operations and prove our claims. We start by considering shrinking the image size by $k$ and $k$ is an integer. Here the low-frequency region of an image $\boldsymbol{X}\in \mathbb{R}^{H\times W}$ refers to the signals within range $[-H/k, H/k-1] \times [-W/k, W/k-1]$ in $\mathcal{F}(\boldsymbol{X})$, while the rest is named as the high-frequency region.

\textbf{{Proposition 1.1.}}
\textit{Suppose that the original image is $\boldsymbol{X}$, and that the image generated from the low-frequency cropping operation is $\boldsymbol{X}_c = \mathcal{F}^{-1}\circ \mathcal{C}_{H/k,W/k} \circ \mathcal{F} (\boldsymbol{X}), k\in\mathbb{N}^+$. We have that all the signals in the spectral map of $\boldsymbol{X}_c$ is only from the low frequency region of $\boldsymbol{X}$, while we can always recover $\mathcal{C}_{H/k,W/k} \circ \mathcal{F} (\boldsymbol{X})$ from $\boldsymbol{X}_c$.}

\textbf{\textit{Proof.}} The proof of this proposition is simple and straightforward. Take Fourier transform on both sides of the above transformation equation, we get
$$ \mathcal{F}(\boldsymbol{X}_c) = \mathcal{C}_{H/k,W/k} \circ \mathcal{F} (\boldsymbol{X}). $$
Denote the spectral of $\boldsymbol{X}_c$ as $\boldsymbol{F}_c = \mathcal{F}(\boldsymbol{X}_c)$ and similarly $\boldsymbol{F}= \mathcal{F} (\boldsymbol{X})$. According to our definition of the cropping operation, we know that
$$ \boldsymbol{F}_c [u,v] = \frac{H/k\cdot W/k}{HW} \cdot \boldsymbol{F}[u,v] = \frac{1}{k^2} \cdot \boldsymbol{F}[u,v].$$
Hence, the spectral information of $\boldsymbol{X}_c$ simply copies $\boldsymbol{X}$'s low frequency parts and conducts a uniform scaling by dividing $k^2$.

\textbf{{Proposition 1.2.}}
\textit{Suppose that the original image is $\boldsymbol{X}$, and that the image generated from the down-sampling operation is $\boldsymbol{X}_{\textnormal{d}} = \mathcal{D}_{1/k} (\boldsymbol{X}), k\in\mathbb{N}^+$. We have that the signals in the spectral map of $\boldsymbol{X}_{\textnormal{d}}$ have a non-zero dependency on the high frequency region of $\boldsymbol{X}$.}

\textbf{\textit{Proof.}} Taking Fourier transform on both sides, we have
$$ \mathcal{F}(\boldsymbol{X}_{\textnormal{d}}) = \mathcal{F}(\mathcal{D}_{1/k} (\boldsymbol{X})). $$
For any $u\in[-H/k, H/k-1], v\in[-W/k, W/k-1]$, according to the definition we have
\begin{align*}
    \mathcal{F}(\boldsymbol{X}_{\textnormal{d}})[u,v] =& \sum_{x=-H/k}^{H/k-1} \sum_{y=-W/k}^{W/k-1} \mathcal{D}_{1/k}(\boldsymbol{X})[x,y]\cdot \exp \left( -j2\pi \left(\frac{kux}{2H}+\frac{kvy}{2W} \right) \right) \\
    =& \sum_{x=-H/k}^{H/k-1} \sum_{y=-W/k}^{W/k-1}  \sum_{s=0}^{k-1} \sum_{t=0}^{k-1} \boldsymbol{w}_{s,t} \cdot \boldsymbol{X}[kx+s, ky+t] \cdot \exp \left( -j2\pi \left(\frac{kux}{2H}+\frac{kvy}{2W} \right) \right), \tag{*1}
\end{align*}
while at the same time we have the inverse DFT for $\boldsymbol{X}$:
\begin{align*}
    \boldsymbol{X}[x,y] =&  \frac{1}{2H \cdot 2W}\sum_{u'=-H}^{H-1} \sum_{v'=-W}^{W-1} 
    \boldsymbol{F}[u',v']\cdot \exp \left( j2\pi \left(\frac{u'x}{2H}+\frac{v'y}{2W} \right) \right). \tag{*2}
\end{align*}
Plugging (*2) into (*1), it is easy to see that essentially each $\boldsymbol{F}_{\textnormal{d}}(u,v)=\mathcal{F}(\boldsymbol{X}_{\textnormal{d}})[u,v]$ is a linear combination of the original signals $\boldsymbol{F}[u',v']$. Namely, it can be represented as
$$ \boldsymbol{F}_{\textnormal{d}}(u,v) = \sum_{u'=-H}^{H-1} \sum_{v'=-W}^{W-1} \alpha(u,v,u',v') \cdot \boldsymbol{F}(u',v'). $$
Therefore, we can compute the dependency weight for any given tuple $(u,v,u',v')$ as

\begin{align*}
    \alpha(u,v,u',v') =& \frac{1}{4HW} \sum_{x=-H}^{H-1} \sum_{y=-W}^{W-1} \boldsymbol{w}_{x_r,y_r}
    \cdot  \exp \left( -j2\pi \left(\frac{u x_p}{2H}+\frac{v y_p}{2W} \right) \right) \cdot \exp \left( j2\pi \left(\frac{u'x}{2H}+\frac{v'y}{2W} \right) \right) \\
     =& \frac{1}{4HW} \sum_{x=-H}^{H-1} \sum_{y=-W}^{W-1} \boldsymbol{w}_{x_r,y_r}
    \cdot  \exp \left( j2\pi \left(\frac{u'x-u x_p}{2H}+\frac{v'y-v y_p}{2W} \right) \right),
\end{align*}
where $x_r=x\ \operatorname{mod}\ k, x_p=x-x_r$, same for $y_r, y_p$. Further deduction shows

\begin{align*}
  \alpha(u,v,u',v') =& \frac{1}{4HW} \sum_{x=-H}^{H-1} \sum_{y=-W}^{W-1} \boldsymbol{w}_{x_r,y_r}
  \cdot  \exp \left( j2\pi \left(\frac{(u'-u) x_p + u' x_r}{2H}+\frac{(v'-v) y_p + v' y_r}{2W} \right) \right) \\
  =&\frac{1}{4HW}\sum_{x'=-H/k}^{H/k-1} \sum_{y'=-W/k}^{W/k-1} \sum_{s=0}^{k-1} \sum_{t=0}^{k-1} \boldsymbol{w}_{s,t}
  \cdot  \exp \left( j2\pi k \left(\frac{(u'-u) x'}{2H}+\frac{(v'-v)y'}{2W} \right) \right)  \cdot  \exp \left( j2\pi \left(\frac{u's}{2H}+\frac{v't}{2W} \right) \right) \\
  =&\frac{1}{4HW} \sum_{x'=-H/k}^{H/k-1} \sum_{y'=-W/k}^{W/k-1}  \exp \left( j2\pi k\left(\frac{(u'-u) x'}{2H}+\frac{(v'-v)y'}{2W} \right) \right)   \sum_{s=0}^{k-1} \sum_{t=0}^{k-1} \boldsymbol{w}_{s,t} \cdot  \exp \left( j2\pi \left(\frac{u's}{2H}+\frac{v't}{2W} \right) \right). \\
\end{align*}
Denote $\beta(u',v') = \sum_{s=0}^{k-1} \sum_{t=0}^{k-1} \boldsymbol{w}_{s,t} \cdot  \exp \left( j2\pi \left(\frac{u's}{2H}+\frac{v't}{2W} \right) \right)$, which is a constant conditioned on $(u',v')$. Then we know
\begin{align*}
  \alpha(u,v,u',v') =& \frac{\beta(u',v')}{4HW} \sum_{x'=-H/k}^{H/k-1} \sum_{y=-W/k}^{W/k-1}  \exp \left( j2\pi k \left(\frac{(u'-u) x'}{2H}+\frac{(v'-v)y'}{2W} \right) \right) \\
 =&  \frac{\beta(u',v')}{4HW} \sum_{x'=-H/k}^{H/k-1} \exp \left( j2\pi k \left(\frac{(u'-u) x'}{2H}\right)\right) \sum_{y'=-W/k}^{W/k-1}  \exp \left(j2\pi k \left(\frac{(v'-v)y'}{2W} \right) \right)  \\
 =& \begin{cases}
 \frac{\beta(u',v')}{k^2}, &u'-u=a\cdot \frac{2H}{k},v'-v=b\cdot\frac{2W}{k},  \quad a,b\in\mathbb{Z} \\
 0, & \text{otherwise}
 \end{cases}. \tag{*3}
\end{align*}
In general, $\beta(u',v')\not=0$ when $u'\not=c\cdot\frac{2H}{k}, v'\not=d\cdot\frac{2W}{k}, c,d\in\mathbb{Z}, cd\not=0$.
Hence, when $\frac{2H}{k} | (u'-u), \frac{2W}{k} | (v'-v)$, we have $\alpha(u,v,u',v')\not = 0$ given $uv\not=0$, while $\alpha(u,0,u',0)\not = 0$ given $u\not=0$, $\alpha(0,v,0,v')\not = 0$ given $v\not=0$. Therefore, the image generated through down-sampling contains mixed information from both low frequency and high frequency, since most signals have a non-zero dependency on the global signals of the original image.

\textbf{{Proposition 1.3.}}
\textit{The conclusions of Proposition 1.1 and Proposition 1.2 still hold when $k\in\mathbb{Q}^+$ is not an integer.}

\textbf{\textit{Proof.}}
It is obvious that Proposition 1.1 can be naturally extended to $k\in\mathbb{Q}^+$. Therefore, here we focus on Proposition 1.2. First, consider up-sampling an image $\boldsymbol{X}$ by $m\in\mathbb{N}^+$ times with the nearest interpolation, namely
$$\boldsymbol{X}_{\textnormal{up}}[mx+s,my+t] = \boldsymbol{X}[x,y], \quad s,t\in\{0, 1, \ldots, m-1\}. $$
Taking Fourier transform, we have
\begin{align*}
    \mathcal{F}(\boldsymbol{X}_{\textnormal{up}})[u,v] =& \sum_{x=-mH}^{mH-1} \sum_{y=-mW}^{mW-1} \boldsymbol{X}_{\textnormal{up}}[x,y]\cdot \exp \left( -j2\pi \left(\frac{ux}{2mH}+\frac{kvy}{2mW} \right) \right) \\
    =& \sum_{x=-H}^{H-1} \sum_{y=-W}^{W-1}  \sum_{s=0}^{m-1} \sum_{t=0}^{m-1} \boldsymbol{X}[x, y] \cdot \exp \left( -j2\pi \left(\frac{u(mx+s)}{2mH}+\frac{v(my+t)}{2mW} \right) \right). \tag{*4}
\end{align*}
Similar to the proof of Proposition 1.2, by plugging (*2) into (*4), it is easy to see that $\boldsymbol{F}_{\textnormal{up}}(u,v)=\mathcal{F}(\boldsymbol{X}_{\textnormal{up}})[u,v]$ is a linear combination of the signals from the original image. Namely, we have
$$ \boldsymbol{F}_{\textnormal{up}}(u,v) = \sum_{u'=-H}^{H-1} \sum_{v'=-W}^{W-1} \alpha_{\textnormal{up}}(u,v,u',v') \cdot \boldsymbol{F}(u',v'). $$
Given any $(u,v,u',v')$, $\alpha_{\textnormal{up}}(u,v,u',v')$ can be computed as
\begin{align*}
  \alpha_{\textnormal{up}}(u,v,u',v') =& \frac{1}{4HW} \sum_{x=-H}^{H-1} \sum_{y=-W}^{W-1} \sum_{s=0}^{m-1} \sum_{t=0}^{m-1}
   \exp \left( -j2\pi \left(\frac{u(mx+s)}{2mH}+\frac{v(my+t)}{2mW} \right) \right)
  \cdot \exp \left( j2\pi \left(\frac{u'x}{2H}+\frac{v'y}{2W} \right) \right) 
  \\
   =& \frac{1}{4HW} \sum_{x=-H}^{H-1} \sum_{y=-W}^{W-1} 
   \exp \left( j2\pi \left(\frac{(u'-u) x}{2H}+\frac{(v'-v)y}{2W} \right) \right)
   \sum_{s=0}^{m-1} \sum_{t=0}^{m-1}
    \exp \left( -j2\pi \left(\frac{us}{2mH}+\frac{vt}{2mW} \right) \right).
\end{align*}
Denote $\beta_{\textnormal{up}}(u,v) = \sum_{s=0}^{m-1} \sum_{t=0}^{m-1}  \exp \left( -j2\pi \left(\frac{us}{2mH}+\frac{vt}{2mW} \right) \right)$, which is a constant conditioned on $(u,v)$. Then we know
\begin{align*}
  \alpha_{\textnormal{up}}(u,v,u',v') =& 
  \frac{\beta_{\textnormal{up}}(u,v)}{4HW} 
  \sum_{x=-H}^{H-1} \sum_{y=-W}^{W-1}  \exp \left( j2\pi \left(\frac{(u'-u) x}{2H}+\frac{(v'-v)y}{2W} \right) \right) \\
 =&  \frac{\beta_{\textnormal{up}}(u,v)}{4HW}  
 \sum_{x=-H}^{H-1} \exp \left( j2\pi \left(\frac{(u'-u) x}{2H} \right) \right) 
 \sum_{y=-W}^{W-1} \exp \left( j2\pi \left(\frac{(v'-v) y}{2W} \right) \right)   \\
 =& \begin{cases}
  \beta_{\textnormal{up}}(u,v), &u'-u=a\cdot 2H, v'-v=b\cdot 2W,  \quad a,b\in\mathbb{Z} \\
 0, & \text{otherwise}
 \end{cases}.  \tag{*5}
\end{align*}
Since $-H \leq u \leq H-1, -W \leq v \leq W-1$, we have $\beta_{\textnormal{up}}(u,v)\not=0$. Thus, we have $\alpha_{\textnormal{up}}(u,v,u',v')\not = 0$ when $2H | (u'-u), 2W | (v'-v)$. 

Now we return to Proposition 1.3. Suppose that the original image is $\boldsymbol{X}$, and that the image obtained through down-sampling is $\boldsymbol{X}_{\textnormal{d}} = \mathcal{D}_{1/k} (\boldsymbol{X})$, where $k\in\mathbb{Q}^+$ may not be an integer. We can always find two integers $m_0$ and $k_0$ such that $\frac{k_0}{m_0} = k$. Consider first up-sampling $\boldsymbol{X}$ by $m_0$ times with the nearest interpolation and then performing down-sampling by $k_0$ times. By combining (*3) and (*5), it is easy to verify that Proposition 1.3 is true.  $\hfill\qedsymbol$

\section{Relationships between Low-pass Filtering and 2D Sinc Convolution}
\label{sec:2d_sinc}

In this section, we theoretically demonstrate that the low-pass filtering with a $B\!\times\!B$ square filter is equivalent to performing a convolution operation on the image with a kernel function called 2D sinc function. This is actually a classical result in the field of signal processing, indicating that low-pass filtering can be efficiently implemented by the convolution with a certain kernel. 
\subsection{Preliminaries}
The basic preliminaries (\emph{e.g.}, 2D discrete Fourier transform and the inverse transform) are the same as Appendix \textcolor{red}{C.1}). In particular, here we define the procedure of low-pass filtering with a $B\!\times\!B$ square filter as $\mathcal{T}_{B, B}^{\textnormal{low}}(\cdot)$, namely,
\begin{equation*}
  \mathcal{T}_{B, B}^{\textnormal{low}}(\boldsymbol{F})[u,v] =
        \begin{cases}
        \boldsymbol{F}[u,v],  & -\frac{B}{2} \le u, v \le \frac{B}{2}-1  \\
        0, & \textnormal{otherwise}
        \end{cases}.
\end{equation*}


\subsection{Proposition}

\textbf{Proposition 2.1.} \textit{Low-pass filtering with a $B\!\times\!B$ square filter is equivalent to performing a constant-kernel convolution on the original image.}

\noindent\textit{\textbf{Proof.}}
Taking inverse 2D discrete Fourier transform over $\mathcal{T}_{B, B}^{\textnormal{low}}(\boldsymbol{F})$, we have
\begin{align*}
  \mathcal{F}^{-1}(\mathcal{T}_{B, B}^{\textnormal{low}}(\boldsymbol{F}))[x,y] =& \frac{1}{4HW} \sum_{u=-\frac{B}{2}}^{\frac{B}{2}-1} \sum_{v=-\frac{B}{2}}^{\frac{B}{2}-1} \boldsymbol{F}[u,v] \cdot \exp\left( j2\pi \left( \frac{ux}{2H} + \frac{vy}{2W} \right) \right)\\
    =& \frac{1}{4HW} \sum_{u=-\frac{B}{2}}^{\frac{B}{2}-1} \sum_{v=-\frac{B}{2}}^{\frac{B}{2}-1} \exp\left( j2\pi \left( \frac{ux}{2H} + \frac{vy}{2W} \right) \right) \sum_{s=-H}^{H-1} \sum_{t=-W}^{W-1} \boldsymbol{X}[s,t] \cdot \exp\left( -j2\pi \left( \frac{us}{2H} + \frac{vt}{2W} \right) \right) \\
    =& \frac{1}{4HW} \sum_{u=-\frac{B}{2}}^{\frac{B}{2}-1} \sum_{v=-\frac{B}{2}}^{\frac{B}{2}-1} \sum_{s=-H}^{H-1} \sum_{t=-W}^{W-1} \exp\left( j2\pi \left( \frac{u(x-s)}{2H} + \frac{v(y-t)}{2W} \right) \right)  \boldsymbol{X}[s,t].
\end{align*}
Here we can see, every pixel of the new transformed image $\mathcal{F}^{-1}(\mathcal{T}_{B, B}^{\textnormal{low}}(\boldsymbol{F}))$ is essentially a linear combination of the original pixels, which can be represented as
\begin{align*}
  \mathcal{F}^{-1}(\mathcal{T}_{B, B}^{\textnormal{low}}(\boldsymbol{F}))[x,y] =& \sum_{s=-H}^{H-1} \sum_{t=-W}^{W-1} \alpha(x,y,s,t) \boldsymbol{X}[s,t]. \\
     \alpha(x,y,s,t) =& \frac{1}{4HW} \sum_{u=-\frac{B}{2}}^{\frac{B}{2}-1} \sum_{v=-\frac{B}{2}}^{\frac{B}{2}-1} \exp\left( j2\pi \left( \frac{u(x-s)}{2H} + \frac{v(y-t)}{2W} \right) \right).
\end{align*}
It can be observed that the value of $\alpha(x,y,s,t)$ only depends on the relative position $(x-s, y-t)$. So we further simplify the notation of $\alpha$ using the relative position of two points: $\alpha(x',y'), x'=x-s, y'=y-t$, where 
\begin{align*}
    \alpha(x',y') =& \frac{1}{4HW} \sum_{u=-\frac{B}{2}}^{\frac{B}{2}-1} \sum_{v=-\frac{B}{2}}^{\frac{B}{2}-1} \exp\left( j2\pi \left( \frac{ux'}{2H} + \frac{vy')}{2W} \right) \right) \\
    =& \frac{1}{8HW} \sum_{u=-\frac{B}{2}}^{\frac{B}{2}-1} \sum_{v=-\frac{B}{2}}^{\frac{B}{2}-1} \exp\left( j2\pi \left( \frac{ux'}{2H} + \frac{vy')}{2W} \right) \right) + \exp\left( -j2\pi \left( \frac{ux'}{2H} + \frac{vy')}{2W} \right) \right)\\
    =& \frac{1}{4HW} \sum_{u=-\frac{B}{2}}^{\frac{B}{2}-1} \sum_{v=-\frac{B}{2}}^{\frac{B}{2}-1} \cos \left(  \frac{\pi u x'}{H} + \frac{\pi vy'}{W} \right).
\end{align*}
Therefore, we prove that $\mathcal{T}_{B, B}^{\textnormal{low}}(\cdot)$ corresponds to a convolution operation on $\boldsymbol{X}$ in the spatial domain with the kernel $\alpha(\cdot, \cdot)$.

\subsection{Infinite Limit}
To better understand this kernel function, we can fix the ratio $\gamma=\frac{B}{2H}$ as a constant, and let $H\to\infty$. For the sake of simplicity, we assume $H=W=N$, then the function $\alpha(x',y')$ will converge to a limit function as 
\begin{align*}
    \alpha(x',y')=& \lim_{N\to \infty} \frac{1}{(2N)^2} \sum_{s=-\frac{B}{2}}^{\frac{B}{2}-1} \sum_{t=-\frac{B}{2}}^{\frac{B}{2}-1} \exp \left( -j2\pi \left(\frac{x' s}{2N}+\frac{y' t}{2N} \right) \right) \\
    =& \lim_{N\to \infty} \frac{1}{(2N)^2} \sum_{s=-\gamma N}^{\gamma N} \sum_{t=-\gamma N}^{\gamma N} \exp \left( -j2\pi \left(x'\cdot \frac{s}{2N}+y' \cdot \frac{t}{2N} \right) \right) \\
    =&\frac{1}{(4\gamma)^2} \int_{-\gamma}^{\gamma} \int_{-\gamma}^{\gamma} \exp(-j2\pi(x' s+ y' t)) \mathrm{d}s\  \mathrm{d}t \\
    =&\frac{1}{(4\gamma)^2} \int_{-\gamma}^{\gamma}\exp(-j2\pi x' s) \mathrm{d}s \int_{-\gamma}^{\gamma} \exp(-j2\pi y' t)\ \mathrm{d}t \\
    =& \frac{\sin(2\pi \gamma x')\sin(2\pi \gamma y')}{4\pi^2 \gamma^2 x' y' }.
\end{align*}
This limit function is called 2D sinc function. It is a classical function in signal processing, since it is the Fourier transform and inverse Fourier transform of the 2D square wave function. Here, $\gamma$ controls the width of the kernel. The larger $B$ is, the larger $\gamma$ is, and the 2D sinc function becomes more strike-like at point $(0,0)$, which means every pixel only takes value from itself.




\end{document}